\documentclass[10pt,twocolumn,letterpaper]{article}

\usepackage[pagenumbers]{wacv} 

\usepackage{graphicx}
\usepackage{booktabs}

\definecolor{wacvblue}{rgb}{0.21,0.49,0.74}
\usepackage[pagebackref,breaklinks,colorlinks,allcolors=wacvblue]{hyperref}

\usepackage[accsupp]{axessibility}  

\usepackage[table]{xcolor}
\usepackage{glossaries}

\glsdisablehyper

\newacronym{bev}{BEV}{Bird's Eye View}
\newacronym{nerf}{NeRF}{Neural Radiance Fields}
\newacronym{lidar}{LiDAR}{Light Detection and Ranging}
\newacronym{iou}{IoU}{Intersection over Union}
\newacronym{miou}{mIoU}{mean Intersection over Union}
\newacronym{rayiou}{RayIoU}{Ray Intersection over Union}
\newacronym{vlm}{VLM}{Vision–Language Model}
\newacronym{vfm}{VFM}{Visual Foundation Model}
\newacronym{3dgs}{3DGS}{3D Gaussian Splatting}  
\definecolor{others}{rgb}{0, 0, 0}
\definecolor{barrier}{rgb}{1, 0.47058824, 0.19607843}
\definecolor{bicycle}{rgb}{1, 0.75294118, 0.79607843}
\definecolor{bus}{rgb}{1, 1, 0.0}
\definecolor{car}{rgb}{0.0, 0.58823529, 0.96078431}
\definecolor{construction}{rgb}{0, 1, 1}
\definecolor{motorcycle}{rgb}{0.78431372549 , 0.70588235294, 0}
\definecolor{pedestrian}{rgb}{1, 0, 0}
\definecolor{cone}{rgb}{1, 0.94117647, 0.58823529}
\definecolor{trailer}{rgb}{0.52941176, 0.23529412, 0}
\definecolor{truck}{rgb}{0.62745098, 0.1254902, 0.94117647}
\definecolor{driveable}{rgb}{1, 0, 1}
\definecolor{flat}{rgb}{0.54509804,0.5372549,0.5372549}
\definecolor{sidewalk}{rgb}{0.29411765,0,0.29411765}
\definecolor{terrain}{rgb}{0.58823529,0.94117647,0.31372549}
\definecolor{manmade}{rgb}{0.90196078,0.90196078,0.98039216}
\definecolor{vegetation}{rgb}{0,0.68627451,0}

\definecolor{roadcolor}{RGB}{234,51,246}
\definecolor{sidewalkcolor}{RGB}{68,8,72}
\definecolor{parkingcolor}{RGB}{241,156,249}
\definecolor{othergroundcolor}{RGB}{160,32,76}
\definecolor{buildingcolor}{RGB}{246,202,69}
\definecolor{carcolor}{RGB}{111,149,238}
\definecolor{truckcolor}{RGB}{74,32,172}
\definecolor{bicyclecolor}{RGB}{136,227,242}
\definecolor{motorcyclecolor}{RGB}{37,59,146}
\definecolor{othervehiclecolor}{RGB}{96,81,242}
\definecolor{vegetationcolor}{RGB}{79, 173, 50}
\definecolor{terraincolor}{RGB}{171, 238, 105}
\definecolor{personcolor}{RGB}{234, 60, 49}
\definecolor{fencecolor}{RGB}{238, 128, 69}
\definecolor{polecolor}{RGB}{252, 241, 161}
\definecolor{trafficsigncolor}{RGB}{233, 51, 35}
\definecolor{other-struct.color}{RGB}{255, 150, 0}
\definecolor{other-objectcolor}{RGB}{50, 255, 255}

\definecolor{winered}{rgb}{1, 0, 0} 

\definecolor{ceiling}{RGB}{214,  38, 40}   %
\definecolor{floor}{RGB}{43, 160, 4}     %
\definecolor{wall}{RGB}{158, 216, 229}  %
\definecolor{window}{RGB}{114, 158, 206}  %
\definecolor{chair}{RGB}{204, 204, 91}   %
\definecolor{bed}{RGB}{255, 186, 119}  %
\definecolor{sofa}{RGB}{147, 102, 188}  %
\definecolor{table}{RGB}{30, 119, 181}   %
\definecolor{tvs}{RGB}{160, 188, 33}   %
\definecolor{furniture}{RGB}{255, 127, 12}  %
\definecolor{objects}{RGB}{196, 175, 214} %

\usepackage{amsmath}
\usepackage{amssymb}
\usepackage{amsfonts}
\usepackage{multirow}
\usepackage{verbatim}
\usepackage{adjustbox}

\usepackage{subcaption}
\usepackage{graphicx}

\definecolor{winered}{rgb}{1, 0, 0} 
\definecolor{lightgray}{rgb}{0.85, 0.85, 0.85}

\usepackage{float}

\usepackage{tocloft}
\makeatletter
\renewcommand*\l@subsection{\@dottedtocline{2}{0.5em}{3.0em}}
\makeatother

\usepackage{pifont}
\newcommand{\cmark}{\ding{51}}%
\newcommand{\xmark}{\ding{55}}%
\newcommand{\myangle}{90}

\newcommand{\method}{{Super\-Quadric\-Occ}}
\newcommand{\methodShort}{{SQOcc}}
\newcommand{\render}{{Super\-Quadric\-Occ-Render}}
\newcommand{\renderShort}{{SQOcc-R}}

\title{SuperQuadricOcc: Real-Time Self-Supervised Semantic Occupancy Estimation with Superquadric Volume Rendering}

\author{
Seamie Hayes$^{1}$,
Alexandre Boulch$^{2}$,
Andrei Bursuc$^{2}$,
Reenu Mohandas$^{1}$,
Ganesh Sistu$^{1}$,\\
Tim Brophy$^{1}$,
Ciaran Eising$^{1}$\\[12pt]
$^{1}$University of Limerick, 
$^{2}$Valeo.ai
}

\begin{document}
\maketitle

\begin{abstract}
Self-supervision for semantic occupancy estimation is appealing as it removes the need for labour-intensive manual annotation, thus allowing one to scale to larger autonomous driving datasets.
Superquadrics offer an expressive shape family very suitable for this task, yet their deployment in a self-supervised setting has been hindered by the lack of efficient rendering methods to bridge the 3D scene representation with 2D training pseudo-labels.
To address this, we introduce \method{}, the first self-supervised occupancy model to leverage superquadrics for scene representation.
To overcome the rendering limitation, we propose \render{}, a real-time volume renderer that preserves the fidelity of the superquadric shape during rendering.
It relies on spatial superquadric–voxel indexing, restricting each ray sample to query only nearby superquadrics, thereby greatly reducing memory usage and computational cost.
Using drastically fewer primitives than prior Gaussian-based methods, \method{} achieves state-of-the-art RayIoU on Occ3D-nuScenes with real-time inference and a substantially reduced memory footprint. It also outperforms Gaussian-based baselines on downstream occupancy forecasting and trajectory estimation. 
Our code is open-source: \href{https://github.com/seamie6/SuperQuadricOcc}{ SuperQuadricOcc GitHub}.

\end{abstract}

\addtocontents{toc}{\protect\setcounter{tocdepth}{-1}}

\section{Introduction}
\label{sec:intro}
\begin{figure*}
    \centering
    \includegraphics[width=\textwidth]{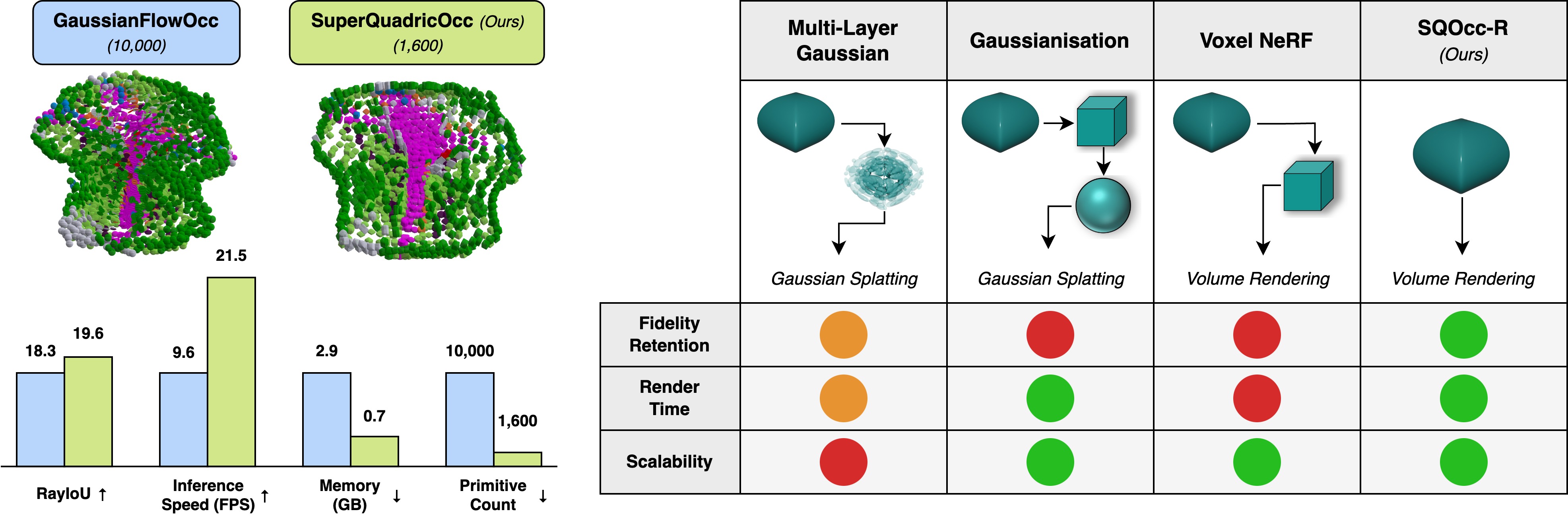}
     \caption{\textbf{\method{} and \render.} 
     \emph{\methodShort{}} achieves higher RayIoU, faster inference, and lower memory consumption with far fewer primitives than GaussianFlowOcc \cite{boeder2025gaussianflowocc} highlighting the efficiency of the superquadric scene representation.
     \emph{\renderShort{}}'s direct rendering of primitives preserves geometric accuracy in contrast to other rendering methods (Gaussian or voxels). Green, yellow, and red are relative comparisons denoting good, average, and poor, respectively.}
    \label{fig:hook}
\end{figure*}

Safe autonomous perception requires accurate modeling of complex scenes. Current research primarily focuses on semantic occupancy estimation, which represents the world as a discretised voxel grid. This addresses limitations of previous spatial understanding methods, such as the rigid cuboid prediction space of 3D object detection and the lack of verticality in \gls{bev}.

In a supervised setting, where the predictions and the ground truth space are aligned~\cite{huang2024gaussianformer}, semantic occupancy estimation has evolved from dense voxel-based formats to more efficient Gaussian representations, and most recently, to superquadrics \cite{zuo2025quadricformer, yu2026superocc}.
The latter's reduced-memory footprint and high inference speed make them highly attractive for real-time autonomous perception.
In contrast, self-supervised methods do not rely on comprehensive ground-truth annotations, significantly reducing the need for costly manual labelling and improving scalability for large datasets and models. However, supervision is now derived from 2D pseudo-labels, and hence rendering techniques are required to bridge 3D scene representations with 2D supervision space \cite{boeder2025gaussianflowocc}. While voxel-based \cite{zhang2023occnerf} and Gaussian-based methods \cite{jiang2025gausstr} directly benefit rendering techniques such as volume rendering and \gls{3dgs}~\cite{kerbl20233dgs}, superquadrics lack a native rasterisation method.

We address these challenges by introducing \method{} (\methodShort{} for short) and \render{} (\renderShort{} for short).
To our knowledge, the former is the first self-supervised model based on superquadrics for semantic occupancy estimation. 
The latter is a real-time CUDA-accelerated superquadric volume renderer specifically designed for superquadrics.
It relies on spatial indexing to establish a superquadric–voxel correspondence which restricts the influence of superquadrics to nearby regions, similar to the efficiency of screen-space rasterisation in \gls{3dgs} \cite{kerbl20233dgs}. 
As a result, each ray sample attends only to superquadrics with a significant contribution,
substantially reducing computational overhead and enabling the rendering of large numbers of superquadrics.

\methodShort{} achieves state-of-the-art results on Occ3D-nuScenes \cite{tian2023occ3d} across scene reconstruction, occupancy forecasting, and trajectory estimation, while enabling real-time inference and a substantially reduced memory footprint through its compact superquadric representation.
Our rendering method demonstrates superior performance in quality metrics due to better preserving geometric fidelity and significantly reduces memory consumption compared to baseline render methods.
Furthermore, it proves scalability with increasing superquadric count. 
As illustrated in \autoref{fig:hook}, our major contributions are the following:
\begin{enumerate}
    \item \textbf{Scene Representation.}
    We introduce \methodShort{}, the first self-supervised semantic occupancy model to represent scenes using superquadrics.
    \item \textbf{View Renderer.} We develop \renderShort{}, a superquadric volume renderer that utilises spatial indexing to efficiently manage thousands of primitives and achieve 58 FPS real-time inference for six nuScenes cameras.
    \item \textbf{Performance.} We obtain state-of-the-art performance on Occ3D-nuScenes, 
    achieving 19.6 RayIoU and 17.1 mIoU. Our model reaches real-time inference at 21.5 FPS with a memory footprint of 0.7 GB.
     \item \textbf{Downstream Tasks.} Compared to Gaussian-based methods, we demonstrate superior performance on occupancy forecasting and trajectory estimation.
\end{enumerate}

\begin{figure*}[t]
    \centering
    \includegraphics[width=\textwidth]{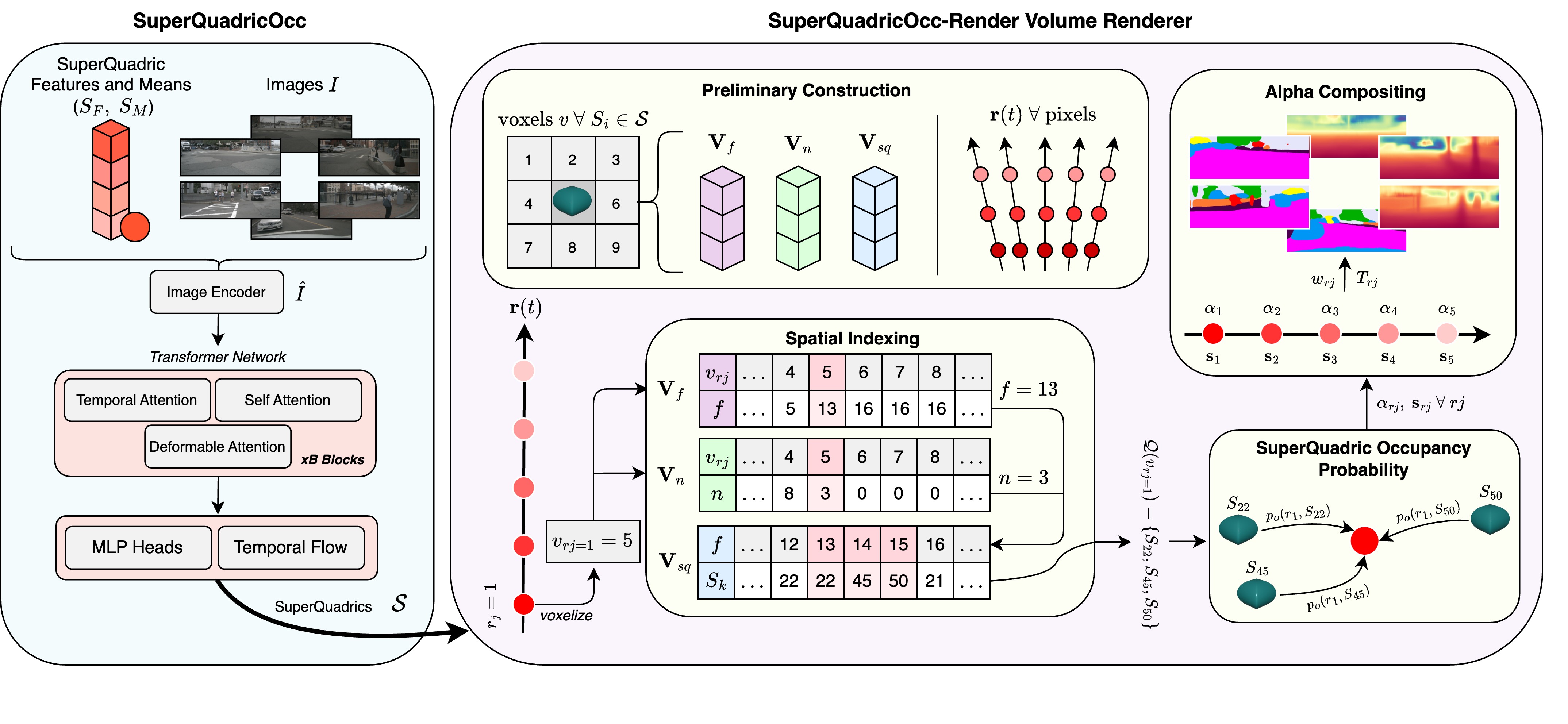}
    \vspace{-4mm}
     \caption{\textbf{Model architecture.} In \method{}, the transformer network refines superquadric means $S_M$ and features $S_F$, which are eventually transformed into full primitives using MLPs for rendering. In \render{}, following preliminary construction, we iterate over each ray sample $r_j$, and query the spatial indexing tensors $\mathbf{V}_{sq}$, $\mathbf{V}_{n}$, $\mathbf{V}_{f}$ to attain the list of superquadrics $\mathcal{Q}(v_{rj})$. We calculate the ray sample values, $\alpha_{rj}$, $\mathbf{s}_{rj}$, and perform alpha compositing for the final semantic $\hat{S}$ and depth $\hat{D}$ renders. During evaluation, superquadrics $\mathcal{S}$ are voxelized.}
    \label{fig:model_arch}
\end{figure*}

\section{Related Work}
\textbf{\emph{Semantic Occupancy Estimation.}}
Interest in semantic occupancy estimation for autonomous vehicle perception has been enabled by large-scale occupancy datasets \cite{wang2023openoccupancy, tian2023occ3d, wei2023surroundocc, tong2023scene}, built upon benchmarks such as nuScenes \cite{nuScenes} and SemanticKITTI \cite{behley2019semantickitti}.
Scene representations have rapidly evolved from strictly voxel-based methods \cite{hou2024fastocc, zuo2023pointocc, wang2024opus, ma2024cotr}, to Gaussian-based approaches \cite{huang2024gaussianformer, huang2025gaussianformer2, zuo2025gaussianworld, zhao2025gaussianformer3d}, and more recently to superquadric-based representations \cite{zuo2025quadricformer, yu2026superocc}.
This progression is largely driven by the advantages of sparse scene representations, including reduced memory consumption and faster inference times. 

\noindent \textbf{\emph{Self-Supervision for Semantic Occupancy.}}
Beyond representation choice, supervision strategy forms another key axis of development. 
Fully supervised methods initially dominated the field \cite{huang2023tri, li2023fb, liao2025stcocc}. 
However, more recently, self-supervised methods have become increasingly common, particularly due to the lack of need for manual annotation, greatly reducing labour costs and increasing scalability.
Such methods rely solely on 2D pseudo-labels generated by \glspl{vlm} \cite{radford2021learning, barsellotti2025talking} and \glspl{vfm} \cite{barsellotti2025talking, hu2024metric3d, zhang2023simple, ren2024grounded} for supervision, with more recent methods utilizing 3D pseudo-labels \cite{boeder2026shelfocc, lilja2026queryocc}. 
Voxel-based approaches \cite{zhang2023occnerf, huang2024selfocc} suffer from high computational costs and slow inference, driven by large convolutional operations and inefficient supervision via voxel-wise volume rendering.
Gaussian-based methods subsequently emerged but are hindered in inference by a reliance on large Vision Foundation Models (VFMs) \cite{jiang2025gausstr, zhang2025tt} or high primitive counts \cite{boeder2025gaussianflowocc, zhang2025tt}, both of which compromise real-time deployment and memory efficiency. 
In contrast, \methodShort{} achieves real-time inference and a reduced memory footprint, all while improving performance metrics without the need for inference-time VFMs.

\noindent \textbf{\emph{Novel View Synthesis}}
aims to render a scene or object from viewpoints different from those of the images used to extract priors.
Currently, the main approaches are \gls{nerf} and \gls{3dgs} \cite{kerbl20233dgs}.
The former uses an implicit scene representation and volume rendering, while the latter employs Gaussian-based scene representation and screen-space rasterisation.
Both approaches have been widely applied across many domains \cite{xiao2025neural, gao2022nerf, chen2025survey3dgaussiansplatting} ranging from medical imaging \cite{wang2022neural, batlle2023lightneus, chen2023cunerf, wang2024endogslam, liu2024endogaussian} and robotics \cite{rosinol2023nerf, liso2024loopy, li2023dense, deng2024compact, hong2024liv} to 3D 
reconstruction \cite{azinovic2022neural, wang2021neus, oechsle2021unisurf, hu2024gaussianavatar, jiang2024hifi4g}.

For 3D object reconstruction specifically, PartGS \cite{gao2025self} and GaussianBlock \cite{jiang2024gaussianblock} draw inspiration from work that places Gaussians on mesh surfaces \cite{guedon2024sugar, waczynska2024games}, extending this idea by positioning Gaussians on superquadric surfaces to act as a rendering proxy. 
OccNeRF \cite{zhang2023occnerf} adopts a NeRF-style voxel-volume rendering formulation, while GaussianOcc \cite{gaussianocc} introduces a voxel-to-Gaussian module for \gls{3dgs}. 
We adapt these methods as baselines for our alternate rendering approaches.

ISCO \cite{alaniz2023iterative} reconstructs single objects using superquadrics and NeRF-style volume rendering for supervision against ground-truth 2D object silhouettes. However, it is not suitable for modelling the complexity of outdoor scenes as it is limited to ten primitives maximum, naively querying every superquadric for every ray sample. In contrast, our approach is capable of handling thousands of superquadric primitives through an efficient spatial indexing strategy, enabling real-time rendering with minimal fidelity loss.

\section{Methodology}
\subsection{SuperQuadrics}
We represent the scene using a set of $N$ superquadric primitives, $\mathcal{S} = \{S_i\}_{i=1}^{N}$, where each primitive is defined as:
$S_i = \{\mathbf{m}_i, \mathbf{s}_i, \mathbf{r}_i, \boldsymbol{\varepsilon}_i, \sigma_i, \mathbf{c}_i\}$
where $\mathbf{m}$ is the center, $\mathbf{s} = (s_x, s_y, s_z)$ is the scale along the three principal axes, $\mathbf{r} \in \mathbb{R}^4$ is a quaternion rotation, and $\boldsymbol{\varepsilon} = (\varepsilon_1, \varepsilon_2)$ controls the squareness and roundness of the shape. 
$\sigma \in \mathbb{R}$ is the density, and $\mathbf{c} \in \mathbb{R}^C$ encodes semantic information over $C$ classes. 
The implicit surface equation of a canonical superquadric \cite{barr1981superquadrics} is:
\begin{equation}
f(\mathbf{x}) =
\left(
\left( \frac{x}{s_x} \right)^{\frac{2}{\varepsilon_2}}
+
\left( \frac{y}{s_y} \right)^{\frac{2}{\varepsilon_2}}
\right)^{\frac{\varepsilon_2}{\varepsilon_1}}
+
\left( \frac{z}{s_z} \right)^{\frac{2}{\varepsilon_1}}
\end{equation}
where $\mathbf{x} = (x,y,z)$ is expressed in the canonical frame. To model occupancy probability, Zuo \textit{et al.} \cite{zuo2025quadricformer} proposed an exponential decay formulation analogous to that of a Gaussian distribution.
The occupancy probability of superquadric $S$ at a 3D point $\mathbf{x}$ is defined as:
\begin{equation}
\label{eq:sq_occupancy_prob}
p_o(\mathbf{x}, S) = \exp\big(-f(\mathbf{x}_S)\big),
\quad
\mathbf{x}_S = R(\mathbf{x} - \mathbf{m})
\end{equation}
\noindent where $\mathbf{x}_S$ is $\mathbf{x}$ transformed into the canonical superquadric coordinate system, and $R$ is the rotation matrix corresponding to quaternion $\mathbf{r}$.
In this exponential formulation, the $\boldsymbol{\varepsilon}$ parameters now also control the rate of decay.
We adopt this formulation both for rendering and voxelization.

\subsection{Model Framework}
The architecture of \method{} is based on GaussianFlowOcc \cite{boeder2025gaussianflowocc}, adapted for a superquadric scene representation to allow for fair comparison of Gaussians versus superquadrics in a self-supervised setting. It is illustrated in \autoref{fig:model_arch}. We utilise a ResNet-50 backbone to extract features, $\hat{I}$, which together with learnable randomly initialised superquadric primitive features $S_F \in \mathbb{R}^{N \times D}$ and means $S_M \in \mathbb{R}^{N \times 3}$, are passed to the Transformer network, which iteratively refines $S_F$ and $S_M$ in $B$ blocks before predicting the final properties of the superquadric primitive set $\mathcal{S}$.

The Transformer network uses temporal attention $\text{TA}(S_F^{T}, S_F^{T-1})$ against previous features for efficient temporal propagation. Following this, self-attention is employed $\text{SA}(S_F^{T}, S_F^{T})$ for spatial interaction, and then deformable attention $\text{DA}(\hat{I}, S_F)$ enriches the primitive features with image features. All attention mechanisms use induced attention \cite{lee2019set}, with positional encoding of $S_M$.

After refinement, each superquadric property $\{\mathbf{s}, \mathbf{r}, \mathbf{\varepsilon}, \sigma, \mathbf{c}\}$, excluding mean, is attained by passing $S_F$ to individual lightweight MLPs. The temporal flow module propagates $S_M$ for each frame in the temporal horizon ($t \in [-T,T]$). This is achieved by passing learnable tokens $\Psi \in \mathbb{R}^{2T \times D}$ and $S_F$ to an MLP to get an offset vector $\vec{v}(t)$ for each $t$ in the horizon:

\begin{equation}
    \vec{v}(t) = \mathrm{MLP} \big(S_F \oplus \Psi(t) \big)
\end{equation}
These offsets are added to $S_M$. Following this, the superquadric sets $\mathcal{S}$ in the temporal horizon are passed to the \render{} Volume Renderer.
\subsection{\render{} Volume Renderer}
\label{subsec:sq_nerf}
In this section, we describe our proposed superquadric rendering method, which leverages CUDA for real-time rendering.
\render{} directly queries the superquadric occupancy probability equation, resulting in minimal fidelity loss compared to alternative rendering strategies. 
We first outline the spatial indexing and ray construction, followed by a detailed description of the forward and backward passes. An illustration is given in \autoref{fig:model_arch}.

\subsubsection{Preliminary Construction} 
To reduce computation while preserving rendering fidelity, we restrict each ray sample to interact only with superquadrics within a localised neighbourhood. We parameterise the scene as a voxel grid with bounds and resolution matching those defined in \autoref{subsec:dataset_metrics}. For each $S_i \in \mathcal{S}$, we construct a tensor that stores the voxel indices ($v$), within a cubic neighbourhood $\mathcal{N}_v$. This tensor is denoted as
$\mathbf{V}_c \in \mathbb{Z}^{\,N \times (2\mathcal{N}_v + 1)^3 \times 3}$.

We then apply hashing to convert $\mathbf{V}_c$ into three spatial indexing tensors 
for efficient mapping from all $v$ in the entire voxel grid to superquadric indices, $S_{k}$:
\begin{enumerate}
    \item $\mathbf{V}_{sq}$: Stores the indices $S_{k}$ of superquadrics associated with voxel $v$.
    \item $\mathbf{V}_f$: Given $v$, this tensor provides the starting index offset $f$ in $\mathbf{V}_{sq}$ from which to retrieve the first $S_{k}$.
    \item $\mathbf{V}_n$: Queried after $\mathbf{V}_f$. It specifies the number of entries, $n$, in $\mathbf{V}_{sq}$ corresponding to $v$, indicating how many superquadrics influence $v$.
\end{enumerate}

We then utilise $S_k$ to obtain the relevant superquadric parameters. Subsequently, we construct ray samples following standard volume rendering procedures. For each pixel ray $\mathbf{r}(t)=\mathbf{o}+t\mathbf{d}$, we sample $L$ uniformly spaced points $\mathbf{x}_{rj} = \mathbf{o} + t_{rj}\mathbf{d}$ along the unit direction $\mathbf{d}$ in 3D space, where $t_{rj}$ is uniformly sampled within the interval $[t_{\text{near}}, t_{\text{far}}]$ and represents the distance from the camera center along $\mathbf{r}$. Both the forward and backward passes are implemented in CUDA, with 256 rays per block.

\begin{table*}[t]
    \caption{\textbf{Semantic occupancy model comparison.} $^\dagger$ - RayIoU from released checkpoints; $^*$ - local training/evaluation. Loss denotes the loss space. Rep. denotes the scene representation: Gauss = Gaussian, Quad = Superquadric. FPS includes voxelisation. Mem. is the peak inference memory (GB), excluding rendering. @Xm is RayIoU at distance threshold X; Dyn. is dynamic-class RayIoU. All models are self-supervised. Best results between GaussianFlowOcc and \methodShort{} are in \textbf{bold}.}
  \centering
  \setlength{\tabcolsep}{3pt}
  \begin{adjustbox}{width=0.9\textwidth}
  \begin{tabular}{
    @{}l
    |cc
    |cc
    |cc
    |ccccc
    |ccccc
    @{}
}
    \toprule
    & & & & & & &
    \multicolumn{5}{c|}{\textbf{Occ3D-nuScenes \cite{tian2023occ3d}}} &
    \multicolumn{5}{c}{\textbf{OpenOccv2 \cite{tong2023scene}}} \\
    \textbf{Method} 
    & \textbf{Loss} 
    & \textbf{Rep.} 
    & \textbf{FPS} 
    & \textbf{Mem.} 
    & \textbf{IoU} 
    & \textbf{mIoU} 
    & \textbf{RayIoU} 
    & \textbf{@1m} 
    & \textbf{@2m} 
    & \textbf{@4m} 
    & \textbf{Dyn.}
    & \textbf{RayIoU} 
    & \textbf{@1m} 
    & \textbf{@2m} 
    & \textbf{@4m} 
    & \textbf{Dyn.} \\
    \midrule
    \rowcolor{black!9}
    QueryOcc \cite{lilja2026queryocc} & 3D & Query & 11.6 &  & 55.0 & 21.3 & 23.6 &  &  &  & 21.7 & & & & &\\
        \rowcolor{black!9}
    ShelfOcc \cite{boeder2026shelfocc} & 3D & Voxel &  &  & 56.1 & 22.9 & 22.6 & 16.3 & 22.7 & 28.9 & & & & & &\\
    SelfOcc$^\dagger$ \cite{huang2024selfocc} & 2D & Voxel & 7.4 & 3.0 & 45.0 & 10.5 & 10.9 & 7.6 & 10.9 & 14.1 & 7.2 & 9.7 & 6.2 & 9.6 & 13.1 & 6.8 \\
    OccNeRF$^\dagger$ \cite{zhang2023occnerf} & 2D & Voxel & 5.4 & 12.7 & 46.4 & 10.8 & 11.8 & 7.8 & 11.6 & 16.0 & 10.0 & 12.1 & 8.4 & 12.1 & 16.0 & 10.0 \\
    GaussianOcc$^\dagger$ \cite{gaussianocc} & 2D & Voxel & 5.4 & 11.8 & 42.9 & 11.3 & 13.4 & 9.8 & 13.5 & 16.9 & 11.0 & 12.5 & 9.1 & 12.6 & 15.8 & 10.3 \\
    GaussTR$^\dagger$ \cite{jiang2025gausstr} & 2D & Gauss & 0.3 & 2.0 & 44.5 & 13.8 & 14.6 & 10.6 & 14.7 & 18.5 & 17.8 & 14.9 & 10.9 & 15.1 & 18.8 & 17.5 \\
    QueryOcc \cite{lilja2026queryocc} & 2D & Query & 11.6 &  & 53.8 & 13.8 & 15.0 &  &  &  & 9.8 & & & & &\\ 
    EasyOcc \cite{hayes2025easyocc} & 2D & Voxel & 5.4 & & 38.9 & 15.7 & 16.5 & 12.3 & 16.7 & 20.6 & 14.3 & 16.2 & 12.1 & 16.4 & 20.0 & 13.7   \\
    TT-Occ \cite{zhang2025tt} & 2D & Gauss & 0.7 & 9.9 & & 16.7 & 15.2 & 11.3 & 14.6 & 17.8 & \\
    GaussianFlowOcc \cite{boeder2025gaussianflowocc} & 2D & Gauss & 10.2 & & 46.9 & 17.1 & 18.7 & 13.4 & 18.8 & 23.8 & \\
        \midrule
    GaussianFlowOcc$^*$ \cite{boeder2025gaussianflowocc} & 2D & Gauss & 9.6 & 2.9 & 40.5 & 16.7 & 18.3 & 12.6 & 18.3 & 24.0 & 17.0 & 18.2 & 12.9 & 18.4 & 23.4 & 17.0 \\
    \rowcolor{blue!10}
    \methodShort{} (Ours) & 2D & Quad & \textbf{21.5} & \textbf{0.7} & \textbf{43.5} & \textbf{17.1} & \textbf{19.6} & \textbf{13.5} & \textbf{19.6} & \textbf{25.6} & \textbf{17.6} & \textbf{19.9} & \textbf{14.2} & \textbf{20.1} & \textbf{25.4} & \textbf{17.6} \\
    \bottomrule
  \end{tabular}
  \end{adjustbox}
  \label{tab:sota_occ3d_no_class}
\end{table*}

\subsubsection{Forward Pass} 
For each ray $\mathbf{r}$, we traverse the ray samples $r_j$ in a front-to-back manner. We first determine voxel index $v_{rj}$ and query the tensors $\mathbf{V}_f$ and $\mathbf{V}_{n}$ to retrieve $n$ and $f$. We then use these to query $\mathbf{V}_{sq}$ to obtain $\mathcal{Q}(v_{rj})$, which contains all $S_{k}$ in the neighbourhood $\mathcal{N}_v$. All $r_j$ outside voxel bounds, as well as those for which $\mathbf{V}_{n}=0$, are skipped to avoid unnecessary computation. 

Next, 
for each $S_k \in \mathcal{Q}(v_{rj})$, we
transform $r_j$ into the local coordinate frame of $S_k$, and compute the occupancy probability $p_o(r_j, S_k)$ (Equation \ref{eq:sq_occupancy_prob}). The probabilities for each $S_k$ are summed to obtain the alpha value, $\alpha_{rj}$, which is upper-bounded by 1, and the semantic vector $\mathbf{s}_{rj}$ for $r_j$:

\begin{equation}
\label{eq:alpha_sem_sq}
\begin{aligned}
\alpha_{r_j} &=
\sum_{S_k \in \mathcal{Q}(v_{r_j})} p_o(r_j, S_k)\,\sigma_{S_k}, \\
\mathbf{s}_{r_j} &=
\sum_{S_k \in \mathcal{Q}(v_{r_j})}
p_o(r_j, S_k)\,\mathbf{c}_{S_k}.
\end{aligned}
\end{equation}

where $\mathbf{c}_{S_k}$ represents the semantic logits of $S_k$. After processing every ray sample in $\mathbf{r}$, the transmittance and corresponding weighting term are computed using front-to-back alpha-compositing  to account for occlusion effects:

\begin{equation}
\label{eq:alpha_composite}
w_{rj} = T_{rj} \alpha_{rj},
\quad
T_{rj} = \prod_{\ell=1}^{j-1} (1 - \alpha_{r\ell})
\end{equation}
Given $t_{rj}$, the ray sample distance along the viewing direction, the final depth and semantic output for each ray (pixel) are then given by:
\begin{equation}
\label{eq:depth_sem}
\hat{D}_p = \sum_{j=1}^{L} w_{rj} t_{rj},
\quad
\hat{S}_p = \sum_{j=1}^{L} w_{rj} \mathbf{s}_{rj}
\end{equation}

\subsubsection{Backward Pass}
For the backward pass, we divide the procedure into three separate stages that facilitate backpropagation of gradients to the superquadric parameters. We use a superquadric-major backward pass instead of a ray-major formulation, allowing interactions for each primitive to be grouped and reduced locally before writing gradients to global memory.

\textbf{Process 1.} First, we recompute the forward compositing quantities for $\mathbf{r}$ required for gradient propagation, including the transmittance $T$, alpha $\alpha$, and the semantic-gradient inner product $\left\langle \nabla_{\hat{\mathbf{S}}*r}\mathcal{L}, \mathbf{s}*{rj} \right\rangle$. Using these quantities, we analytically backpropagate through the front-to-back alpha compositing Equations~(\ref{eq:alpha_composite}) to obtain the opacity gradient $\frac{\partial \mathcal{L}}{\partial \alpha_{rj}}$. We store only the compact per-sample quantities needed by later stages, namely $w_{rj}=T_{rj}\alpha_{rj}$ and $\frac{\partial \mathcal{L}}{\partial \alpha_{rj}}$, rather than storing all forward-pass intermediates.

\textbf{Process 2.} We construct the set of valid sample--superquadric interactions $(S_k,r_j)$ using the spatial indexing structure. These interactions are sorted by superquadric index and grouped into contiguous ranges for each active primitive. This converts the backward pass from a ray-wise accumulation into a primitive-wise reduction problem.

\textbf{Process 3.}
For each grouped superquadric, we propagate the stored opacity gradients $\frac{\partial \mathcal{L}}{\partial \alpha_{rj}}$ through the occupancy probability formulation in Equation~(\ref{eq:sq_occupancy_prob}) to compute gradients for the parameters ${\mathbf{m}, \mathbf{s}, \mathbf{r}, \boldsymbol{\varepsilon}, \sigma, \mathbf{c}}$. Gradients are accumulated over all grouped interactions belonging to the same primitive using tiled reductions, followed by a final per-primitive reduction.

\subsection{Downstream Tasks}
\label{subsec:downstream}
\paragraph{Occupancy Forecasting.}
We utilize the temporal flow module to propagate the superquadric means at the current time, $S_M^T$, for future timesteps ($S_M^{T+1}$, $S_M^{T+2}$, $S_M^{T+3}$), and then voxelize. This requires no model fine-tuning.

\paragraph{Trajectory Estimation.}
For trajectory estimation, we follow ShelfGaussian \cite{zhao2026shelfgaussian}. We initialise learnable ego queries,
$\mathbf{Q}_{ego} \in \mathbb{R}^{N_{step} \times D}$ for $N_{step}$ future timesteps, and condition them on temporally propagated superquadric features.
Specifically, the temporal flow module propagates the superquadric means $S_M$ over $N_{step}$, producing
$\tilde{S}_M \in \mathbb{R}^{N_{step} \times 3}$. We then inject this temporal position information into the superquadric
features $S_F$ using a positional embedding:
The ego queries then attend to these temporally encoded features, and an MLP predicts the future ego trajectory
$\hat{\mathbf{X}} \in \mathbb{R}^{N_{step} \times 2}$:
\begin{equation}
    \hat{\mathbf{X}}
    =
    \mathrm{MLP}\!\left(
    \mathrm{CrossAttn}\!\left(
    \mathbf{Q}_{ego},
    \tilde{S}_F
    \right)
    \right)
\end{equation}

\section{Experiments}
\subsection{Datasets and Metrics}
\label{subsec:dataset_metrics}
For training, we utilise 2D pseudo-semantic labels \cite{ren2024grounded} and depth labels \cite{hu2024metric3d} provided by GaussianFlowOcc \cite{boeder2025gaussianflowocc}. These pseudo-labels are derived from nuScenes \cite{nuScenes} and hence permit model evaluation on nuScenes-derived datasets. 

To evaluate occupancy estimation, we use the Occ3D-nuScenes \cite{tian2023occ3d} and OpenOccv2 \cite{tong2023scene} datasets, both built upon the nuScenes dataset \cite{nuScenes}. These datasets provide dense occupancy annotations spanning $[-40\text{m}, 40\text{m}]$ along the $X$ and $Y$ axes and $[-1\text{m}, 5.4\text{m}]$ along the $Z$ axis, with a voxel resolution of $0.4\text{m}$, defined in the ego-vehicle coordinate frame. Evaluation is conducted using the \gls{iou}, \gls{miou}, and RayIoU metrics, with emphasis on RayIoU as it discourages over-dense predictions and favours geometrically faithful semantics over dense but inaccurate occupancy \cite{tang2024sparseocc}.

For downstream tasks, we evaluate occupancy forecasting using the \gls{iou}, \gls{miou}, and RayIoU metrics. Following ST-P3~\cite{hu2022st}, we report L2 displacement error and collision rate (CR) for trajectory estimation.

For depth estimation, we retain the same perception range as occupancy estimation for consistency. Ground-truth depth is obtained from nuScenes LiDAR data projected into the camera frame, sourced from GaussianOcc \cite{gaussianocc}. The depth values are clamped to the range $[0.1\text{m}, 80\text{m}]$ to ensure consistency with prior work. Evaluation follows standard depth metrics reported in the literature: Abs Rel, Sq Rel, RMSE, RMSE log, and threshold metrics ($\delta$).

\subsection{Implementation Details}
\label{subsec:main_impl}
We train \method{} following the configuration of GaussianFlowOcc \cite{boeder2025gaussianflowocc}. We employ $N=1600$ superquadrics and a ResNet-50 backbone \cite{he2016deep} pretrained on ImageNet \cite{deng2009imagenet} with an input image resolution of $256\times704$. The Gaussian Transformer network consists of $B=3$ sequential blocks, with the induced attention module using 500 inducing points. We set $D=256$. The temporal flow module operates over a temporal horizon of $T=6$.
For \render{}, we use a neighbourhood size of $\mathcal{N}_v\!=\!5$, $L\!=\!100$ ray samples, and a 
sampling range of $[t_{\text{near}}, t_{\text{far}}]=[0.1, 40.0]$, without early ray termination. 
%
%
All methods are trained for 18 epochs with a batch size of 1 on NVIDIA GPUs.
For fair comparison, we report memory usage, model inference time, and rendering time with $1\!\times\!\text{A100}$.
We train the trajectory estimator for 12 epochs, using the frozen \methodShort{} model, with $N_{step}=6$.


\subsection{Main Results}

\subsubsection{Semantic Occupancy}
Since self-supervised occupancy methods can obtain high voxel IoU through over-dense predictions, we treat RayIoU as the primary reconstruction metric. 
In \autoref{tab:sota_occ3d_no_class}, \methodShort{} improves RayIoU over GaussianFlowOcc on both Occ3D-nuScenes and OpenOccv2, while also outperforming it in IoU and mIoU. We achieve higher metrics for all RayIoU distance thresholds, indicating that superquadrics provide more faithful scene coverage. Further, we outperform for RayIoU on dynamic classes, namely bicycle, car, and pedestrian. Finally, \methodShort{} provides a 75\% reduction in memory and over 2$\times$ faster inference at 21.5 FPS, primarily due to the 84\% reduction in primitive count.



ShelfOcc~\cite{boeder2026shelfocc} and QueryOcc~\cite{lilja2026queryocc} use 3D losses, which substantially improve performance. Prior work shows that combining 2D rendering and 3D losses further boosts performance~\cite{hayes2025easyocc,chambon2025gaussrender} and would therefore likely also benefit our model. However, these 3D-labels are not open-source. In a fair comparison, QueryOcc trained with only a 2D rendering loss fails to outperform \methodShort{}, highlighting the effectiveness of our scene representation strategy.

Detailed class-wise mIoU and RayIoU results, along with additional NYUv2 \cite{silberman2012indoor} generalisation experiments, are provided in the supplementary material (B.12, B.1).

\subsubsection{Downstream Tasks}
For a fair comparison, we adapt GaussianFlowOcc~\cite{boeder2025gaussianflowocc} to the downstream setup described in \autoref{subsec:downstream}.

\paragraph{Occupancy Forecasting.}
In \autoref{tab:sota_forecasting}, our method outperforms GaussianFlowOcc for all metrics, notably for FPS and RayIoU, demonstrating the effectiveness of superquadrics for downstream tasks due to their ability to represent the scene with fewer primitives. Comparing both methods to their copy\&paste variants ($\dagger$), it is evident that the temporal flow module significantly aids in forecasting accuracy, with a small impact on inference speed due to propagation and additional voxelization.
Furthermore, \methodShort{} outperforms several strongly supervised methods, highlighting the effectiveness of sparse scene representation for occupancy forecasting.

\paragraph{Trajectory Estimation.}
In \autoref{tab:trajectory}, \methodShort{} outperforms GaussianFlowOcc, namely on the collision rate (CR) metric,  indicating stronger safety-relevant trajectory estimation. This suggests that the compact superquadric representation better preserves temporally useful free-space and object-extent structure than a larger set of local Gaussian primitives. Furthermore, our method rivals the performance of the similarly trained ShelfGaussian \cite{zhao2026shelfgaussian} despite it being supervised with LiDAR data.

\begin{table}[ht]
\centering
\setlength{\tabcolsep}{3pt}
\caption{\textbf{Downstream tasks.}
Sup.- supervision type.
FPS - runtime from camera input to all occupancy forecasts (1s, 2s, 3s) and trajectories, if applicable.
Metrics are averaged over 1s, 2s, and 3s forecasts; timewise results are in the supplementary (B.13).
Best results are in \textbf{bold}. $^\dagger$: copying the current occupancy as future estimates (no forecasting). $^*$: locally trained and evaluated.}

\label{tab:sota_occ_traj}

\begin{subtable}{\linewidth}
\centering
\begin{adjustbox}{width=\linewidth}
\begin{tabular}{l|c|c|cc|cc}
\toprule
&
&
&
&
&
\multicolumn{2}{c}{\textbf{RayIoU}} \\
\textbf{Method}
& \textbf{Sup.}
& \textbf{FPS}
& \textbf{IoU}
& \textbf{mIoU}
& \textbf{Occ3D}
& \textbf{OpenOccv2} \\
\midrule
\rowcolor{black!8}
OccWorld \cite{zheng2024occworld} & Strong & 2.8 & 16.5 & 8.6 & & \\

\rowcolor{black!8}
OccLLaMa \cite{wei2024occllama} & Strong & & 23.0 & 8.7 & & \\

\rowcolor{black!8}
Occ-LLM \cite{xu2025occ} & Strong & & 23.8 & 10.2 & & \\

\rowcolor{black!8}
DOME \cite{gu2024dome} & Strong & & 28.8 & 18.3 & & \\

\rowcolor{black!8}
COME \cite{shi2026come} & Strong & & 44.0 & 22.3 & & \\

\rowcolor{black!8}
SparseWorld-TC \cite{du2026sparseworld} & Strong & 2.1 & 53.5 & 29.9 & & \\

\midrule

\rowcolor{black!8}
GaussianFlowOcc$^\dagger$ \cite{boeder2025gaussianflowocc}
& Self & 9.6 & 33.6 & 9.3 & 10.3 & 9.8 \\

\rowcolor{black!8}
\methodShort{}$^\dagger$ (Ours)
& Self & 21.5 & 37.2 & 10.0 & 11.0 & 11.0 \\

\midrule

GaussianFlowOcc$^*$ \cite{boeder2025gaussianflowocc}
& Self & 7.9 & 37.7 & 13.8 & 13.7 & 13.9 \\

\rowcolor{blue!10}
\methodShort{} (Ours)
& Self & \textbf{17.9} & \textbf{39.3} & \textbf{14.1} & \textbf{14.4} & \textbf{14.6} \\
\bottomrule
\end{tabular}
\end{adjustbox}
\caption{\textbf{Occupancy forecasting.}}
\label{tab:sota_forecasting}
\end{subtable}

\vspace{0.5em}

\begin{subtable}{\linewidth}
\centering
\begin{adjustbox}{width=\linewidth}
\begin{tabular}{l|cc|c|cc}
\toprule
\textbf{Method}
& \textbf{Sup.}
& \textbf{Ego Traj.}
& \textbf{FPS}
& \textbf{L2 (m) $\downarrow$}
& \textbf{CR(\%) $\downarrow$} \\
\midrule

\rowcolor{black!8}
ST-P3 \cite{hu2022st}
& Strong & \xmark & & 2.65 & 4.23 \\

\rowcolor{black!8}
OccWorld \cite{zheng2024occworld}
& Strong & \xmark & 2.8 & 1.40 & 0.87 \\

\rowcolor{black!8}
OccLLaMa \cite{wei2024occllama}
& Strong & \xmark & & 1.20 & 0.70 \\

\rowcolor{black!8}
GaussianAD \cite{zheng2024gaussianad}
& Strong & \xmark & & 0.64 & 0.42 \\

\rowcolor{black!8}
Uni-AD \cite{hu2023planning}
& Strong & \checkmark & & 0.46 & 0.37 \\


\rowcolor{black!8}
BEV-Planner \cite{li2024ego}
& Self & \checkmark & & 0.35 & 0.42 \\

\midrule

\rowcolor{black!8}
ShelfGaussian \cite{zhao2026shelfgaussian}
& Self$^*$ & \xmark & & 1.77 & 2.67 \\

\rowcolor{black!8}
ShelfGaussian \cite{zhao2026shelfgaussian}
& Self$^*$ & \checkmark & & 0.35 & 0.32 \\

\midrule

GaussianFlowOcc$^*$ \cite{boeder2025gaussianflowocc}
& Self & \xmark & 7.9 & 2.41 & 2.69 \\

\rowcolor{blue!10}
SQOcc (Ours)
& Self & \xmark & \textbf{17.9} & \textbf{2.22} & \textbf{1.91} \\
\bottomrule
\end{tabular}
\end{adjustbox}
\caption{\textbf{Trajectory estimation.} Self$^*$ - LiDAR supervision. Ego Traj. - models that use ego trajectory. CR - collision rate}
\label{tab:trajectory}
\end{subtable}

\end{table}



\subsection{Rendering Method Study}

To benchmark our proposed \render{} approach, we compare it with alternative rendering methods. For Gaussianisation and Voxel NeRF rendering, we first voxelize the superquadric set $\mathcal{S}$ (supplementary A.4), with voxel bounds and resolution specified in \autoref{subsec:dataset_metrics}, producing a density voxel grid $\mathbf{V}_d$ and a semantic voxel grid $\mathbf{V}_s$.

\subsubsection{Baselines}
\paragraph{\textbf{Multi-Layer Gaussian Approximation (MLGA).}}
To appropriately model the superquadric occupancy distribution as Gaussian, we extend upon previous works \cite{gao2025self, jiang2024gaussianblock}. Given a canonical superquadric $S=(\boldsymbol{s}, \varepsilon_1, \varepsilon_2)$, we augment its scale parameters with $n$ positive scalars $P = \{\, p_1, p_2, \ldots, p_n \,\} \subset \mathbb{R}^+$ to define a scaled family of superquadrics: $S_P = \{\, (p\,\boldsymbol{s}, \varepsilon_1,\, \varepsilon_2) \mid p \in P \,\}$. 

For each $S_p \in S_P$, a tessellated unit icosphere is constructed. Each vertex on the icosphere, defined by spherical coordinates $(\eta, \omega)$, is mapped to the superquadric surface through signed powers of trigonometric functions, controlled by $\boldsymbol{\varepsilon}$. For each mapped vertex, we place a 3D Gaussian at the centre, for which the scales $s_x$ and $s_y$ provide sufficient coverage.

For a Gaussian $G_p$ with center $\mathbf{x}_{p}$ sampled on superquadric $S_{p}$, we define a corresponding Gaussian $\mathbf{x}_{p+1}$ on $S_{p+1}$ at the same angular coordinates $(\eta, \omega)$. For $G_p$, we define the scale $s_z$ and the opacity $\sigma$:
\[
s_z
= \frac{1}{2} \|\mathbf{x}_{p+1}(\eta, \omega) - \mathbf{x}_{p}(\eta, \omega)\|,
\quad
    \sigma = \sigma_{S_p} \cdot e^{f(\mathbf{m})}
\]
where $\mathbf{m}$ is the Gaussian mean. The scale $s_z$ enables each Gaussian to decay gradually along its normal, thereby better approximating the superquadric occupancy probability. 
Our formulation for opacity $\sigma$ ensures that the occupancy evaluated at the Gaussian’s centre exactly matches the corresponding value in the superquadric distribution. Gaussians are subsequently rendered using \gls{3dgs}.

\paragraph{\textbf{Gaussianisation.}} 
For this method, we follow the strategy adopted in prior state-of-the-art work \cite{gaussianocc, chambon2025gaussrender}. A spherical Gaussian primitive is instantiated for each voxel in the scene as $(\mathbf{m}, \mathbf{s}, \sigma, \mathbf{c})$ with: the mean $\mathbf{m}$ at voxel center, the scale hyperparameter $\mathbf{s}$ constant across all axes, the opacity $\sigma$ with value from $\mathbf{V}_d$, the semantics $\mathbf{c}$: value from $\mathbf{V}_s$. Gaussians are subsequently rendered using \gls{3dgs}.

\paragraph{\textbf{Voxel NeRF.}} 
We adopt the approach introduced in OccNeRF \cite{zhang2023occnerf} for volumetric rendering of voxel grids, using the render \Cref{eq:depth_sem}. In this formulation, alpha $\alpha_{rj}$ and semantic feature $\mathbf{s}_{rj}$ are obtained via trilinear interpolation of the voxel grids $\mathbf{V}_d$ and $\mathbf{V}_s$, respectively.

\paragraph{\textbf{Implementation details for baselines.}}
For MLGA, we use an 80-faced icosahedron, $S_{n=9}$ where $P=[0.5, 0.6, 0.75, 0.9, 1.05, 1.2, 1.6, 2.0, 2.5]$. Due to limitations of this strategy, particularly in retaining fidelity across multiple views, we omit the temporal module during training as it leads to severely degraded results.
For Gaussianisation, we set the Gaussian scale to $0.15$. 
For Voxel NeRF, we use $L\!=\!75$ ray samples with the same ray sampling range as \renderShort{}. For \renderShort{}, we use the implementation parameters stated in \autoref{subsec:main_impl}.

For depth evaluation, we render at resolution $396 \times 704$ and bilinearly interpolate to the nuScenes \cite{nuScenes} image resolution of $900\times1600$ in line with previous methods \cite{boeder2025gaussianflowocc}.

For MLGA, Gaussianisation, and \renderShort{}, we used $4\!\times\!\text{A100 (40GB)}$, and $4\!\times\!\text{H100 (80GB)}$ for Voxel NeRF due to memory constraints.


\begin{table}[ht]
    \centering
    \setlength{\tabcolsep}{3pt}
    \caption{\textbf{Superquadric rendering comparison.}
    All methods use the same superquadric scene representation and \method{} architecture, differing only in render proxy: MLGA, Gaussianisation, Voxel NeRF, or SQOcc-R.
    All models are self-supervised. Best results are in \textbf{bold}.}


\begin{subtable}{\linewidth}
\begin{adjustbox}{width=\textwidth}
\begin{tabular}{l|cc|cc|cc|c}
\toprule
&
&
&
\multicolumn{2}{c|}{\textbf{RayIoU}}
&
\multicolumn{2}{c|}{\textbf{Memory (GB)}}
&
\textbf{Render} \\
\textbf{Method}
& \textbf{IoU}
& \textbf{mIoU}
& \textbf{Occ3D }
& \textbf{OpenOccv2}
& \textbf{Training}
& \textbf{Inference}
& \textbf{Time (ms)} \\
\midrule
    MLGA & 33.7 & 12.7 & 12.1 & 13.1 &  & 1.35 & 44.0 \\
    Gaussianisation  & \textbf{43.7} & 17.0 & 16.4 & 16.6 & \textbf{6.2} & \textbf{0.8} & \textbf{13.9} \\
    Voxel NeRF & 42.9 & \textbf{17.1} & 18.8 & 18.9 & 69.4 & 11.1 & 102.4 \\
    \rowcolor{blue!10}
    \renderShort{} (Ours) & 43.5 & \textbf{17.1} & \textbf{19.6} & \textbf{19.9} & 13.7 & 2.8 & 17.2 \\
        \bottomrule
\end{tabular}
\end{adjustbox}
\caption{\textbf{Semantic occupancy.} Memory is peak usage across training and inference, including the rendering pipeline during inference. Render time is averaged over rendering six surround-view validation images.}
\label{tab:ablation_render_occ}
\end{subtable}

\begin{subtable}{\linewidth}
\centering
\begin{adjustbox}{width=\linewidth}
\begin{tabular}{l|l|cc|cc|ccc}
    \toprule
    & 
    & \textbf{Abs}
    & \textbf{Sq}
    & \multicolumn{2}{c|}{\textbf{RMSE} }
    & \multicolumn{3}{c}{$\boldsymbol{\delta}$}
    \\
    \textbf{Method}
    & \multirow{-1}{*}{Rep.}
    & \textbf{Rel} $\downarrow$
    & \textbf{Rel} $\downarrow$
    & $\downarrow$
    & \textbf{log} $\downarrow$
    & $\boldsymbol{<\!\!1.25} \uparrow$
    & $\boldsymbol{<\!\!1.25^2} \uparrow$
    & $\boldsymbol{<\!\!1.25^3} \uparrow$ \\
    \midrule
    \rowcolor{black!8}
    SurroundDepth~\cite{wei2023surrounddepth} & Image
    & 0.280 & 4.401 & 7.467 & 0.364 & 0.661 & 0.844 & 0.917 \\
    \rowcolor{black!8}
    SimpleOcc~\cite{gan2023simple} & Voxel
    & 0.224 & 3.383 & 7.165 & 0.333 & 0.753 & 0.877 & 0.930 \\
    \rowcolor{black!8}
    OccNeRF~\cite{zhang2023occnerf} & Voxel
    & 0.202 & 2.883 & 6.697 & 0.319 & 0.768 & 0.882 & 0.931 \\
    \rowcolor{black!8}
    SelfOcc~\cite{huang2024selfocc} & Voxel
    & 0.215 & 2.743 & 6.706 & 0.316 & 0.753 & 0.875 & 0.932 \\
    \rowcolor{black!8}
    GaussianOcc~\cite{gaussianocc}  & Voxel
    & 0.197 & 1.846 & 6.733 & 0.312 & 0.746 & 0.873 & 0.931 \\
    \rowcolor{black!8}
    GaussianFlowOcc \cite{boeder2025gaussianflowocc} & Gauss
    & 0.278 & 2.522 & 5.232 & 0.389 & 0.677 & 0.826 & 0.898 \\
    \midrule
    MLGA & Quad & 
    0.194 & 1.934 & 8.236 & 0.569 & 0.714 & 0.854 & 0.914 \\
    Gaussianisation & Quad &
    0.203 & 1.880 & \textbf{8.071} & \textbf{0.351} & 0.711 & 0.852 & 0.919 \\
    Voxel NeRF & Quad & 
    0.208 & 2.023 & 8.383 & 0.387 & 0.683 & 0.822 & 0.894 \\
    \rowcolor{blue!10}
    \renderShort{} (Ours) & Quad &
    \textbf{0.190} & \textbf{1.825} & 8.163 & 0.353 & \textbf{0.717} & \textbf{0.857} & \textbf{0.920} \\
    \bottomrule
\end{tabular}
\end{adjustbox}
\caption{\textbf{Depth estimation.} Rep. denotes scene representation: Gauss - Gaussian; Quad - Superquadric.} 
\label{tab:ablation_render_depth}
\end{subtable}
\end{table}

\begin{table*}[ht]
\centering
\scriptsize
\setlength{\tabcolsep}{2.0pt}
\renewcommand{\arraystretch}{1.05}
\caption{
\textbf{Ablation study} for \render{}. The parameter highlighted in grey is the preferred value.
}

\begin{subtable}[t]{0.245\textwidth}
\centering
\caption{Neighbourhood size}
\resizebox{\linewidth}{!}{%
\begin{tabular}{c|cc|cc|c}
\toprule
&
&
&
\multicolumn{2}{c|}{\textbf{RayIoU}}
&
\textbf{Lat.} \\
$\boldsymbol{\mathcal{N}_v}$
& \textbf{IoU}
& \textbf{mIoU}
& \textbf{Occ3D}
& \textbf{OpenOccv2}
& \textbf{(ms)} \\
\midrule
3 & 38.9 & 14.5 & 14.2 & 15.2 & \textbf{13.5} \\
\rowcolor{black!5}
5 & \textbf{41.1} & 15.8 & 17.2 & 17.9 & 17.2 \\
\rowcolor{white}
7 & 40.7 & \textbf{15.9} & \textbf{18.1} & \textbf{18.4} & 29.7 \\
\bottomrule
\end{tabular}}
\end{subtable}
\hfill
\begin{subtable}[t]{0.245\textwidth}
\centering
\caption{Ray termination}
\resizebox{\linewidth}{!}{%
\begin{tabular}{c|cc|cc|c}
\toprule
&
&
&
\multicolumn{2}{c|}{\textbf{RayIoU}}
&
\textbf{Lat.} \\
$\boldsymbol{\kappa}$
& \textbf{IoU}
& \textbf{mIoU}
& \textbf{Occ3D}
& \textbf{OpenOccv2}
& \textbf{(ms)} \\
\midrule
$1e^{-2}$ & 39.5 & 15.2 & 16.7 & 17.4 & \textbf{15.4} \\
$1e^{-4}$ & \textbf{41.2} & 15.7 & \textbf{17.4} & \textbf{18.1} & 16.9 \\
\rowcolor{black!5}
None & 41.1 & \textbf{15.8} & 17.2 & 17.9 & 17.2 \\
\bottomrule
\end{tabular}}
\end{subtable}
\hfill
\begin{subtable}[t]{0.245\textwidth}
\centering
\caption{Number of superquadrics}
\resizebox{\linewidth}{!}{%
\begin{tabular}{c|cc|cc|c}
\toprule
&
&
&
\multicolumn{2}{c|}{\textbf{RayIoU}}
&
\textbf{Lat.} \\
$\boldsymbol{N}$
& \textbf{IoU}
& \textbf{mIoU}
& \textbf{Occ3D}
& \textbf{OpenOccv2}
& \textbf{(ms)} \\
\midrule
1,200  & 39.7 & 15.0 & 16.3 & 17.0 & \textbf{16.3} \\
\rowcolor{black!5}
1,600  & 41.1 & 15.8 & 17.2 & 17.9 & 17.2 \\
\rowcolor{white}
2,000  & 41.4 & \textbf{16.1} & \textbf{17.7} & \textbf{18.3} & 20.2 \\
4,000  & 40.2 & 15.9 & 17.5 & 18.0 & 34.8 \\
10,000 & \textbf{42.0} & 15.9 & 16.4 & 17.0 & 91.2 \\
\bottomrule
\end{tabular}}
\end{subtable}
\hfill
\begin{subtable}[t]{0.245\textwidth}
\centering
\caption{Number of ray samples}
\resizebox{\linewidth}{!}{%
\begin{tabular}{c|cc|cc|c}
\toprule
&
&
&
\multicolumn{2}{c|}{\textbf{RayIoU}}
&
\textbf{Lat.} \\
$\boldsymbol{L}$
& \textbf{IoU}
& \textbf{mIoU}
& \textbf{Occ3D}
& \textbf{OpenOccv2}
& \textbf{(ms)} \\
\midrule
50  & 40.6 & 15.6 & 17.3 & 17.9 & \textbf{12.7} \\
75  & 40.9 & 15.5 & 16.9 & 17.7 & 15.3 \\
\rowcolor{black!5}
100 & 41.1 & 15.8 & 17.2 & 17.9 & 17.2 \\
\rowcolor{white}
125 & 41.1 & 15.8 & 17.1 & 17.9 & 19.8 \\
150 & \textbf{41.5} & \textbf{15.9} & \textbf{17.5} & \textbf{18.1} & 21.1 \\
\bottomrule
\end{tabular}}
\end{subtable}

\label{tab:main_abl}
\end{table*}

\subsubsection{Quantitative Evaluation}
\paragraph{\textbf{3D Occupancy.}}
In \autoref{tab:ablation_render_occ}, \renderShort{} outperforms all other methods in RayIoU, namely by 19.5\% on OpenOccv2 compared to Gaussianisation. Methods with the temporal flow module exhibit similar IoU and mIoU results, due to lack of penalisation for overly dense estimations. For MLGA, we disable the temporal flow module as it severely degraded performance. This can be attributed to the mismatch in decay between Gaussians (quadratic) and superquadrics (not necessarily quadratic), indicating that a Gaussian approximation is not an appropriate render proxy.

\paragraph{\textbf{Rendering.}}

As shown in \autoref{tab:ablation_render_occ}, \renderShort{} reduces memory by 92.9\% during training and 74.7\% during inference relative to Voxel NeRF, thanks to efficient spatial indexing, while reducing render time by 83.2\% and achieving 58 FPS (17.2 ms) across the six nuScenes cameras. MLGA is slower than \renderShort{} and Gaussianisation due to rendering over 1M Gaussians. \renderShort{} provides the best fidelity-efficiency trade-off among renderers: it achieves the highest RayIoU while remaining real-time, whereas Gaussianisation is faster but loses substantial RayIoU, suggesting that it does not preserve superquadric fidelity as effectively.



\paragraph{\textbf{Depth Estimation.}}
For superquadric-based depth estimation in \autoref{tab:ablation_render_depth}, \renderShort{} outperforms the other approaches (MLGA, Gaussianisation, Voxel NeRF) for three important depth metrics (Abs Rel, Sq Rel, $\delta$) \cite{eigen2014depth,khan2020deep}. The overall improvements can be attributed to the direct rendering of superquadrics, which enables more accurate geometric reconstruction compared to the voxel-based proxy rendering employed by Gaussianisation and Voxel NeRF.


\renderShort{} remains competitive with state-of-the-art methods, outperforming all baselines on Abs Rel and Sq Rel. Its higher RMSE indicates a few large depth errors, likely in distant regions, as limiting the depth range to $40m$ reduces RMSE from 8.163 to 4.444. Methods using photometric consistency losses improve threshold-based depth metrics \cite{zhang2023occnerf, gaussianocc, huang2024selfocc, gan2023simple}, but substantially underperform on occupancy estimation.



\subsection{Ablation Studies}
\label{subsec:ablations}
For the ablation study on \renderShort{} in \autoref{tab:main_abl}, 
we train on the first half of the dataset (300 scenes) to reduce computational cost, and evaluate on the full validation set.

\paragraph{Neighbourhood Value \texorpdfstring{$\mathcal{N}_v$}{Nv}.}
We ablate $\mathcal{N}_v$, which defines the cubic region around each superquadric within which ray samples are permitted to query it. Increasing $\mathcal{N}_v$ improves performance, as each ray sample can consider a larger set of superquadrics, thereby enhancing rendering fidelity. We select $\mathcal{N}_v = 5$ to balance rendering fidelity, computational efficiency, and rendering time.

\paragraph{Ray Termination $\kappa$.}
Early ray termination has been shown to accelerate rendering in prior work \cite{li2022nerfacc, muller2022instant}. Sampling along a ray stops once the accumulated transmittance $T$ falls below a threshold $\kappa$. We observe that for $\kappa=1e^{-2}$, performance degrades significantly, while for $\kappa=1e^{-4}$ we see reduced latency with noisy performance results. Thus, we choose not to apply early ray termination.

\paragraph{SuperQuadric Count $N$.}
We observe that at approximately $N=2,000$ superquadrics, the model reaches a saturation point.
Increasing $N$ further provides minimal improvements in performance metrics while increasing render time.  We find that render time increases linearly with increasing $N$ count, and at $N=10,000$ the latency reaches 91.6 ms with degraded RayIoU due to overly dense scene coverage. Thus, we select $N=1,600$ to balance rendering cost and model inference speed.

\paragraph{Ray Sample Number $L$.}
Increasing the number of ray samples $L$ improves performance metrics by enhancing rendering fidelity. Performance gains begin to saturate at approximately $L=100$. While $L=100$ yields the strongest overall results, using $L=50$ remains competitive, with similar metrics and rendering time reduced by 26.2\% compared to $L=100$, making it a practical alternative depending on computational constraints.

\section{Conclusion}
In this paper, we present a real-time self-supervised occupancy model, \method{}, which uses superquadrics for scene representation to achieve state-of-the-art RayIoU on the Occ3D-nuScenes dataset for scene reconstruction and important downstream tasks, including occupancy forecasting and trajectory estimation. To address the rendering challenge, we introduce a CUDA-accelerated superquadric renderer, \render{}, that enables real-time rendering of thousands of superquadric primitives. Its spatial indexing strategy allows efficient querying of relevant superquadrics for each ray sample within the volume rendering framework, reducing rendering time while preserving fidelity, a limitation of alternative rendering approaches.
Compared to other candidate superquadric rendering methods, \render{} demonstrates superior results, achieving rendering speeds comparable to Gaussian-based methods.
Overall, our results demonstrate the effectiveness of superquadric scene representations. They further show that efficient superquadric rendering is practical and can approach the performance of Gaussian rasterisation, encouraging further research into superquadric scene representations.
 

\section*{Acknowledgement}
This publication has emanated from research supported in part by a grant from Taighde Éireann – Research Ireland under Grant number 18/CRT/6049. For the purpose of Open Access, the author has applied a CC BY public copyright licence to any Author Accepted Manuscript version arising from this submission

{
    \small
    \bibliographystyle{ieeenat_fullname}
    \bibliography{main}

@String(PAMI = {IEEE Trans. Pattern Anal. Mach. Intell.})

@String(CVPR= {IEEE Conf. Comput. Vis. Pattern Recog.})

@String(ICCV= {Int. Conf. Comput. Vis.})

@String(ECCV= {Eur. Conf. Comput. Vis.})

@String(TOG= {ACM Trans. Graph.})

@String(PAMI  = {IEEE TPAMI})

@String(CVPR  = {CVPR})

@String(ICCV  = {ICCV})

@String(ECCV  = {ECCV})

@String(TOG   = {ACM TOG})

@inproceedings{huang2024gaussianformer,
  title={{Gaussianformer: Scene as gaussians for vision-based 3d semantic occupancy prediction}},
  author={Huang, Yuanhui and Zheng, Wenzhao and Zhang, Yunpeng and Zhou, Jie and Lu, Jiwen},
  booktitle={ECCV},
  year={2024},
}

@inproceedings{gaussianocc,
  title={Gaussianocc: Fully self-supervised and efficient 3d occupancy estimation with gaussian splatting},
  author={Gan, Wanshui and Liu, Fang and Xu, Hongbin and Mo, Ningkai and Yokoya, Naoto},
  booktitle={ICCV},
  year={2025}
}

@inproceedings{huang2025gaussianformer2,
  title={{Gaussianformer-2: Probabilistic gaussian superposition for efficient 3d occupancy prediction}},
  author={Huang, Yuanhui and Thammatadatrakoon, Amonnut and Zheng, Wenzhao and Zhang, Yunpeng and Du, Dalong and Lu, Jiwen},
  booktitle={CVPR},
  year={2025}
}

@inproceedings{zuo2025gaussianworld,
  title={Gaussianworld: Gaussian world model for streaming 3d occupancy prediction},
  author={Zuo, Sicheng and Zheng, Wenzhao and Huang, Yuanhui and Zhou, Jie and Lu, Jiwen},
  booktitle={CVPR},
  year={2025}
}

@inproceedings{jiang2025gausstr,
  title={Gausstr: Foundation model-aligned gaussian transformer for self-supervised 3d spatial understanding},
  author={Jiang, Haoyi and Liu, Liu and Cheng, Tianheng and Wang, Xinjie and Lin, Tianwei and Su, Zhizhong and Liu, Wenyu and Wang, Xinggang},
  booktitle={CVPR},
  year={2025}
}

@inproceedings{chambon2025gaussrender,
  title={Gaussrender: Learning 3d occupancy with gaussian rendering},
  author={Chambon, Loick and Zablocki, Eloi and Boulch, Alexandre and Chen, Mickael and Cord, Matthieu},
  booktitle={ICCV},
  year={2025}
}

@inproceedings{boeder2025gaussianflowocc,
  title={Gaussianflowocc: Sparse and weakly supervised occupancy estimation using gaussian splatting and temporal flow},
  author={Boeder, Simon and Gigengack, Fabian and Risse, Benjamin},
  booktitle={ICCV},
  year={2025}
}

@article{zhang2025tt,
  title={TT-Occ: Test-Time Compute for Self-Supervised Occupancy via Spatio-Temporal Gaussian Splatting},
  author={Zhang, Fengyi and Yang, Huitong and Zhang, Zheng and Huang, Zi and Luo, Yadan},
  journal={arXiv preprint arXiv:2503.08485},
  year={2025}
}

@article{zhao2025gaussianformer3d,
  title={{GaussianFormer3D: Multi-Modal Gaussian-based Semantic Occupancy Prediction with 3D Deformable Attention}},
  author={Zhao, Lingjun and Wei, Sizhe and Hays, James and Gan, Lu},
  journal={arXiv preprint arXiv:2505.10685},
  year={2025}
}

@article{tian2023occ3d,
  title={Occ3d: A large-scale 3d occupancy prediction benchmark for autonomous driving},
  author={Tian, Xiaoyu and Jiang, Tao and Yun, Longfei and Mao, Yucheng and Yang, Huitong and Wang, Yue and Wang, Yilun and Zhao, Hang},
  journal={NeurIPS},
  year={2023}
}

@inproceedings{wei2023surroundocc,
  title={Surroundocc: Multi-camera 3d occupancy prediction for autonomous driving},
  author={Wei, Yi and Zhao, Linqing and Zheng, Wenzhao and Zhu, Zheng and Zhou, Jie and Lu, Jiwen},
  booktitle={ICCV},
  year={2023}
}

@inproceedings{behley2019semantickitti,
  title={Semantickitti: A dataset for semantic scene understanding of lidar sequences},
  author={Behley, Jens and Garbade, Martin and Milioto, Andres and Quenzel, Jan and Behnke, Sven and Stachniss, Cyrill and Gall, Jurgen},
  booktitle={ICCV},
  year={2019}
}

@inproceedings{wang2023openoccupancy,
  title={Openoccupancy: A large scale benchmark for surrounding semantic occupancy perception},
  author={Wang, Xiaofeng and Zhu, Zheng and Xu, Wenbo and Zhang, Yunpeng and Wei, Yi and Chi, Xu and Ye, Yun and Du, Dalong and Lu, Jiwen and Wang, Xingang},
  booktitle={ICCV},
  year={2023},
}

@inproceedings{nuScenes,
  author    = {Caesar, H. and Bankiti, V. and Lang, A. H. and Vora, S. and Liong, V. E. and Xu, Q. and Krishnan, A. and Pan, Y. and Baldan, G. and Beijbom, O.},
  title     = {nuScenes: A Multimodal Dataset for Autonomous Driving},
  booktitle = {CVPR},
  year      = {2020}
}

@inproceedings{sun2020waymo,
  title={Scalability in perception for autonomous driving: Waymo open dataset},
  author={Sun, Pei and Kretzschmar, Henrik and Dotiwalla, Xerxes and Chouard, Aurelien and Patnaik, Vijaysai and Tsui, Paul and Guo, James and Zhou, Yin and Chai, Yuning and Caine, Benjamin and others},
  booktitle={CVPR},
  year={2020}
}

@inproceedings{huang2023tri,
  title={Tri-perspective view for vision-based 3d semantic occupancy prediction},
  author={Huang, Yuanhui and Zheng, Wenzhao and Zhang, Yunpeng and Zhou, Jie and Lu, Jiwen},
  booktitle={CVPR},
  year={2023}
}

@article{li2023fb,
  title={Fb-occ: 3d occupancy prediction based on forward-backward view transformation},
  author={Li, Zhiqi and Yu, Zhiding and Austin, David and Fang, Mingsheng and Lan, Shiyi and Kautz, Jan and Alvarez, Jose M},
  journal={arXiv preprint arXiv:2307.01492},
  year={2023}
}

@article{chen2025survey3dgaussiansplatting,
      title={A Survey on 3D Gaussian Splatting}, 
      author={Guikun Chen and Wenguan Wang},
      year={2025},
      journal={arXiv preprint arXiv:2401.03890},
}

@article{zuo2023pointocc,
  title={Pointocc: Cylindrical tri-perspective view for point-based 3d semantic occupancy prediction},
  author={Zuo, Sicheng and Zheng, Wenzhao and Huang, Yuanhui and Zhou, Jie and Lu, Jiwen},
  journal={arXiv preprint arXiv:2308.16896},
  year={2023}
}

@inproceedings{cao2022monoscene,
  title={Monoscene: Monocular 3d semantic scene completion},
  author={Cao, Anh-Quan and De Charette, Raoul},
  booktitle={CVPR},
  year={2022}
}

@inproceedings{ma2024cotr,
  title={Cotr: Compact occupancy transformer for vision-based 3d occupancy prediction},
  author={Ma, Qihang and Tan, Xin and Qu, Yanyun and Ma, Lizhuang and Zhang, Zhizhong and Xie, Yuan},
  booktitle={CVPR},
  year={2024}
}

@inproceedings{hou2024fastocc,
  title={Fastocc: Accelerating 3d occupancy prediction by fusing the 2d bird’s-eye view and perspective view},
  author={Hou, Jiawei and Li, Xiaoyan and Guan, Wenhao and Zhang, Gang and Feng, Di and Du, Yuheng and Xue, Xiangyang and Pu, Jian},
  booktitle={ICRA},
  year={2024},
}

@inproceedings{kerbl20233dgs,
  title={{3D Gaussian splatting for real-time radiance field rendering.}},
  author={Kerbl, Bernhard and Kopanas, Georgios and Leimk{\"u}hler, Thomas and Drettakis, George},
  booktitle={ACM Trans. Graph.},
    year={2023}
}

@inproceedings{tang2024sparseocc,
  title={Sparseocc: Rethinking sparse latent representation for vision-based semantic occupancy prediction},
  author={Tang, Pin and Wang, Zhongdao and Wang, Guoqing and Zheng, Jilai and Ren, Xiangxuan and Feng, Bailan and Ma, Chao},
  booktitle={CVPR},
  year={2024}
}

@inproceedings{zhang2023simple,
  title={A simple framework for open-vocabulary segmentation and detection},
  author={Zhang, Hao and Li, Feng and Zou, Xueyan and Liu, Shilong and Li, Chunyuan and Yang, Jianwei and Zhang, Lei},
  booktitle={ICCV},
  year={2023}
}

@inproceedings{lee2019set,
  title={Set transformer: A framework for attention-based permutation-invariant neural networks},
  author={Lee, Juho and Lee, Yoonho and Kim, Jungtaek and Kosiorek, Adam and Choi, Seungjin and Teh, Yee Whye},
  booktitle={ICML},
  year={2019},
}

@inproceedings{he2016deep,
  title={Deep residual learning for image recognition},
  author={He, Kaiming and Zhang, Xiangyu and Ren, Shaoqing and Sun, Jian},
  booktitle={CVPR},
  year={2016}
}

@article{hu2024metric3d,
  title={Metric3d v2: A versatile monocular geometric foundation model for zero-shot metric depth and surface normal estimation},
  author={Hu, Mu and Yin, Wei and Zhang, Chi and Cai, Zhipeng and Long, Xiaoxiao and Chen, Hao and Wang, Kaixuan and Yu, Gang and Shen, Chunhua and Shen, Shaojie},
  journal={PAMI},
  year={2024},
}

@article{zhang2023occnerf,
  title={Occnerf: Self-supervised multi-camera occupancy prediction with neural radiance fields},
  author={Zhang, Chubin and Yan, Juncheng and Wei, Yi and Li, Jiaxin and Liu, Li and Tang, Yansong and Duan, Yueqi and Lu, Jiwen},
  journal={CoRR},
  year={2023}
}

@article{ren2024grounded,
  title={Grounded sam: Assembling open-world models for diverse visual tasks},
  author={Ren, Tianhe and Liu, Shilong and Zeng, Ailing and Lin, Jing and Li, Kunchang and Cao, He and Chen, Jiayu and Huang, Xinyu and Chen, Yukang and Yan, Feng and others},
  journal={arXiv preprint arXiv:2401.14159},
  year={2024}
}

@inproceedings{barsellotti2025talking,
  title={Talking to dino: Bridging self-supervised vision backbones with language for open-vocabulary segmentation},
  author={Barsellotti, Luca and Bianchi, Lorenzo and Messina, Nicola and Carrara, Fabio and Cornia, Marcella and Baraldi, Lorenzo and Falchi, Fabrizio and Cucchiara, Rita},
  booktitle={ICCV},
  year={2025}
}

@inproceedings{radford2021learning,
  title={Learning transferable visual models from natural language supervision},
  author={Radford, Alec and Kim, Jong Wook and Hallacy, Chris and Ramesh, Aditya and Goh, Gabriel and Agarwal, Sandhini and Sastry, Girish and Askell, Amanda and Mishkin, Pamela and Clark, Jack and others},
  booktitle={ICML},
  year={2021},
}

@article{zuo2025quadricformer,
  title={QuadricFormer: Scene as Superquadrics for 3D Semantic Occupancy Prediction},
  author={Zuo, Sicheng and Zheng, Wenzhao and Han, Xiaoyong and Yang, Longchao and Pan, Yong and Lu, Jiwen},
  journal={arXiv preprint arXiv:2506.10977},
  year={2025}
}

@inproceedings{huang2024selfocc,
  title={Selfocc: Self-supervised vision-based 3d occupancy prediction},
  author={Huang, Yuanhui and Zheng, Wenzhao and Zhang, Borui and Zhou, Jie and Lu, Jiwen},
  booktitle={CVPR},
  year={2024}
}

@article{zheng2024gaussianad,
  title={Gaussianad: Gaussian-centric end-to-end autonomous driving},
  author={Zheng, Wenzhao and Wu, Junjie and Zheng, Yao and Zuo, Sicheng and Xie, Zixun and Yang, Longchao and Pan, Yong and Hao, Zhihui and Jia, Peng and Lang, Xianpeng and others},
  journal={arXiv preprint arXiv:2412.10371},
  year={2024}
}

@article{wang2024distillnerf,
  title={Distillnerf: Perceiving 3d scenes from single-glance images by distilling neural fields and foundation model features},
  author={Wang, Letian and Kim, Seung Wook and Yang, Jiawei and Yu, Cunjun and Ivanovic, Boris and Waslander, Steven and Wang, Yue and Fidler, Sanja and Pavone, Marco and Karkus, Peter},
  journal={NeurIPS},
  year={2024}
}

@article{wang2024opus,
  title={Opus: occupancy prediction using a sparse set},
  author={Wang, Jiabao and Liu, Zhaojiang and Meng, Qiang and Yan, Liujiang and Wang, Ke and Yang, Jie and Liu, Wei and Hou, Qibin and Cheng, Ming-Ming},
  journal={NeurIPS},
  year={2024}
}

@inproceedings{gao2025self,
  title={{Self-supervised Learning of Hybrid Part-aware 3D Representations of 2D Gaussians and Superquadrics}},
  author={Gao, Zhirui and Yi, Renjiao and Huang, Yuhang and Chen, Wei and Zhu, Chenyang and Xu, Kai},
  booktitle={ICCV},
  year={2025}
}

@article{jiang2024gaussianblock,
  title={GaussianBlock: Building Part-Aware Compositional and Editable 3D Scene by Primitives and Gaussians},
  author={Jiang, Shuyi and Zhao, Qihao and Rahmani, Hossein and Soh, De Wen and Liu, Jun and Zhao, Na},
  journal={arXiv preprint arXiv:2410.01535},
  year={2024}
}

@article{barr1981superquadrics,
  title={Superquadrics and angle-preserving transformations},
  author={Barr, Alan H},
  journal={IEEE Computer graphics and Applications},
  year={1981},
  publisher={IEEE Computer Society}
}

@inproceedings{guedon2024sugar,
  title={Sugar: Surface-aligned gaussian splatting for efficient 3d mesh reconstruction and high-quality mesh rendering},
  author={Gu{\'e}don, Antoine and Lepetit, Vincent},
  booktitle={CVPR},
  year={2024}
}

@inproceedings{tong2023scene,
  title={Scene as occupancy},
  author={Tong, Wenwen and Sima, Chonghao and Wang, Tai and Chen, Li and Wu, Silei and Deng, Hanming and Gu, Yi and Lu, Lewei and Luo, Ping and Lin, Dahua and others},
  booktitle={ICCV},
  year={2023}
}

@article{waczynska2024games,
  title={Games: Mesh-based adapting and modification of gaussian splatting},
  author={Waczy{\'n}ska, Joanna and Borycki, Piotr and Tadeja, S{\l}awomir and Tabor, Jacek and Spurek, Przemys{\l}aw},
  journal={arXiv preprint arXiv:2402.01459},
  year={2024}
}

@inproceedings{alaniz2023iterative,
  title={Iterative superquadric recomposition of 3d objects from multiple views},
  author={Alaniz, Stephan and Mancini, Massimiliano and Akata, Zeynep},
  booktitle={ICCV},
  year={2023}
}

@article{gan2023simple,
  title={A simple attempt for 3d occupancy estimation in autonomous driving},
  author={Gan, Wanshui and Mo, Ningkai and Xu, Hongbin and Yokoya, Naoto},
  journal={CoRR},
  year={2023}
}

@inproceedings{wei2023surrounddepth,
  title={Surrounddepth: Entangling surrounding views for self-supervised multi-camera depth estimation},
  author={Wei, Yi and Zhao, Linqing and Zheng, Wenzhao and Zhu, Zheng and Rao, Yongming and Huang, Guan and Lu, Jiwen and Zhou, Jie},
  booktitle={CoRL},
  year={2023},
}

@article{yu2026superocc,
  title={SuperOcc: Toward Cohesive Temporal Modeling for Superquadric-based Occupancy Prediction},
  author={Yu, Zichen and Liu, Quanli and Wang, Wei and Zhang, Liyong and Zhao, Xiaoguang},
  journal={arXiv preprint arXiv:2601.15644},
  year={2026}
}

@article{xiao2025neural,
  title={Neural Radiance Fields for the Real World: A Survey},
  author={Xiao, Wenhui and Chierchia, Remi and Cruz, Rodrigo Santa and Li, Xuesong and Ahmedt-Aristizabal, David and Salvado, Olivier and Fookes, Clinton and Lebrat, Leo},
  journal={arXiv preprint arXiv:2501.13104},
  year={2025}
}

@article{gao2022nerf,
  title={Nerf: Neural radiance field in 3d vision, a comprehensive review},
  author={Gao, Kyle and Gao, Yina and He, Hongjie and Lu, Dening and Xu, Linlin and Li, Jonathan},
  journal={arXiv preprint arXiv:2210.00379},
  year={2022}
}

@inproceedings{wang2022neural,
  title={Neural rendering for stereo 3d reconstruction of deformable tissues in robotic surgery},
  author={Wang, Yuehao and Long, Yonghao and Fan, Siu Hin and Dou, Qi},
  booktitle={MICCAI},
  year={2022},
}

@inproceedings{batlle2023lightneus,
  title={Lightneus: Neural surface reconstruction in endoscopy using illumination decline},
  author={Batlle, V{\'\i}ctor M and Montiel, Jos{\'e} MM and Fua, Pascal and Tard{\'o}s, Juan D},
  booktitle={MICCAI},
  year={2023},
}

@inproceedings{chen2023cunerf,
  title={Cunerf: Cube-based neural radiance field for zero-shot medical image arbitrary-scale super resolution},
  author={Chen, Zixuan and Yang, Lingxiao and Lai, Jian-Huang and Xie, Xiaohua},
  booktitle={ICCV},
  year={2023}
}

@inproceedings{rosinol2023nerf,
  title={Nerf-slam: Real-time dense monocular slam with neural radiance fields},
  author={Rosinol, Antoni and Leonard, John J and Carlone, Luca},
  booktitle={IROS},
  year={2023},
}

@inproceedings{liso2024loopy,
  title={Loopy-slam: Dense neural slam with loop closures},
  author={Liso, Lorenzo and Sandstr{\"o}m, Erik and Yugay, Vladimir and Van Gool, Luc and Oswald, Martin R},
  booktitle={CVPR},
  year={2024}
}

@article{li2023dense,
  title={Dense rgb slam with neural implicit maps},
  author={Li, Heng and Gu, Xiaodong and Yuan, Weihao and Yang, Luwei and Dong, Zilong and Tan, Ping},
  journal={arXiv preprint arXiv:2301.08930},
  year={2023}
}

@article{li2022nerfacc,
  title={Nerfacc: A general nerf acceleration toolbox},
  author={Li, Ruilong and Tancik, Matthew and Kanazawa, Angjoo},
  journal={arXiv preprint arXiv:2210.04847},
  year={2022}
}

@inproceedings{azinovic2022neural,
  title={Neural rgb-d surface reconstruction},
  author={Azinovi{\'c}, Dejan and Martin-Brualla, Ricardo and Goldman, Dan B and Nie{\ss}ner, Matthias and Thies, Justus},
  booktitle={CVPR},
  year={2022}
}

@article{wang2021neus,
  title={Neus: Learning neural implicit surfaces by volume rendering for multi-view reconstruction},
  author={Wang, Peng and Liu, Lingjie and Liu, Yuan and Theobalt, Christian and Komura, Taku and Wang, Wenping},
  journal={arXiv preprint arXiv:2106.10689},
  year={2021}
}

@inproceedings{oechsle2021unisurf,
  title={Unisurf: Unifying neural implicit surfaces and radiance fields for multi-view reconstruction},
  author={Oechsle, Michael and Peng, Songyou and Geiger, Andreas},
  booktitle={ICCV},
  year={2021}
}

@article{hayes2025easyocc,
  title={{EasyOcc: 3D Pseudo-Label Supervision for Fully Self-Supervised Semantic Occupancy Prediction Models}},
  author={Hayes, Seamie and Sistu, Ganesh and Eising, Ciar{\'a}n},
  journal={arXiv preprint arXiv:2509.26087},
  year={2025}
}

@inproceedings{deng2009imagenet,
  title={Imagenet: A large-scale hierarchical image database},
  author={Deng, Jia and Dong, Wei and Socher, Richard and Li, Li-Jia and Li, Kai and Fei-Fei, Li},
  booktitle={CVPR},
  year={2009},
}

@article{deng2024compact,
  title={Compact 3d gaussian splatting for dense visual slam},
  author={Deng, Tianchen and Chen, Yaohui and Zhang, Leyan and Yang, Jianfei and Yuan, Shenghai and Liu, Jiuming and Wang, Danwei and Wang, Hesheng and Chen, Weidong},
  journal={arXiv preprint arXiv:2403.11247},
  year={2024}
}

@inproceedings{hong2024liv,
  title={Liv-gaussmap: Lidar-inertial-visual fusion for real-time 3d radiance field map rendering},
  author={Hong, Sheng and He, Junjie and Zheng, Xinhu and Zheng, Chunran},
  booktitle={RA-L},
  year={2024},
}

@inproceedings{wang2024endogslam,
  title={Endogslam: Real-time dense reconstruction and tracking in endoscopic surgeries using gaussian splatting},
  author={Wang, Kailing and Yang, Chen and Wang, Yuehao and Li, Sikuang and Wang, Yan and Dou, Qi and Yang, Xiaokang and Shen, Wei},
  booktitle={MICCAI},
  year={2024},
}

@article{liu2024endogaussian,
  title={Endogaussian: Real-time gaussian splatting for dynamic endoscopic scene reconstruction},
  author={Liu, Yifan and Li, Chenxin and Yang, Chen and Yuan, Yixuan},
  journal={arXiv preprint arXiv:2401.12561},
  year={2024}
}

@inproceedings{hu2024gaussianavatar,
  title={Gaussianavatar: Towards realistic human avatar modeling from a single video via animatable 3d gaussians},
  author={Hu, Liangxiao and Zhang, Hongwen and Zhang, Yuxiang and Zhou, Boyao and Liu, Boning and Zhang, Shengping and Nie, Liqiang},
  booktitle={CVPR},
  year={2024}
}

@inproceedings{jiang2024hifi4g,
  title={Hifi4g: High-fidelity human performance rendering via compact gaussian splatting},
  author={Jiang, Yuheng and Shen, Zhehao and Wang, Penghao and Su, Zhuo and Hong, Yu and Zhang, Yingliang and Yu, Jingyi and Xu, Lan},
  booktitle={CVPR},
  year={2024}
}

@inproceedings{liao2025stcocc,
  title={Stcocc: Sparse spatial-temporal cascade renovation for 3d occupancy and scene flow prediction},
  author={Liao, Zhimin and Wei, Ping and Chen, Shuaijia and Wang, Haoxuan and Ren, Ziyang},
  booktitle={CVPR},
  year={2025}
}

@article{muller2022instant,
  title={Instant neural graphics primitives with a multiresolution hash encoding},
  author={M{\"u}ller, Thomas and Evans, Alex and Schied, Christoph and Keller, Alexander},
  journal={ACM TOG},
  year={2022},
}

@article{ye2025gsplat,
  title={{gsplat: An open-source library for Gaussian splatting}},
  author={Ye, Vickie and Li, Ruilong and Kerr, Justin and Turkulainen, Matias and Yi, Brent and Pan, Zhuoyang and Seiskari, Otto and Ye, Jianbo and Hu, Jeffrey and Tancik, Matthew and others},
  journal={Journal of Machine Learning Research},
  year={2025}
}

@article{lee2025timing,
  title={Timing guarantees for inference of AI models in embedded systems},
  author={Lee, Seunghoon and Kang, Woosung and Bertogna, Marko and Chwa, Hoon Sung and Lee, Jinkyu},
  journal={Real-Time Systems},
  year={2025},
  publisher={Springer}
}

@inproceedings{lilja2026queryocc,
  title={QueryOcc: Query-based self-supervision for 3D semantic occupancy},
  author={Lilja, Adam and Lan, Ji and Fu, Junsheng and Hammarstrand, Lars},
  booktitle={Proceedings of the IEEE/CVF Conference on Computer Vision and Pattern Recognition},
  year={2026}
}

@inproceedings{boeder2026shelfocc,
  title={ShelfOcc: Native 3D supervision beyond LiDAR for vision-based occupancy estimation},
  author={Boeder, Simon and Gigengack, Fabian and Roesler, Simon and Caesar, Holger and Risse, Benjamin},
  booktitle={Proceedings of the IEEE/CVF Conference on Computer Vision and Pattern Recognition},
  year={2026}
}

@article{eigen2014depth,
  title={Depth map prediction from a single image using a multi-scale deep network},
  author={Eigen, David and Puhrsch, Christian and Fergus, Rob},
  journal={Advances in neural information processing systems},
  year={2014}
}

@article{khan2020deep,
  title={Deep learning-based monocular depth estimation methods—a state-of-the-art review},
  author={Khan, Faisal and Salahuddin, Saqib and Javidnia, Hossein},
  journal={Sensors},
  year={2020},
  publisher={MDPI}
}

@inproceedings{zheng2024occworld,
  title={Occworld: Learning a 3d occupancy world model for autonomous driving},
  author={Zheng, Wenzhao and Chen, Weiliang and Huang, Yuanhui and Zhang, Borui and Duan, Yueqi and Lu, Jiwen},
  booktitle={European conference on computer vision},
  year={2024},
  organization={Springer}
}

@inproceedings{silberman2012indoor,
  title={Indoor segmentation and support inference from rgbd images},
  author={Silberman, Nathan and Hoiem, Derek and Kohli, Pushmeet and Fergus, Rob},
  booktitle={European conference on computer vision},
  year={2012},
  organization={Springer}
}

@inproceedings{xu2025occ,
  title={Occ-llm: Enhancing autonomous driving with occupancy-based large language models},
  author={Xu, Tianshuo and Lu, Hao and Yan, Xu and Cai, Yingjie and Liu, Bingbing and Chen, Yingcong},
  booktitle={2025 IEEE International Conference on Robotics and Automation (ICRA)},
  year={2025},
  organization={IEEE}
}

@article{wei2024occllama,
  title={Occllama: An occupancy-language-action generative world model for autonomous driving},
  author={Wei, Julong and Yuan, Shanshuai and Li, Pengfei and Hu, Qingda and Gan, Zhongxue and Ding, Wenchao},
  journal={arXiv preprint arXiv:2409.03272},
  year={2024}
}

@article{shi2026come,
  title={Come: Adding scene-centric forecasting control to occupancy world model},
  author={Shi, Yining and Jiang, Kun and Meng, Qiang and Wang, Ke and Wang, Jiabao and Sun, Wenchao and Wen, Tuopu and others},
  journal={Advances in Neural Information Processing Systems},
  volume={38},
  year={2026}
}

@inproceedings{du2026sparseworld,
  title={SparseWorld-TC: Trajectory-Conditioned Sparse Occupancy World Model},
  author={Du, Jiayuan and Zhao, Yiming and Guo, Zhenglong and Pan, Yong and Hou, Wenbo and Hao, Zhihui and Zhan, Kun and Chen, Qijun},
  booktitle={Proceedings of the IEEE/CVF Conference on Computer Vision and Pattern Recognition},
  year={2026}
}

@article{gu2024dome,
  title={Dome: Taming diffusion model into high-fidelity controllable occupancy world model},
  author={Gu, Songen and Yin, Wei and Jin, Bu and Guo, Xiaoyang and Wang, Junming and Li, Haodong and Zhang, Qian and Long, Xiaoxiao},
  journal={arXiv preprint arXiv:2410.10429},
  year={2024}
}

@inproceedings{zhao2026shelfgaussian,
  title={ShelfGaussian: Shelf-supervised open-vocabulary gaussian-based 3D scene understanding},
  author={Zhao, Lingjun and Luo, Yandong and Hays, James and Gan, Lu},
  booktitle={Proceedings of the IEEE/CVF Conference on Computer Vision and Pattern Recognition},
  year={2026}
}

@inproceedings{li2024ego,
  title={Is ego status all you need for open-loop end-to-end autonomous driving?},
  author={Li, Zhiqi and Yu, Zhiding and Lan, Shiyi and Li, Jiahan and Kautz, Jan and Lu, Tong and Alvarez, Jose M},
  booktitle={Proceedings of the IEEE/CVF Conference on Computer Vision and Pattern Recognition},
  year={2024}
}

@inproceedings{hu2023planning,
  title={Planning-oriented autonomous driving},
  author={Hu, Yihan and Yang, Jiazhi and Chen, Li and Li, Keyu and Sima, Chonghao and Zhu, Xizhou and Chai, Siqi and Du, Senyao and Lin, Tianwei and Wang, Wenhai and others},
  booktitle={Proceedings of the IEEE/CVF conference on computer vision and pattern recognition},
  year={2023}
}

@inproceedings{hu2022st,
  title={St-p3: End-to-end vision-based autonomous driving via spatial-temporal feature learning},
  author={Hu, Shengchao and Chen, Li and Wu, Penghao and Li, Hongyang and Yan, Junchi and Tao, Dacheng},
  booktitle={European Conference on Computer Vision},
  year={2022},
  organization={Springer}
}
}

\maketitlesupplementary
\appendix

\addtocontents{toc}{\protect\setcounter{tocdepth}{3}}
\tableofcontents
\section*{Overview}
We divide the supplementary material into three sections:
\begin{enumerate}
    \item \textbf{Supplementary Information.} In \autoref{sec:supp_info}, we provide additional details that assist the reader in understanding concepts not included in the main paper, such as the loss formulation and the evaluation metrics.

    \item \textbf{Supplementary Ablations.} In \autoref{sec:supp_res}, we present additional ablation studies such as generalisation to the NYUv2 \cite{silberman2012indoor} dataset, and insight into the rendering methods used in \method{}, specifically \render{}. We also provide detailed metrics for occupancy estimation and downstream tasks.
    
    \item \textbf{Supplementary Visualisations.} In \autoref{sec:supp_vis}, we include visualisations for \method{} on occupancy and depth, NYUv2, and downstream tasks. Video visualisations are also provided.
    
\end{enumerate}
\newpage

\setcounter{equation}{0}
\setcounter{figure}{0}
\setcounter{table}{0}

\section{Supplementary Information}
\label{sec:supp_info}
In this section, we provide additional information that could not be included in the main paper due to length constraints.

\subsection{Limitations and Future Work}
In this section, we discuss the limitations and future directions of our proposed model, SuperQuadricOcc (SQOcc), as well as those of the rendering method, SuperQuadricOcc-Render (SQOcc-R).

In \methodShort{}, the use of superquadrics improves performance metrics while reducing inference time. However, incorporating 3D pseudo-labels alongside our rendering pipeline could further enforce spatial consistency \cite{hayes2025easyocc}, and increase performance metrics \cite{chambon2025gaussrender, boeder2026shelfocc, zhao2026shelfgaussian}. Additional work on constraining certain class types to specific epsilon $\boldsymbol{\varepsilon}$ and scale $\mathbf{s}$ ranges may help the model develop a more structured representation, which could further reduce the number of primitives required.

Within the spatial indexing strategy of SQOcc-R, incorporating the superquadric occupancy probability into the indexing process may reduce inference time by avoiding queries to superquadrics with low opacity or rapidly decaying influence. However, this may introduce challenges, as it could result in a dynamic number of superquadrics per voxel, disrupting the batching mechanism that currently enables efficient computation.

Future work on a native superquadric rasteriser would allow direct screen-space rendering, enabling faster and more accurate rendering. This would remove the need for a Gaussian or voxel rendering proxy and may lead to improved performance compared with SQOcc-R.
\subsection{Problem Statement}
The task of semantic occupancy estimation aims to predict the semantic class $\mathcal{C}=\{c_0, c_1, \dots, c_n\}$ for each voxel in a discretised 3D space using a model $M$. For the nuScenes dataset, the model may utilise surround-view camera images, LiDAR, radar, or a combination of these modalities. Let the input data over a temporal horizon $T=\{t-n, \dots, t\}$ be denoted as $D_n=\{D_{t-n}, \dots, D_t\}$. The model processes this sequence as $M(D_n)$ to produce a scene representation. The form of the output depends on the chosen representation and stage of operation, and may consist of voxel grids, Gaussian primitives, or superquadrics.

In our self-supervised approach, SQOcc outputs a set of refined superquadric primitives $\mathcal{S}$. During training, $\mathcal{S}$ is rendered, and losses are computed against 2D pseudo-labels generated by \glspl{vfm} \cite{hu2024metric3d, ren2024grounded}. During inference, the superquadrics are voxelised, and evaluation metrics are computed against dense ground-truth voxel grids.
\subsection{Multi-Layer Gaussian Approximation}
The design of the Multi-Layer Gaussian Approximation (MLGA) is discussed in the main paper in Section 4.4. Here, we provide additional detail on its implementation.

As described in the main paper, the goal of MLGA is to approximate the superquadric occupancy probability function $p_o(\mathbf{x}, S)$ for an arbitrary superquadric $S$ using a set of Gaussians. We first reduce $S$ to its canonical form $S=(\boldsymbol{s}, \varepsilon_1, \varepsilon_2)$, and later apply translation and rotation to place it in world space. Our implementation follows PartGS \cite{gao2025self} and GaussianBlock \cite{jiang2024gaussianblock}, which builds upon SuGaR \cite{guedon2024sugar}, and GaMeS \cite{waczynska2024games}, methods that place Gaussians on mesh surfaces.

We begin by augmenting the scale parameters of $S$ with $n$ positive scalars $P = \{\, p_1, p_2, \ldots, p_n \,\} \subset \mathbb{R}^+$ to define a scaled family of superquadrics: $S_P = \{\, (p\,\boldsymbol{s}, \varepsilon_1,\, \varepsilon_2) \mid p \in P \,\}$. This produces a layered superquadric structure analogous to level sets.

Following PartGS \cite{gao2025self}, for $S_p \in S_P$, we construct a superquadric mesh by first generating an icosphere. Let $(x,y,z)$ denote the vertices of the icosphere. For each vertex, we compute the spherical parameters $\eta = \arcsin(y)$ and $\omega = \text{atan2}(x, z)$. The vertex is then mapped onto the superquadric surface using signed powers of trigonometric functions controlled by $\varepsilon_1$ and $\varepsilon_2$:
\begin{equation}
\begin{aligned}
\mathbf{v} &= [v_1, v_2, v_3], \\
v_1 &= s_1 \cos^{\epsilon_1}(\eta)\cos^{\epsilon_2}(\omega), \\
v_2 &= s_2 \sin^{\epsilon_1}(\eta), \\
v_3 &= s_3 \cos^{\epsilon_1}(\eta)\sin^{\epsilon_2}(\omega).
\end{aligned}
\end{equation}
For each face $V$ with vertices $v_1,v_2,v_3 \in \mathbb{R}^3$, instantiate a Gaussian $G$ with mean $\mathbf{m}_G$ as the centroid of the face. The Gaussian rotation $R_G$ is defined by constructing orthonormal vectors $r_1, r_2, r_3 \in \mathbb{R}^3$ to form the rotation matrix $R_G = [r_1, r_2, r_3].$

The vector $r_1$ is aligned with the normal vector of $V$, while $r_2$ is defined as the vector from the centroid of the triangle to the first vertex $v_1$. The third vector $r_3$ is obtained by orthonormalising the vector from the centroid to the second vertex with respect to $r_1$ and $r_2$:
\begin{equation}
r_2 = \frac{v_1 - m}{\|v_1 - m\|},
\quad
r_3 =
\frac{\text{ort}(v_2 - m; r_1, r_2)}
{\|\text{ort}(v_2 - m; r_1, r_2)\|}
\end{equation}
where $\text{ort}$ denotes orthonormalisation using the Gram--Schmidt process. For the Gaussian scale $S_G$, we assign $ s_x = \|m - v_1\|$ and $ s_y = \langle v_2, r_3 \rangle$, allowing the Gaussians to adequately cover the vertex face.

We build upon previous works, extending this approximation to 3D space by introducing a $z$ scale $s_z$. For a Gaussian $G_p$ with centre $\mathbf{x}_{p}$ sampled on superquadric $S_{p}$, we define a corresponding Gaussian $\mathbf{x}_{p+1}$ on $S_{p+1}$ at the same angular coordinates $(\eta, \omega)$ and assign $s_z$ as half the euclidean distance:
\begin{equation}
s_z
= \frac{1}{2} \|\mathbf{x}_{p+1}(\eta, \omega) - \mathbf{x}_{p}(\eta, \omega)\|
\end{equation}
This assignment enables the Gaussians to model the space between scaled superquadrics, producing a more uniform density. The final Gaussian scale $S_G$ is then defined as:
\begin{equation}
S_G = \text{diag}(s_x, s_y, s_z)
\end{equation}

Previous works assign the Gaussian opacity directly from the parent superquadric. In our case, since Gaussians are used to model occupancy probability, we instead define:
\begin{equation}
\sigma = \sigma_{S_p} \cdot e^{f(\mathbf{m})}
\end{equation}
This scaling aligns the Gaussian occupancy with that of the superquadric, ensuring that the occupancy evaluated at the Gaussian centre matches the corresponding value in the superquadric distribution.

Finally, all Gaussians are transformed into the world coordinate system using rotation and translation defined by the parent superquadric mean $\mathbf{m}$ and quaternion $\mathbf{r}$.
\subsection{SuperQuadric Voxelisation}
To enable direct comparison with previous state-of-the-art methods, we develop a superquadric voxelization module for evaluation. We extend the Gaussian voxelization strategy of Boeder et al. \cite{boeder2025gaussianflowocc} to superquadric primitives. We iterate over each voxel $v$ with centre $\boldsymbol{p}$, summing the contributions from all superquadrics $S$ within a cubic neighbourhood of $\mathcal{N}_v$. Following previous works \cite{boeder2025gaussianflowocc}, we set $\mathcal{N}_v=5$. For each $S_i$ within the cubic neighbourhood $\mathcal{N}_v$, we  sum the occupancy probability to obtain a density and semantic voxel:
\begin{equation}
\label{eq:voxelisation}
\begin{aligned}
v_d(\mathbf{p}) &= \sum_{S_i \in N_v} p_o(\boldsymbol{p}, S_i)\,\sigma_i, \\
v_s(\mathbf{p}) &= \sum_{S_i \in N_v} p_o(\boldsymbol{p}, S_i)\,\mathbf{c}_i
\end{aligned}
\end{equation}
Voxels below a defined threshold $\tau$ in $v_d$ are assigned as free in $v_s$ to obtain the final prediction. We assign $\tau=0.05$ through empirical tuning.
\subsection{Loss Formulation}
For fair comparison, we adopt the full loss formulation of GaussianFlowOcc \cite{boeder2025gaussianflowocc}. Supervision uses 2D pseudo-labels from VFMs: Grounded-SAM \cite{ren2024grounded} for semantics ($S$) and Metric3Dv2 \cite{hu2024metric3d} for depth ($D$). During training, the chosen render method provides rendered semantics ($\hat{S}$) and depth ($\hat{D}$). The semantic loss $\mathcal{L}_{\text{sem}}$ is cross-entropy with $\lambda_s=2$, and the depth loss $\mathcal{L}_{\text{depth}}$ is MSE with $\lambda_d=0.05$. The full loss formulation is provided in Equation (\ref{eq:loss}). Training is performed using Adam with a learning rate of $10^{-4}$.
\begin{equation}
\label{eq:loss}
\mathcal{L}
=
\lambda_{s} \ \mathcal{L}_{\text{sem}}(\hat{S}, S)
+
\lambda_{d} \ \mathcal{L}_{\text{depth}}(\hat{D}, D)
\end{equation}
Following the training of \methodShort{}, we train the trajectory estimator, adapted from ShelfGaussian \cite{zhao2026shelfgaussian}, for 12 epochs while retaining \methodShort{} as frozen. We use an L1 loss on the estimated trajectories and ground-truth trajectories as per prior work \cite{zhao2026shelfgaussian}. Training is performed using Adam with a learning rate of $10^{-4}$.

For experimentation on NYUv2 \cite{silberman2012indoor}, we utilize Cross-Entropy and Binary Cross-Entropy for the semantic and density voxel, respectively, which are obtained from Equation (\ref{eq:voxelisation}). For optional depth supervision, we use MSE.
\subsection{Datasets}
\label{subsec:datasets}
\subsubsection{nuScenes}
The nuScenes \cite{nuScenes} dataset contains 1000 scenes of urban driving collected in Boston and Singapore. The ego-vehicle is equipped with six surround-view cameras, one 32-laser LiDAR sensor, and five radar sensors. The dataset is split into 700 scenes for training, 150 for validation, and 150 unannotated scenes for testing. Each scene contains approximately 40 key frames, representing around 20 seconds of driving. For each annotated key frame, ground-truth 3D bounding boxes and \gls{bev} labels are provided to support training and validation. 

For nuScenes-lidarseg, each LiDAR point cloud is manually annotated with a semantic class. This, alongside the previously mentioned ground-truth labels, enables the construction of dense 3D occupancy datasets.

\subsubsection{Occ3D}
The Occ3D \cite{tian2023occ3d} dataset follows the design of earlier occupancy datasets \cite{wei2023surroundocc, wang2023openoccupancy}, using multi-frame aggregation to densify the point cloud for each sample for both the nuScenes \cite{nuScenes} and Waymo \cite{sun2020waymo} datasets. Additional refinements, including occlusion reasoning and image guidance, are applied to generate higher-quality ground-truth labels. The Occ3D-nuScenes dataset contains 600 scenes for training and 150 for validation. The evaluation space spans $[-40\text{m}, 40\text{m}]$ along the $X$ and $Y$ axes and $[-1\text{m}, 5.4\text{m}]$ along the $Z$ axis, with a voxel resolution of $0.4\text{m}^3$, defined in the ego-vehicle coordinate frame.

Occ3D-nuScenes is the most widely used dataset for evaluating self-supervised models, largely due to the inclusion of a camera mask that restricts the evaluation space to regions visible in the camera views. However, the RayIoU metric also enables similar evaluation on other datasets.

\subsubsection{OpenOccv2}
The OpenOccv2 \cite{tong2023scene} dataset provides occupancy labels in a similar manner to Occ3D, using multi-frame LiDAR aggregation, with an equivalent evaluation space and number of scenes. It provides denser occupancy labels with reduced noise due to dedicated noise removal, which more effectively penalises models that overpredict. This dataset does not include a camera mask, and therefore, we do not report IoU or mIoU; instead, we report RayIoU.

\subsubsection{NYUv2} 
\label{subsubsec:datasets_nyuv2} 
The NYUv2 \cite{silberman2012indoor} indoor dataset is collected from a Kinect sensor, which provides RGBD images, containing 1449 samples from 464 indoor scenes. Ground-truth semantic occupancy grids are provided, with 13 known classes, 1 free, and 1 unknown. We utilize a 795/654 train/test split and perform evaluation on a 1:4 scale as per prior work \cite{cao2022monoscene}. The scene range is $4.8m\times4.8m\times2.88m$, and the resolution is 0.08m. Unlike nuScenes, there is no temporal aspect to the data.
\subsection{Evaluation Metrics}
For semantic occupancy evaluation, we use three standard metrics: \gls{iou}, \gls{miou}, and RayIoU. For depth evaluation, we use Abs Rel, Sq Rel, RMSE, RMSE log, and threshold accuracy metrics ($\delta$).

\subsubsection{Intersection over Union (IoU)} For \gls{iou}, we first convert the semantic voxel grid into a binary grid by assigning the \textit{free} class to 0 and all other classes to 1. The \gls{iou} formula is given as follows: 
\begin{equation}
\label{eq:iou}
\text{IoU} = \frac{TP}{TP + FP + FN}
\end{equation}
where $TP$, $FP$, and $FN$ denote true positives, false positives, and false negatives, respectively. \gls{iou} measures the model's ability to estimate occupancy and is traditionally less preferred since it does not penalise semantic misclassifications, which are important in automotive perception. 

\subsubsection{Mean Intersection over Union (mIoU)} This metric is more commonly used as it properly penalises misclassifications, providing a better measure of the model's ability to estimate both occupancy and semantic class. The formulation is given as follows:
\begin{equation}
\label{eq:miou}
\text{mIoU} = \frac{1}{C} \sum_{\substack{c=1}}^{C} \frac{TP_c}{TP_c + FP_c + FN_c}
\end{equation}
where $C$ denotes the total number of semantic classes. The empty class $c=0$ is excluded from the calculation.

\subsubsection{RayIoU} \label{subsubsec:rayiou} First introduced in SparseOcc \cite{tang2024sparseocc}, this metric addresses issues such as over-penalisation and over-prediction that can artificially inflate mIoU scores while not accurately reflecting overall scene completion. RayIoU operates by projecting rays from the LiDAR viewpoint into the ground-truth voxel grid and traversing them until a voxel is intersected. The same ray is then projected into the predicted voxel grid. A true positive is assigned if the semantic class matches and the L1 distance between the ground-truth ray and predicted ray is within a defined threshold. The formulation is similar to that of \gls{miou}:
\begin{equation}
\label{eq:rayiou}
\text{RayIoU} = \frac{1}{C} \sum_{\substack{c=1}}^{C} \frac{TP_c}{TP_c + FP_c + FN_c}
\end{equation}

\subsubsection{Trajectory Metrics}
L2 measures the Euclidean distance between estimated and ground truth trajectories. The collision rate (CR) measures the percentage of instances where the estimated trajectory intersects with dynamic agents in the scene for the predicted timestep.

\subsubsection{Depth Metrics}
We use the depth metrics adopted in previous works to evaluate depth estimation accuracy:
\begin{enumerate}
\item \textbf{Abs Rel:} 
$\frac{1}{|T|}\sum_{d\in T}\frac{|d-d^*|}{d^*}$

\item \textbf{Sq Rel:} 
$\frac{1}{|T|}\sum_{d\in T}\frac{|d-d^*|^2}{d^*}$

\item \textbf{RMSE:} 
$\sqrt{\frac{1}{|T|}\sum_{d\in T}|d-d^*|^2}$

\item \textbf{RMSE$_{\log}$:} 
$\sqrt{\frac{1}{|T|}\sum_{d\in T}|\log d-\log d^*|^2}$

\item $\boldsymbol{\delta} \mathbf{< t}:$
percentage of $d$ such that $\max\!\left(\frac{d}{d^*},\frac{d^*}{d}\right) < t$
\end{enumerate}

where $d$ denotes the ground-truth depth map, $d^{*}$ denotes the predicted depth map, and $T$ represents the total number of evaluated pixels.
\subsection{Primitive Count}
In Table \ref{tab:primitive_count}, we compare the total primitive count across various models, highlighting the efficiency of SQOcc's scene representation through the use of superquadrics.

\begin{table}[ht]
    \caption{\textbf{Gaussian/SuperQuadric primitive count comparison.} * GaussianOcc uses Gaussians during training only, converting each voxel to a Gaussian. The remaining models are for training and inference.}
  \centering
  \begin{adjustbox}{width=0.84\linewidth}
  \begin{tabular}{
    l
    |c|c
}
    \toprule
\textbf{Method} & \textbf{Representation} & \textbf{Primitives} \\
    \midrule 
    TT-Occ \cite{zhang2025tt} & Gaussian & 2,968,647 \\
    GaussianOcc* \cite{gaussianocc} & Gaussian & 640,000\\
    GaussianFlowOcc \cite{boeder2025gaussianflowocc} & Gaussian & 10,000 \\
    GaussTR \cite{jiang2025gausstr} & Gaussian & 1,800 \\
    \midrule
    SQOcc (Ours) & SuperQuadric & \textbf{1,600} \\
    \bottomrule
  \end{tabular}
  \end{adjustbox}
  \label{tab:primitive_count}
\end{table}
\begin{table*}[t]
\centering
\caption{\textbf{Model comparison} on the NYUv2 \cite{silberman2012indoor} dataset. The \textit{(+/-)} indicates the change if rendering loss versus ground truth depth is enabled. Class-wise metrics are for mIoU. The best performer is highlighted in \textbf{bold}.}
\begin{adjustbox}{width=\linewidth}
\begin{tabular}{l|c|c|ccc|ccccccccccc}
\toprule
\textbf{Method}
& \textbf{Num. Primitives}
& \textbf{FPS}
& \textbf{RayIoU}
& \textbf{IoU}
& \textbf{mIoU}
& \rotatebox{90}{\textcolor{ceiling}{$\blacksquare$} ceiling}
& \rotatebox{90}{\textcolor{floor}{$\blacksquare$} floor}
& \rotatebox{90}{\textcolor{wall}{$\blacksquare$} wall}
& \rotatebox{90}{\textcolor{window}{$\blacksquare$} window}
& \rotatebox{90}{\textcolor{chair}{$\blacksquare$} chair}
& \rotatebox{90}{\textcolor{bed}{$\blacksquare$} bed}
& \rotatebox{90}{\textcolor{sofa}{$\blacksquare$} sofa}
& \rotatebox{90}{\textcolor{table}{$\blacksquare$} table}
& \rotatebox{90}{\textcolor{tvs}{$\blacksquare$} tvs}
& \rotatebox{90}{\textcolor{furniture}{$\blacksquare$} furniture}
& \rotatebox{90}{\textcolor{objects}{$\blacksquare$} objects} \\
\midrule

GaussianFlowOcc \cite{boeder2025gaussianflowocc} & 1,600 & 29.8
& 13.3 \textit{(+1.8)}
& 33.1 \textit{(-1.2)}
& 19.1 \textit{(-0.4)}
& 6.1 & 72.7 & 7.3 & 6.6 & 8.1 & 36.7 & 29.1 & 10.2 & 4.4 & 21.3 & 7.8 \\

\rowcolor{black!5}
GaussianFlowOcc \cite{boeder2025gaussianflowocc} & 10,000 & 10.6
& 16.2 \textit{(-1.3)}
& 34.9 \textit{(-3.0)}
& 20.5 \textit{(-0.6)}
& 5.5 & 82.8 & 6.9 & \textbf{7.0} & 8.6 & 37.0 & 29.1 & 10.6 & 7.4 & 21.9 & 8.5 \\

SQOcc (Ours) & 1,600 & \textbf{30.5}
& 17.0 \textit{(+2.8)}
& 35.5 \textit{(+0.4)}
& 20.8 \textit{(+0.2)}
& 7.5 & 87.0 & 7.1 & 6.6 & 8.7 & 36.8 & 29.5 & 11.8 & 4.3 & 22.1 & 7.8 \\

\rowcolor{black!5}
SQOcc (Ours) & 10,000 & 11.1
& \textbf{19.4} \textit{(+0.9)}
& \textbf{36.5} \textit{(-1.5)}
& \textbf{21.8} \textit{(-1.5)}
& \textbf{8.2} & \textbf{87.8} & \textbf{7.6} & 6.7 & \textbf{10.0} & \textbf{38.2} & \textbf{29.8} & \textbf{11.9} & \textbf{8.8} & \textbf{22.4} & \textbf{8.9} \\
\bottomrule
\end{tabular}
\end{adjustbox}

\label{tab:nyuv2}
\end{table*}

Many Gaussian-based models employ large numbers of Gaussian primitives. For example, TT-Occ \cite{zhang2025tt} uses a large primitive count due to its strategy of initialising Gaussians on a pixel basis. GaussTR uses fewer primitives; however, it relies heavily on large foundation models during inference, which significantly limits real-time deployment. In contrast, SQOcc requires relatively few superquadric primitives due to their expressive shape family, enabling compact scene modelling. Combined with our efficient rendering proxy used for supervision, this results in improved performance compared with previous methods.

\subsection{Scale and Epsilon Constraints}
We constrain the scale $\mathbf{s}$ to the range $[0.01,1]$, following previous works \cite{boeder2025gaussianflowocc}. We constrain the density $\sigma$ to $[0,1]$. For the epsilon parameters $\boldsymbol{\varepsilon}$, we restrict their range to $[0.1, 2.0]$, consistent with prior work \cite{zuo2025quadricformer}.

\section{Supplementary Ablations}
\label{sec:supp_res}
In this section, we present additional ablation results for SuperQuadricOcc (SQOcc), focusing on the rendering pipelines used for supervision.

\subsection{Domain Generalisation: NYUv2}
To demonstrate the adaptability of our approach across domains, we experiment with GaussianFlowOcc and \method{} on the NYUv2 dataset \cite{silberman2012indoor}. This dataset is described in Section \ref{subsubsec:datasets_nyuv2}, and we evaluate the model on the IoU, mIoU, and RayIoU metrics. 

For RayIoU, we adapt the formulation from SparseOcc \cite{tang2024sparseocc}, where it is performed for autonomous driving purposes, and rays are emitted from the LiDAR sensor. Here, we compute pixel-wise RayIoU, where for each pixel in the image, a ray is cast into the scene using the known extrinsic and intrinsic calibration parameters. These serve as the evaluation rays, and evaluation is performed as described in Section \ref{subsubsec:rayiou}. For distance thresholds, we use  (0.2m, 0.4m, 0.8m) compared to (1m, 2m, 4m) for nuScenes due to the smaller world bounds.

We additionally experiment with adding depth supervision, using Gaussian Splatting with GaussianFlowOcc, and \render{} with \method{}. We utilize the same model setup and hyperparameters for \renderShort{} to further display generalisation.

As evident in Table \ref{tab:nyuv2}, \methodShort{} (1,600) outperforms GFlow (10,000) for all metrics, namely RayIoU (+0.8), with faster inference ($\sim$3x). Our method with 10,000 primitives achieves top performance for all except one class, showing that superquadrics can better represent indoor scenes compared to Gaussians, in addition to outdoor dynamic scenes, as seen in the main paper. When incorporating depth supervision, results are mixed,  perhaps due to rendering enforcing less densification in such a close environment.

Visualisations are provided in \autoref{fig:nyuv2_vis}.
\subsection{Rendering Ablation Studies}
We evaluate our ablations on the Occ3D-nuScenes \cite{tian2023occ3d} and OpenOccv2 \cite{tong2023scene} datasets. To reduce computational cost, we train Gaussianisation, Voxel NeRF, and SQOcc-R on the first half of the dataset (300 scenes) and evaluate them on the full validation set. MLGA is trained on the full dataset, as it trains significantly faster without the temporal module.

\subsubsection{Multi-Layer Gaussian Approximation (MLGA)} Here, we investigate several components of the MLGA superquadric approximation proposed in the main paper.

The hyperparameters in \autoref{tab:mlga_abl} were selected during an earlier stage of development, using mIoU/IoU and rendering latency as the primary criteria. RayIoU is reported for completeness and consistency with the final evaluation protocol. Consequently, some selected settings would differ if RayIoU alone were used for model selection. This does not affect the main conclusion of the ablation: MLGA can serve as an effective rendering proxy at moderate primitive counts, but it is sensitive to excessive primitive overlap and does not preserve the superquadric occupancy distribution as reliably as \renderShort{}. Its degradation when combined with the temporal flow module further indicates limitations as a supervision source for superquadric rendering. Overall, this sensitivity to hyperparameter choice, which is not observed to the same extent for \renderShort{}, Gaussianisation, or Voxel NeRF, motivates our use of \renderShort{} as the primary renderer.


\begin{table*}[t]
\centering
\caption{\textbf{Multi-Layer Gaussian Approximation (MLGA) ablation studies.} The parameter highlighted in grey is the selected development value,  corresponding to a compromise between prediction metrics and latency. Lat. denotes latency, the average time to render each sample of the entire validation set, measured in milliseconds. The best performer is highlighted in \textbf{bold}.}
\label{tab:mlga_ablation}

\setlength{\tabcolsep}{4pt}

\begin{subtable}{0.49\linewidth}
\centering
\caption{SuperQuadric Count}
\label{tab:num_quadrics}
\resizebox{0.8\linewidth}{!}{
\begin{tabular}{c|cc|cc|c}
\toprule
&
&
&
\multicolumn{2}{c|}{\textbf{RayIoU}}
& \\
$\mathbf{N}$ & \textbf{IoU} & \textbf{mIoU} & \textbf{Occ3D} & \textbf{OpenOccv2} & \textbf{Lat.} \\
\midrule
1200  & 32.1 & 12.4 & \textbf{12.9} & \textbf{13.5} & \textbf{35.3} \\
\rowcolor{black!8}
1600  & 33.7 & \textbf{12.7} & 12.1 & 13.1 & 44.0 \\
2000  & \textbf{34.0} & \textbf{12.7} & 11.9 & 12.6 & 44.7 \\
5000  & 27.8 & 7.0 & 5.9 & 6.5 & 60.5 \\
10000 & 27.9 & 4.6 & 3.6 & 4.1 & 121.4 \\
\bottomrule
\end{tabular}}
\end{subtable}\hfill
\begin{subtable}{0.49\linewidth}
\centering
\caption{Icosphere Faces}
\label{tab:level}
\resizebox{0.85\linewidth}{!}{
\begin{tabular}{cc|cc|cc|c}
\toprule
&
&
&
&
\multicolumn{2}{c|}{\textbf{RayIoU}}
& \\
\textbf{Level} & \textbf{Faces} & \textbf{IoU} & \textbf{mIoU} & \textbf{Occ3D} & \textbf{OpenOccv2} & \textbf{Lat.} \\
\midrule
0 & 20   & 32.2 & \textbf{12.7} & 12.0 & 13.0 & \textbf{36.0} \\
\rowcolor{black!8}
1 & 80   & \textbf{33.7} & \textbf{12.7} & \textbf{12.1} & \textbf{13.1} & 44.0 \\
2 & 320  & 33.3 & 12.3 & 10.4 & 11.3 & 73.0 \\
3 & 1280 & OOM & OOM & OOM & OOM & OOM \\
\bottomrule
\end{tabular}}
\end{subtable}

\vspace{0.5em}

\begin{subtable}{0.49\linewidth}
\centering
\caption{Gaussian z-scale ${s_z}$}
\label{tab:zscale}
\resizebox{0.85\linewidth}{!}{
\begin{tabular}{c|cc|cc|c}
\toprule
&
&
&
\multicolumn{2}{c|}{\textbf{RayIoU}}
& \\
$\mathbf{{s_z}}$ & \textbf{IoU} & \textbf{mIoU} & \textbf{Occ3D} & \textbf{OpenOccv2} & \textbf{Lat.} \\
\midrule
0.01 & 33.2 & 12.6 & \textbf{12.7} & \textbf{13.7} & \textbf{43.9} \\
0.1 & 29.7 & 11.3 & 10.0 & 10.8 & 44.3 \\
\rowcolor{black!8}
$\frac{1}{2}$-Euclidean & \textbf{33.7} & \textbf{12.7} & 12.1 & 13.1 & 44.0 \\
\bottomrule
\end{tabular}}
\end{subtable}\hfill
\begin{subtable}{0.49\linewidth}
\centering
\caption{Scale Gaussian Opacity}
\label{tab:scale_opacity}
\resizebox{0.85\linewidth}{!}{
\begin{tabular}{c|cc|cc|c}
\toprule
&
&
&
\multicolumn{2}{c|}{\textbf{RayIoU}}
& \\
\textbf{Scale Opacity} & \textbf{IoU} & \textbf{mIoU} & \textbf{Occ3D} & \textbf{OpenOccv2} & \textbf{Lat.} \\
\midrule
\rowcolor{black!8}
\cmark & \textbf{33.7} & \textbf{12.7} & \textbf{12.1} & \textbf{13.1} & 44.0 \\
\xmark & 22.7 & 8.2 & 11.9 & 11.4 & \textbf{40.6} \\
\bottomrule
\end{tabular}}
\end{subtable}

\vspace{0.5em}

\begin{subtable}{\linewidth}
\centering
\caption{Scale Values}
\label{tab:scale_values}
\resizebox{0.7\linewidth}{!}{
\begin{tabular}{c|cc|cc|c}
\toprule
&
&
&
\multicolumn{2}{c|}{\textbf{RayIoU}}
& \\
\textbf{Scale Values} & \textbf{IoU} & \textbf{mIoU} & \textbf{Occ3D} & \textbf{OpenOccv2} & \textbf{Lat.} \\
\midrule
\rowcolor{black!8}
{[}0.5, 0.6, 0.75, 0.9, 1.05, 1.2, 1.6, 2.0, 2.5{]} 
& 33.7 & \textbf{12.7} & 12.1 & 13.1 & 44.0 \\
{[}0.1, 0.22, 0.40, 0.58, 0.76, 0.94, 1.42, 1.90, 2.5{]} 
& 32.9 & 12.4 & 11.5 & 12.2 & \textbf{40.1} \\
{[}0.5, 0.65, 0.88, 1.1, 1.33, 1.55, 2.15, 2.75, 3.5{]} 
& 32.5 & 12.2 & 12.8 & 13.4 & 49.4 \\
{[}0.5, 0.7, 0.9, 1.1, 1.3, 1.5, 1.7, 1.9, 2.1, 2.3, 2.5{]} 
& \textbf{33.8} & 12.5 & \textbf{13.0} & \textbf{14.0} & 51.4 \\
\bottomrule
\end{tabular}}
\end{subtable}

\label{tab:mlga_abl}
\end{table*}

\textbf{Superquadric Count $N$.} In Table \ref{tab:num_quadrics}, at lower primitive counts ($<$2000), the MLGA strategy provides effective supervision. However, when the primitive count exceeds 2000, performance degrades significantly, possibly due to confusion in the render proxy caused by excessive overlapping superquadrics and therefore Gaussians. We note this as a key limitation of this method. Choosing $N=1600$ provides the best balance between model performance and render time.

\textbf{Icosphere Faces.} Table \ref{tab:level} examines the number of faces per scaled superquadric in the superquadric-to-Gaussian module. Results show that using 80 faces per scaled superquadric provides the best balance, offering sufficient surface coverage without the computational overhead observed with 320 faces. Using 20 faces may also be considered effective, as it reduces latency with minimal decrease in the other metrics. At 320 faces, results show a noticeable drop, indicating that using too many Gaussians may interfere with the supervision signal. Notably, configurations with 1280 faces result in an out-of-memory error.

\textbf{Gaussian z-scale $s_z$.} Table \ref{tab:zscale} evaluates the formulation of the Gaussian z-scale $s_z$, where $s_z=0.01$ provides the best performance. For the half-Euclidean distance, it provides adequate performance, but perhaps a smaller scaling factor, such as $\frac{1}{4}$ or $\frac{1}{8}$, could allow for a better approximation, allowing the Gaussian to decay more gradually over space. Latency remains approximately the same across all variants. 

\textbf{Scaling of Gaussian Opacity.} Table \ref{tab:scale_opacity} evaluates the inclusion of the superquadric scaling term, $e^{f(\mathbf{m})}$, in the Gaussian opacity formulation. Including this term improves performance by enabling Gaussians to better approximate the superquadric distribution, resulting in a more consistent supervision signal. Not using scale opacity reduces render time, likely due to fewer superquadrics appearing in the camera view as a result of less complete scene reconstruction, as indicated by the results.

\textbf{Scale Values.} Table \ref{tab:scale_values} examines the effect of different scale values for the scaled family of superquadrics, $S_K$. We observe that all choices yield adequate performance. In particular, the fourth choice, which uses a constant difference between consecutive scale values, provides strong coverage of the distribution.


\subsubsection{Gaussianisation and Voxel NeRF}
Here, we investigate the optimal Gaussian scale $\mathbf{s}$ and the Voxel NeRF ray sample number $L$.

\begin{table}[htpb]
\centering
\caption{\textbf{Gaussianisation and Voxel NeRF ablation studies.} The parameter highlighted in grey is the selected value, corresponding to a compromise between prediction metrics and latency. Lat. denotes latency, the average time to render each sample of the entire validation set, measured in milliseconds. The best performer is highlighted in \textbf{bold}.}
\label{tab:ablation}

\begin{subtable}{\linewidth}
\centering
\caption{Gaussian scale $s$}
\label{tab:gaussian_scale}
\begin{adjustbox}{width=0.8\linewidth}
\begin{tabular}{c|cc|cc|c}
\toprule
&
&
&
\multicolumn{2}{c|}{\textbf{RayIoU}} 
& \\
\textbf{$s$}
& \textbf{IoU}
& \textbf{mIoU}
& \textbf{Occ3D}
& \textbf{OpenOccv2}
& \textbf{Lat.} \\
\midrule
\rowcolor{black!8}
0.15 & \textbf{43.1} & \textbf{16.0} & \textbf{16.0} & \textbf{16.6} & \textbf{13.8} \\
0.20 & 42.2 & \textbf{16.0} & 15.7 & 15.9 & 15.3 \\
0.28 & 40.5 & 15.5 & 14.7 & 14.7 & 15.6 \\
0.40 & 36.6 & 14.3 & 14.0 & 13.4 & 15.5 \\
\bottomrule
\end{tabular}
\end{adjustbox}
\end{subtable}

\vspace{0.75em}

\begin{subtable}{\linewidth}
\centering
\caption{Voxel NeRF ray samples $L$}
\label{tab:voxelnerf_abl}
\begin{adjustbox}{width=0.8\linewidth}
\begin{tabular}{c|cc|cc|c}
\toprule
&
&
&
\multicolumn{2}{c|}{\textbf{RayIoU}} 
& \\
\textbf{$L$}
& \textbf{IoU}
& \textbf{mIoU}
& \textbf{Occ3D}
& \textbf{OpenOccv2}
& \textbf{Lat.} \\
\midrule
50 & 33.0 & 12.1 & 11.2 & 11.9 & \textbf{86.5} \\
\rowcolor{black!8}
75 & 33.3 & \textbf{12.2} & \textbf{11.5} & \textbf{12.2} & 113.3 \\
100 & 32.3 & 11.7 & 10.6 & 11.0 & 133.9 \\
125 & \textbf{33.6} & 11.9 & 10.7 & 11.4 & 145.3 \\
150 & 32.9 & 11.8 & 10.6 & 11.2 & 160.0 \\
\bottomrule
\end{tabular}
\end{adjustbox}
\end{subtable}

\end{table}

For the Gaussian scale ablation in Table \ref{tab:gaussian_scale}, we find that a constant scale of 0.15 across all three axes produces the best results for quality metrics and render speed. This scale is slightly below the voxel radius $0.20m$, allowing sufficient decay outside the voxel and resulting in a more accurate rendering compared with larger scale values.

For the ray sample number $L$, the model must be trained without temporal labels, as using $L>75$ results in an out-of-memory error, highlighting a limitation of voxel-based volume rendering. We observe in Table \ref{tab:voxelnerf_abl} that $L=75$ provides the best balance between model performance and rendering speed. Interestingly, for $L>75$, performance decreases, which differs from the behaviour observed in SQOcc-R, where results show diminishing returns. This may indicate instability during training with Voxel NeRF.

\subsection{Rendering Method Robustness}
To evaluate the robustness of each rendering method, we ablate them under varying image and voxel resolutions, and additionally study Voxel NeRF and SQOcc-R with different ray sample counts $L$.

\begin{table}[htpb]
\caption{\textbf{Render time for varying image and voxel grid resolutions.} We take the best performing fully trained model from the main results section. Default image resolution is set to 256x704, default voxel resolution is set to 200x200x16. OOM denotes out of memory on a 40GB NVIDIA A100 GPU. Metrics are expressed in milliseconds.}
\centering 
\begin{adjustbox}{width=\linewidth}
\begin{tabular}{l|ccc|ccc}
    \toprule
    & \multicolumn{3}{c|}{\textbf{Image Resolution}} 
    & \multicolumn{3}{c}{\textbf{Voxel Resolution}} \\

    \textbf{Method} 
    & \textbf{128x352} 
    & \textbf{256x704} 
    & \textbf{512x1408} 
    & \textbf{200x200x16} 
    & \textbf{300x300x24} 
    & \textbf{400x400x32} \\
    \midrule
    MLGA & 27.0 & 44.0 & 66.9 & 44.0 & 44.0 & 44.0 \\
    Gaussianisation & 14.1 & \textbf{15.5} & \textbf{29.6} & \textbf{15.5} & 48.6 & 95.9 \\
    Voxel NeRF & 56.4 & 102.4 & OOM & 102.4 & 137.2 & 164.5 \\
    SQOcc-R (Ours) & \textbf{10.1} & 17.7 & 47.4 & 17.7 & \textbf{17.7} & \textbf{17.7} \\
    \bottomrule
\end{tabular}
\end{adjustbox}
\label{tab:image_render}
\end{table}



\subsubsection{Image and Voxel Resolution} 
In Table \ref{tab:image_render}, we observe that Voxel NeRF struggles with increasing image resolution, particularly at larger resolutions where it runs out of memory. Both Gaussianisation and SQOcc-R perform well under these conditions, with Gaussianisation achieving better results at larger image resolutions due to efficient screen-space rasterisation. However, our method shows lower rendering time at smaller resolutions. For increased voxel resolutions, both Gaussianisation and Voxel NeRF struggle significantly, especially Gaussianisation, which requires one Gaussian per voxel, highlighting a limitation of the approach. In contrast, SQOcc-R and MLGA maintain consistent rendering time regardless of voxel resolution, as they do not regress to voxel space.

\subsubsection{\texorpdfstring{Ray Sample Number $L$}{Ray Sample Number L}}
In Table \ref{tab:sample_latency_mem}, we compare Voxel NeRF and SQOcc-R across varying ray sample counts $L$ in terms of memory usage and latency. SQOcc-R consistently maintains approximately an 84\% reduction in latency and around an 80\% reduction in memory across all tested values of $L$. Comparing the $400\times400\times32$ voxel resolution used by Gaussianisation with $L=200$ in SQOcc-R provides a rough comparison at higher scene sampling density. In this case, SQOcc-R significantly outperforms in latency, reducing rendering time from 96ms to 26ms. This demonstrates that our method scales more effectively with increasing scene resolution.

\begin{table}[htpb]
\caption{\textbf{Memory and latency comparison.} We take the best performing fully trained model from the main results section and ablate the ray sample number $L$ during inference render time. The best result in each column is highlighted in \textbf{bold}.}
\centering
\begin{adjustbox}{width=\linewidth}
\begin{tabular}{l|l|ccccccc}
\toprule
&
&
\multicolumn{7}{c}{\textbf{Ray Samples $L$}} \\
\textbf{Metric}
& \textbf{Method}
& \textbf{50}
& \textbf{75}
& \textbf{100}
& \textbf{125}
& \textbf{150}
& \textbf{175}
& \textbf{200} \\
\midrule
    \multirow{2}{*}{Memory (GB)} 
        & Voxel NeRF & 7.5 & 11.1 & 14.8 & 18.4 & 22.0 & 25.6 & 29.2 \\
        & SQOcc-R (Ours) & \textbf{1.6} & \textbf{2.2} & \textbf{2.8} & \textbf{3.4} & \textbf{4.1} & \textbf{4.7} & \textbf{5.3} \\
    \midrule
    \multirow{2}{*}{Latency (ms)} 
        & Voxel NeRF & 88.0 & 102.4 & 126.1 & 141.0 & 153.2 & 167.7 & 178.4 \\
        & SQOcc-R (Ours) & \textbf{13.3} & \textbf{15.2} & \textbf{17.2} & \textbf{20.2} & \textbf{22.3} & \textbf{24.5} & \textbf{26.0} \\
    \bottomrule
\end{tabular}
\end{adjustbox}
\label{tab:sample_latency_mem}
\end{table}


\subsection{Scalability of Methods}
Here, we compare the methods when trained with 10,000 superquadric primitives. We observe that Gaussianisation outperforms our method, SQOcc-R, in terms of mIoU and render time. However, it performs worse in RayIoU compared to SQOcc-R on OpenOccv2, which was also observed in the main paper. This suggests limitations in overall scene completion for this method and indicates a saturation point in superquadric count, similar to the behaviour reported in GaussianFlowOcc \cite{boeder2025gaussianflowocc} when the Gaussian count exceeds their saturation point of 10,000.

Voxel NeRF encounters an out-of-memory error (80GB+) due to the already high memory requirements of the method. With increased GPU memory, training may become feasible, as inference does not require the rendering method to be used. MLGA performance decreases significantly, indicating limitations with the render proxy under this configuration.

\begin{table}[htpb]
\setlength{\tabcolsep}{2.8pt}
\caption{\textbf{Training with 10,000 superquadrics.} Voxel NeRF has no reported values because training produced an out-of-memory error. Models are trained with half the training dataset, as done in the ablation studies. The best result overall is highlighted in \textbf{bold}.} 
\centering 
\begin{adjustbox}{width=\linewidth}
\begin{tabular}{l|cc|cc|c|c}
\toprule
&
&
&
\multicolumn{2}{c|}{\textbf{RayIoU}}
&
\textbf{Latency}
&
\textbf{Memory} \\
\textbf{Method}
& \textbf{IoU}
& \textbf{mIoU}
& \textbf{Occ3D}
& \textbf{OpenOccv2}
& \textbf{(ms)}
& \textbf{(GB)} \\
\midrule
    MLGA & 27.9 & 4.6 & 3.2 & 3.7 & 121.4 & 4.4 \\
    Gaussianisation & \textbf{43.4} & \textbf{17.5} & \textbf{16.4} & 16.4 & \textbf{65.3} & 3.7 \\
    Voxel NeRF & - & - & - & - & - & - \\
    SQOcc-R (Ours)  & 42.0 & 15.9 & 16.4 & \textbf{17.0} & 91.2 & \textbf{3.4} \\
    \bottomrule
\end{tabular}
\end{adjustbox}
\label{tab:10k_results}
\end{table}


We note that comparisons with the fully trained 1,600 superquadric models are not strictly fair, as the models in this study are trained using only half of the dataset, following the ablation study setup.
\subsection{Training Information}
In Table \ref{tab:training_information}, we examine the training time along with forward and backward pass times for all four render methods. This timing study enables temporal flow for all renderers only to measure comparable training cost; it is not the MLGA configuration used for final metrics.

\begin{table}[htpb]
\caption{\textbf{Training information.} Training time is measured while training on $4\!\times\!\text{H100 (80GB)}$ for fairness and due to memory constraints issues with Voxel NeRF. Models are trained for 18 epochs. Forward and Backward denote the average time to process each sample during training (in milliseconds).} 
\centering 
\begin{adjustbox}{width=\linewidth}
\begin{tabular}{l|c|ccc}
    \toprule
    \textbf{Method} & \textbf{Parameters} & \textbf{Training Time} & \textbf{Forward} & \textbf{Backward} \\
    \midrule
    MLGA & 35.7M & 1d 20hr & 220 & 150 \\
    Gaussianisation & 35.7M & 1d 20hr & 240 & 0.110 \\
    Voxel NeRF & 35.7M & 2d 5hr & 590 & 865 \\
    SQOcc-R & 35.7M & 2d 0hr & 240 & 270 \\
    \bottomrule
\end{tabular}
\end{adjustbox}
\label{tab:training_information}
\end{table}

We observe that Gaussian-based methods (MLGA, Gaussianisation) require the shortest training time due to the efficiency of Gaussian rasterisation and the implementation provided by the gsplat library \cite{ye2025gsplat}. The volume rendering-based method, Voxel NeRF, requires a noticeably longer training time. SQOcc-R achieves the second fastest forward pass among all approaches, with a backward pass comparable to Gaussian-based methods due to its efficient spatial indexing approach, which greatly reduces rendering cost.

Coupled with efficient training time, SQOcc-R achieves higher performance metrics, which is important for real-time systems, particularly automotive systems that require safety and reliability guarantees \cite{lee2025timing}.
\subsection{Training on the Same GPU Architecture}
In Table \ref{tab:h100_models}, we train all models on the same GPU architecture for fairness: $4\times$ 80GB NVIDIA H100 GPUs. As mentioned in the main paper, we trained only Voxel NeRF on the 80GB NVIDIA H100 GPUs due to memory constraints.
\begin{table}[htpb]
\caption{\textbf{Training on the same GPU architecture.} We train the models again on $4\!\times\!\text{H100 (80GB)}$. The best result overall is highlighted in \textbf{bold}.} 
\centering 
\begin{adjustbox}{width=0.85\linewidth}
\begin{tabular}{l|cc|cc}
\toprule
&
&
&
\multicolumn{2}{c}{\textbf{RayIoU}} \\
\textbf{Method}
& \textbf{IoU}
& \textbf{mIoU}
& \textbf{Occ3D}
& \textbf{OpenOccv2} \\
\midrule
    MLGA & 34.5 & 12.7 & 13.4 & 13.5 \\
    Gaussianisation & \textbf{43.7} & 17.0 & 17.2 & 16.4 \\
    Voxel NeRF & 42.9 & \textbf{17.1} & 18.8 & 18.9 \\
    SQOcc-R (Ours) & 43.3 & 17.0 & \textbf{19.3} & \textbf{19.8} \\
    \bottomrule
\end{tabular}
\end{adjustbox}
\label{tab:h100_models}
\end{table}


Here, we observe trends similar to those reported in the main paper. Gaussianisation, Voxel NeRF, and SQOcc-R show comparable performance across IoU and mIoU. For RayIoU, which is less sensitive to depth estimation errors and penalises over-prediction, SQOcc-R outperforms the other methods, consistent with the results in the main paper. This is particularly evident on OpenOccv2 \cite{tong2023scene}, which provides denser occupancy labels compared to Occ3D-nuScenes \cite{tian2023occ3d}.
\subsection{Self-Supervised Pre-Training}
To further evaluate the effectiveness of our SQOcc-R rendering module, we train SQOcc using ground-truth voxel labels without pre-training for 18 epochs, and then repeat the experiment using pre-trained weights obtained from SQOcc-R.

\begin{table}[htpb]
\caption{\textbf{Pre-training a strongly-supervised model.} We train SuperQuadricOcc with ground-truth labels, and compare it to a model pre-trained with self-supervision from SQOcc-R. The best result overall is highlighted in \textbf{bold}. RayIoU means contain the \textit{others} and \textit{other\_flat} classes because they are present in the ground-truth labels} 
\centering 
\begin{adjustbox}{width=\linewidth}
\begin{tabular}{cc|cc|cc}
    \toprule
    &
    &
    &
    &
    \multicolumn{2}{c}{\textbf{RayIoU}} \\
    \textbf{Supervision}
    & \textbf{Pretraining}
    & \textbf{IoU}
    & \textbf{mIoU}
    & \textbf{Occ3D \cite{tian2023occ3d}}
    & \textbf{OpenOccv2 \cite{tong2023scene}} \\
    \midrule
    Strong & None & 55.3 & 24.4 & 15.6 & 17.7 \\
    Strong & SQOcc-R & \textbf{57.7} & \textbf{29.2} & \textbf{20.8} & \textbf{23.4}
    \\
    \bottomrule
\end{tabular}
\end{adjustbox}
\label{tab:selfsup_pretraining}
\end{table}


We observe a noticeable improvement in model performance, particularly for RayIoU, where there is a 33.3\% increase on the Occ3D-nuScenes dataset and a 32.3\% increase on the OpenOccv2 dataset. This highlights the benefit of using pre-trained weights and suggests that self-supervised occupancy estimation can further support supervised occupancy estimation.
\subsection{SuperQuadric Decay for Neighbourhood Values \texorpdfstring{$\mathcal{N}_v$}{Nv}}
In Table \ref{tab:superquadric_neighborhood_decay}, we examine the value of the superquadric occupancy distribution, $p_o(\mathbf{x}, S)$, for all predicted superquadrics in the validation set for \methodShort{} trained with SQOcc-R. We compute the average value across all superquadrics for three points corresponding to the maximum distances within the cubic neighbourhood: the maximum along one axis ($\mathbf{x}$), along two axes ($\mathbf{x,y}$), and along three axes ($\mathbf{x,y,z}$).

\begin{table}[htpb]
\caption{\textbf{Decay of superquadric occupancy probability.} For each neighbourhood value $\mathcal{N}$, we examine the value of the superquadric occupancy distribution for varying distances.} 
\centering 
\begin{adjustbox}{width=0.6\linewidth}
\begin{tabular}{c|ccc}
    \toprule
    $\mathcal{N}_v$ & \textbf{x} & \textbf{x,y} & \textbf{x,y,z} \\
    \midrule
    1 & 0.3520 & 0.3243 & 0.0959 \\
    2 & 0.2080 & 0.1769 & 0.0294 \\
    3 & 0.1114 & 0.0917 & 0.0107 \\
    4 & 0.0628 & 0.0498 & 0.0042 \\
    5 & 0.0367 & 0.0280 & 0.0017 \\
    6 & 0.0220 & 0.0162 & 0.0007 \\
    7 & 0.0135 & 0.0096 & 0.0003 \\
    \bottomrule
\end{tabular}
\end{adjustbox}
\label{tab:superquadric_neighborhood_decay}
\end{table}

We observe that for smaller neighbourhood sizes, such as $\mathcal{N}=3$, the values remain relatively large, meaning superquadrics outside this region cannot be safely ignored. In contrast, for $\mathcal{N}=5$ there is a substantial reduction in these values, indicating that superquadrics beyond this region have minimal influence. This observation is further supported by the ablation results in Table 5 of the main paper.
\subsection{Temporal Flow Module}
In the main experiments, only MLGA is trained without temporal flow due to severe performance degradation. In \autoref{tab:non_temporal}, for comparison, we train all methods without temporal flow.

\begin{table}[htpb]
\caption{\textbf{Training models without the temporal flow module.} The best result overall is highlighted in \textbf{bold}.} 
\centering 
\begin{adjustbox}{width=0.93\linewidth}
\begin{tabular}{l|cc|cc}
    \toprule
    &
    &
    &
    \multicolumn{2}{c}{\textbf{RayIoU}} \\
    \textbf{Method}
    & \textbf{IoU}
    & \textbf{mIoU}
    & \textbf{Occ3D}
    & \textbf{OpenOccv2} \\
    \midrule
    GaussianFlowOcc \cite{boeder2025gaussianflowocc} & \textbf{35.9} & 12.0 & 11.8 & 12.7 \\
    \midrule
    MLGA & 33.7 & \textbf{12.7} & \textbf{12.1} & \textbf{13.1} \\
    Gaussianisation & 34.5 & 12.4 & 11.6 & 12.2 \\
    Voxel NeRF & 33.0 & 12.0 & 9.1 & 9.8 \\
    SQOcc-R & 34.7 & 12.6 & 11.9 & 12.6 \\
    \bottomrule
\end{tabular}
\end{adjustbox}
\label{tab:non_temporal}
\end{table}

 
MLGA achieves the best performance across the main metrics of mIoU and RayIoU, although SQOcc-R performs closely behind. This can be attributed to the approximation technique fitting well when constrained to fixed viewpoints. However, when rendering from multiple views, the supervision signal degrades, making the method less scalable. Additionally, when increasing the superquadric count, MLGA struggles due to limitations in the render proxy.
\subsection{MLGA versus Single Gaussian Approximation}
To evaluate the effectiveness of the MLGA strategy, which uses a multi-layer Gaussian approximation, we compare it with a baseline that approximates each superquadric using a single Gaussian positioned at its centre.

\begin{table}[htpb]
\caption{\textbf{MLGA versus a single Gaussian render proxy.} The best result overall is highlighted in \textbf{bold}.} 
\centering 
\begin{adjustbox}{width=0.8\linewidth}
\begin{tabular}{l|cc|cc}
    \toprule
    &
    &
    &
    \multicolumn{2}{c}{\textbf{RayIoU}} \\
    \textbf{Approximation}
    & \textbf{IoU}
    & \textbf{mIoU}
    & \textbf{Occ3D}
    & \textbf{OpenOccv2} \\
    \midrule
    Single Gaussian & 31.4 & 8.7 & 8.7 & 9.3 \\
    MLGA & \textbf{33.7} & \textbf{12.7} & 12.1 & 13.1 \\
    \bottomrule
\end{tabular}
\end{adjustbox}
\label{tab:single_gauss}
\end{table}


The results in Table \ref{tab:single_gauss} show that MLGA performs better than a single Gaussian representation, as it more accurately captures the decay of the superquadric occupancy probability. However, as observed in earlier results, this approximation does not provide a sufficiently reliable proxy when rendering from multiple views or when increasing the number of superquadrics.
\subsection{Depth Estimation}
In \autoref{tab:40m_depth}, we re-evaluate our methods for a depth range of $[0.1m, 40m]$, compared to the $[0.1m, 80m]$ range used in the main paper, which was chosen to enable a fair comparison with state-of-the-art methods. Here, the range is restricted to $40m$ as this corresponds to the depth perception range of \methodShort{}.

\begin{table}[htpb]
\caption{\textbf{Depth Estimation performance on nuScenes.} For 40m range.}
\centering
\begin{adjustbox}{width=\linewidth}
\begin{tabular}{l|cc|cc|ccc}
    \toprule
    & \textbf{Abs}
    & \textbf{Sq}
    & \multicolumn{2}{c|}{\textbf{RMSE} }
    & \multicolumn{3}{c}{$\boldsymbol{\delta}$}
    \\
    \textbf{Method}
    & \textbf{Rel} $\downarrow$
    & \textbf{Rel} $\downarrow$
    & $\downarrow$
    & \textbf{log} $\downarrow$
    & $\boldsymbol{<\!\!1.25} \uparrow$
    & $\boldsymbol{<\!\!1.25^2} \uparrow$
    & $\boldsymbol{<\!\!1.25^3} \uparrow$ \\
    \midrule
MLGA & 0.177 & 1.144 & 4.827 & 0.497 & 0.753 & 0.885 & 0.934
\\
Gaussianisation & 0.185 & 1.062 & \textbf{4.365} & 0.291 & 0.750 & 0.888 & 0.946 \\
Voxel NeRF & 0.189 & 1.171 & 4.693 & 0.326 & 0.722 & 0.859 & 0.922 \\
SQOcc-R (Ours) & \textbf{0.171} & \textbf{0.986} & 4.444 & \textbf{0.290} & \textbf{0.758} & \textbf{0.894} & \textbf{0.948} \\
\bottomrule
\end{tabular}
\end{adjustbox}
\label{tab:40m_depth}
\end{table}

We observe similar trends to those reported in the depth results table in the main paper. SQOcc-R outperforms all methods for Abs Rel, the most important depth metric, which measures absolute relative error. Our method is only slightly outperformed in RMSE by Gaussianisation. MLGA displays noteworthy performance due to the lack of regression to voxel space. Overall, SQOcc-R achieves the best results across all metrics except RMSE.

\subsection{Occupancy Classwise Metrics}

\begin{table*}[t]
    \caption{\textbf{State-of-the-art model comparison} on Occ3D-nuScenes \cite{tian2023occ3d}. $^*$ denotes results reproduced locally. Loss denotes the loss space. Rep. denotes the method of scene representation: Gauss - Gaussian; Quad - Superquadric. Mem. denotes the maximum memory (GB) during inference (rendering pipeline not included). All models are self-supervised. Our alternate render methods (MLGA, Gaussianisation, and Voxel NeRF) are included for reference. The best performer between GaussianFlowOcc and \methodShort{} is highlighted in \textbf{bold}.}
  \centering
  \setlength{\tabcolsep}{3pt}
  \begin{adjustbox}{width=\textwidth}
  \begin{tabular}{
    @{}l
    |>{\columncolor{red!3}}c
            >{\columncolor{red!3}}c
    |>{\columncolor{green!3}}c
        >{\columncolor{green!3}}c
    | >{\columncolor{blue!3}}c
        >{\columncolor{blue!3}}c
    |*{15}{c}
    @{}
}
    \toprule
\textbf{Method} & \textbf{Loss} & \textbf{Rep.} & \textbf{Mem.} & \textbf{FPS} & \textbf{IoU} & \textbf{mIoU} & \rotatebox{\myangle}{\textcolor{barrier}{$\blacksquare$} barrier} & \rotatebox{\myangle}{\textcolor{bicycle}{$\blacksquare$} bicycle} & \rotatebox{\myangle}{\textcolor{bus}{$\blacksquare$} bus} & \rotatebox{\myangle}{\textcolor{car}{$\blacksquare$} car} & \rotatebox{\myangle}{\textcolor{construction}{$\blacksquare$} const. veh.} & \rotatebox{\myangle}{\textcolor{motorcycle}{$\blacksquare$} motorcycle} & \rotatebox{\myangle}{\textcolor{pedestrian}{$\blacksquare$} pedestrian} & \rotatebox{\myangle}{\textcolor{cone}{$\blacksquare$} traffic cone} & \rotatebox{\myangle}{\textcolor{trailer}{$\blacksquare$} trailer} & \rotatebox{\myangle}{\textcolor{truck}{$\blacksquare$} truck} & \rotatebox{\myangle}{\textcolor{driveable}{$\blacksquare$} drive. surf.} & \rotatebox{\myangle}{\textcolor{sidewalk}{$\blacksquare$} sidewalk} & \rotatebox{\myangle}{\textcolor{terrain}{$\blacksquare$} terrain} & \rotatebox{\myangle}{\textcolor{manmade}{$\blacksquare$} manmade} & \rotatebox{\myangle}{\textcolor{vegetation}{$\blacksquare$} vegetation} \\
    \midrule 
    \rowcolor{black!8}
    QueryOcc \cite{lilja2026queryocc} & 3D & Query &  & 11.6 & 55.0 & 21.3 & 7.3 & 6.8 & 26.5 & 20.9 & 4.8 & 10.9 & 15.0 & 13.2 & 3.7 & 17.3 & 69.2 & 34.5 & 38.4 & 25.2 & 25.7 \\
        \rowcolor{black!8}
    ShelfOcc \cite{boeder2026shelfocc} & 3D & Voxel &  &  & 56.1 & 22.9 & 14.0 & 11.4 & 25.3 & 25.8 & 7.3 & 16.6 & 12.9 & 13.4 & 5.4 & 17.2 & 68.0 & 34.7 & 42.7 & 25.6 & 22.9 \\
    DistillNeRF \cite{wang2024distillnerf} & 2D & Voxel &  & 2.8 & 29.1 & 10.1 & 1.4 & 2.1 & 10.2 & 10.1 & 2.6 & 2.0 & 5.5 & 4.6 & 1.4 & 7.9 & 43.0 & 16.9 & 15.0 & 14.1 & 15.1 \\
    SelfOcc \cite{huang2024selfocc} & 2D & Voxel & 3.0 & 7.4 & 45.0 & 10.5 & 0.2 & 0.7 & 5.5 & 12.5 & 0.0 & 0.8 & 2.1 & 0.0 & 0.0 & 8.3 & 55.5 & 26.3 & 26.5 & 14.2 & 5.6 \\
    OccNeRF \cite{zhang2023occnerf} & 2D & Voxel & 12.7 & 5.4 & 46.4 & 10.8 & 0.8 & 0.8 & 5.1 & 12.5 & 3.5 & 0.2 & 3.1 & 1.8 & 0.5 & 3.9 & 52.6 & 20.8 & 24.8 & 18.5 & 13.2 \\
    GaussianOcc \cite{gaussianocc} & 2D & Voxel & 11.8 & 5.4 & 42.9 & 11.3 & 1.8 & 5.8 & 14.6 & 13.6 & 1.3 & 2.8 & 8.0 & 9.8 & 0.6 & 9.6 & 44.6 & 20.1 & 17.6 & 8.6 & 10.3 \\
    GaussTR \cite{jiang2025gausstr} & 2D & Gauss & 2.00 & 0.3 & 44.5 & 13.8 & 6.5 & 8.5 & 21.8 & 24.3 & 6.3 & 15.5 & 7.9 & 1.9 & 6.1 & 17.2 & 37.0 & 17.2 & 7.2 & 21.2 & 10.0 \\
    QueryOcc \cite{lilja2026queryocc} & 2D & Query & 11.6 &  & 53.8 & 13.8 \\
    EasyOcc \cite{hayes2025easyocc} & 2D & Voxel & & 5.4 & 38.9 & 15.7 & 1.9 & 6.7 & 15.1 & 21.7 & 2.7 & 8.1 & 15.3 & 11.1 & 1.4 & 12.8 & 55.8 & 27.9 & 22.1 & 16.1 & 17.3 \\
    TT-Occ \cite{zhang2025tt} & 2D & Gauss & 9.9 & 0.7 &  & 16.7 & 21.5 & 10.5 & 10.7 & 14.7 & 11.9 & 12.3 & 9.7 & 12.2 & 4.4 & 7.9 & 48.3 & 23.7 & 28.3 & 14.1 & 20.2 \\
    GaussianFlowOcc \cite{boeder2025gaussianflowocc} & 2D & Gauss & & 10.2 & 46.9 & 17.1 & 6.8 & 9.7 & 19.0 & 17.2 & 4.2 & 11.8 & 9.3 & 10.3 & 1.8 & 12.3 & 61.0 & 31.2 & 34.8 & 14.7 & 12.4 \\
        \midrule
   GaussianFlowOcc$^*$ \cite{boeder2025gaussianflowocc} & 2D & Gauss & 2.9 & 9.6 & 40.5 & 16.7 & 6.7 & \textbf{9.2} & \textbf{17.5} & \textbf{16.5} & \textbf{4.0} & \textbf{11.9} & \textbf{8.8} & \textbf{9.7} & 0.6 & \textbf{12.3} & 49.3 & 30.8 & \textbf{34.6} & \textbf{17.4} & 20.8 
\\

\rowcolor{black!8}
MLGA & 2D & Quad & 0.7 & 21.5 & 33.7 & 12.7 & 5.8 & 5.7 & 16.3 & 13.1 & 3.7 & 5.7 & 3.9 & 3.5 & 2.6 & 11.9 & 43.6 & 20.7 & 24.1 & 14.4 & 15.3
\\

\rowcolor{black!8}
Gaussianisation & 2D & Quad & 0.7 & 21.5 & 43.7 & 17.0 & 6.0 & 8.3 & 14.0 & 12.6 & 4.0 & 9.6 & 8.7 & 10.3 & 1.5 & 10.3 & 64.0 & 32.4 & 35.9 & 16.9 & 20.3
\\

\rowcolor{black!8}
Voxel NeRF & 2D & Quad & 0.7 & 21.5 & 42.9 & 17.1 & 7.6 & 8.5 & 17.2 & 14.9 & 2.0 & 8.7 & 7.6 & 10.5 & 0.1 & 12.5 & 65.7 & 31.4 & 34.5 & 16.3 & 19.2
\\

\rowcolor{blue!10}
\methodShort{} (Ours) & 2D & Quad & \textbf{0.7} & \textbf{21.5} & \textbf{43.5} & \textbf{17.1} & \textbf{7.0} & 8.3 & 16.6 & 14.8 & 3.4 & 9.1 & 8.0 & 8.8 & \textbf{1.1} & 12.0 & \textbf{64.7} & \textbf{31.4} & 34.3 & 16.6 & \textbf{21.0}
    \\
    \bottomrule
  \end{tabular}
  \end{adjustbox}
  \label{tab:sota_occ3d}
\end{table*}
\begin{table*}[htp]
    \caption{\textbf{State-of-the-art model comparison for RayIoU} on Occ3D-nuScenes \cite{tian2023occ3d}. $^\dagger$ - evaluated from released checkpoint. $^*$ - locally trained and evaluated. All models are self-supervised. The best performer is highlighted in \textbf{bold}.}
  \centering
  \setlength{\tabcolsep}{3pt}
  \begin{adjustbox}{width=\textwidth}
  \begin{tabular}{
    @{}l
    | >{\columncolor{blue!5}}c
    |  >{\columncolor{blue!5}}c
        >{\columncolor{blue!5}}c
        >{\columncolor{blue!5}}c
    |*{15}{c}
    @{}
}
    \toprule
\textbf{Model} 
& \rotatebox{\myangle}{\textbf{RayIoU}}
& \rotatebox{\myangle}{\textbf{RayIoU@1}}
& \rotatebox{\myangle}{\textbf{RayIoU@2}}
& \rotatebox{\myangle}{\textbf{RayIoU@4}}
& \rotatebox{\myangle}{\textcolor{barrier}{$\blacksquare$} barrier} & \rotatebox{\myangle}{\textcolor{bicycle}{$\blacksquare$} bicycle} & \rotatebox{\myangle}{\textcolor{bus}{$\blacksquare$} bus} & \rotatebox{\myangle}{\textcolor{car}{$\blacksquare$} car} & \rotatebox{\myangle}{\textcolor{construction}{$\blacksquare$} const. veh.} & \rotatebox{\myangle}{\textcolor{motorcycle}{$\blacksquare$} motorcycle} & \rotatebox{\myangle}{\textcolor{pedestrian}{$\blacksquare$} pedestrian} & \rotatebox{\myangle}{\textcolor{cone}{$\blacksquare$} traffic cone} & \rotatebox{\myangle}{\textcolor{trailer}{$\blacksquare$} trailer} & \rotatebox{\myangle}{\textcolor{truck}{$\blacksquare$} truck} & \rotatebox{\myangle}{\textcolor{driveable}{$\blacksquare$} drive. surf.} & \rotatebox{\myangle}{\textcolor{sidewalk}{$\blacksquare$} sidewalk} & \rotatebox{\myangle}{\textcolor{terrain}{$\blacksquare$} terrain} & \rotatebox{\myangle}{\textcolor{manmade}{$\blacksquare$} manmade} & \rotatebox{\myangle}{\textcolor{vegetation}{$\blacksquare$} vegetation} \\
    \midrule 
    SelfOcc$^\dagger$ \cite{huang2024selfocc} 
& 10.9 & 7.6 & 10.9 & 14.1 & 0.9 & 1.0 & 15.5 & 15.7 & 0.0 & 0.8 & 5.0 & 0.0 & 0.0 & 19.7 & 48.4 & 13.6 & \textbf{17.5} & 16.4 & 8.7 \\

OccNeRF$^\dagger$ \cite{zhang2023occnerf} 
& 11.8 & 7.8 & 11.6 & 16.0 & 2.4 & 2.1 & 28.3 & 21.0 & 4.6 & 1.0 & 5.8 & 7.5 & 0.6 & 16.5 & 39.8 & 12.5 & 11.4 & 13.1 & 10.9 \\

GaussianOcc$^\dagger$ \cite{gaussianocc} 
& 13.4 & 9.8 & 13.5 & 16.9 & 2.6 & 6.6 & 23.6 & 18.0 & 2.0 & 2.5 & 11.9 & 12.9 & 0.7 & 22.5 & 42.3 & 16.3 & 14.5 & 11.7 & 13.3 \\

GaussTR$^\dagger$ \cite{jiang2025gausstr} 
& 14.6 & 10.6 & 14.7 & 18.5 & 9.6 & \textbf{12.2} & 36.4 & 29.7 & \textbf{8.4} & \textbf{15.0} & 16.8 & 3.0 & \textbf{4.6} & 19.5 & 20.8 & 5.9 & 3.6 & 21.0 & 12.7 \\

TT-Occ \cite{zhang2025tt} 
& 15.2 & 11.3 & 14.6 & 17.8 & \\

EasyOcc \cite{hayes2025easyocc} 
& 16.5 & 12.3 & 16.7 & 20.6 & 3.0 & 4.6 & 21.9 & 27.7 & 4.5 & 6.4 & \textbf{20.7} & 12.7 & 2.5 & 26.3 & \textbf{50.3} & \textbf{17.6} & 14.4 & 15.3 & 20.2 \\

GaussianFlowOcc$^*$ \cite{boeder2025gaussianflowocc} 
& 18.3 & 12.6 & 18.3 & 24.0 & 12.2 & 7.0 & \textbf{38.8} & 29.9 & 5.3 & 10.9 & 16.1 & 13.4 & 0.1 & 28.1 & 36.0 & 13.9 & 15.3 & \textbf{24.6} & 23.1 \\
\midrule

SQOcc (Ours) 
& \textbf{19.6} & \textbf{13.5} & \textbf{19.6} & \textbf{25.6} & \textbf{14.3} & 8.8 & 38.0 & \textbf{31.2} & 4.9 & 8.6 & 18.1 & \textbf{15.1} & 0.3 & \textbf{30.7} & 45.9 & 14.3 & 15.2 & 24.1 & \textbf{23.9} \\
    \bottomrule
  \end{tabular}
  \end{adjustbox}
  \label{tab:rayiou_class_SOTA_occ3d}
\end{table*}

\begin{table*}[htp]
    \caption{\textbf{State-of-the-art model comparison for RayIoU} on OpenOccv2 \cite{tong2023scene}. $^\dagger$ - evaluated from released checkpoint. $^*$ - locally trained and evaluated. All models are self-supervised. Our alternate render methods (MLGA, Gaussianisation, and Voxel NeRF) are included for reference. The best performer between GaussianFlowOcc and SQOcc is highlighted in \textbf{bold}.}
  \centering
  \setlength{\tabcolsep}{3pt}
  \begin{adjustbox}{width=\textwidth}
  \begin{tabular}{
    @{}l
    | >{\columncolor{blue!5}}c
    |  >{\columncolor{blue!5}}c
        >{\columncolor{blue!5}}c
        >{\columncolor{blue!5}}c
        |>{\columncolor{blue!5}}c
    |*{15}{c}
    @{}
}
    \toprule
\textbf{Method} 
& \rotatebox{\myangle}{\textbf{RayIoU}}
& \rotatebox{\myangle}{\textbf{RayIoU@1}}
& \rotatebox{\myangle}{\textbf{RayIoU@2}}
& \rotatebox{\myangle}{\textbf{RayIoU@4}}
& \rotatebox{\myangle}{\textbf{Dynamic}}
& \rotatebox{\myangle}{\textcolor{barrier}{$\blacksquare$} barrier} & \rotatebox{\myangle}{\textcolor{bicycle}{$\blacksquare$} bicycle} & \rotatebox{\myangle}{\textcolor{bus}{$\blacksquare$} bus} & \rotatebox{\myangle}{\textcolor{car}{$\blacksquare$} car} & \rotatebox{\myangle}{\textcolor{construction}{$\blacksquare$} const. veh.} & \rotatebox{\myangle}{\textcolor{motorcycle}{$\blacksquare$} motorcycle} & \rotatebox{\myangle}{\textcolor{pedestrian}{$\blacksquare$} pedestrian} & \rotatebox{\myangle}{\textcolor{cone}{$\blacksquare$} traffic cone} & \rotatebox{\myangle}{\textcolor{trailer}{$\blacksquare$} trailer} & \rotatebox{\myangle}{\textcolor{truck}{$\blacksquare$} truck} & \rotatebox{\myangle}{\textcolor{driveable}{$\blacksquare$} drive. surf.} & \rotatebox{\myangle}{\textcolor{sidewalk}{$\blacksquare$} sidewalk} & \rotatebox{\myangle}{\textcolor{terrain}{$\blacksquare$} terrain} & \rotatebox{\myangle}{\textcolor{manmade}{$\blacksquare$} manmade} & \rotatebox{\myangle}{\textcolor{vegetation}{$\blacksquare$} vegetation} \\
    \midrule 

SelfOcc$^\dagger$ \cite{huang2024selfocc} 
& 9.7 & 6.2 & 9.6 & 13.1 & 6.8 & 0.8 & 0.9 & 14.0 & 14.7 & 0.0 & 0.8 & 4.8 & 0.0 & 0.0 & 18.9 & 39.4 & 12.1 & 14.2 & 16.2 & 8.1 \\

OccNeRF$^\dagger$ \cite{zhang2023occnerf} 
& 12.1 & 8.4 & 12.1 & 16.0 & 10.0 & 2.1 & 1.7 & 29.0 & 21.0 & 4.6 & 0.9 & 5.1 & 6.2 & 0.6 & 16.8 & 42.1 & 14.2 & 14.4 & 12.9 & 10.5 \\

GaussianOcc$^\dagger$ \cite{gaussianocc} 
& 12.5 & 9.1 & 12.6 & 15.8 & 10.3 & 2.4 & 5.4 & 21.9 & 17.4 & 2.1 & 2.2 & 10.8 & 10.7 & 0.6 & 22.2 & 38.3 & 15.1 & 14.0 & 11.3 & 12.8 \\

GaussTR$^\dagger$ \cite{jiang2025gausstr} 
& 14.9 & 10.9 & 15.1 & 18.8 & 17.5 & 9.2 & 11.8 & 34.8 & 29.6 & 8.4 & 15.2 & 16.5 & 2.5 & 4.9 & 19.1 & 24.9 & 8.1 & 4.8 & 21.2 & 12.7 \\

EasyOcc \cite{hayes2025easyocc} 
& 16.2 & 12.1 & 16.4 & 20.0 & 13.7 & 2.8 & 3.8 & 20.8 & 26.4 & 4.7 & 5.8 & 19.9 & 11.0 & 3.1 & 25.3 & 47.2 & 17.1 & 18.4 & 15.4 & 20.8 \\ 

\midrule

GaussianFlowOcc$^*$ \cite{boeder2025gaussianflowocc} 
& 18.2 & 12.9 & 18.4 & 23.4 & 17.0 & 11.1 & 6.2 & \textbf{38.7} & 29.5 & \textbf{5.8} & \textbf{11.1} & 15.5 & 12.0 & 0.2 & 29.0 & 32.9 & 15.4 & 17.1 & \textbf{25.3} & 23.5 \\

\rowcolor{black!8}
MLGA & 13.1 & 8.1 & 12.9 & 18.2 & 13.4 & 8.3 & 6.0 & 31.1 & 24.2 & 4.6 & 5.5 & 9.5 & 5.0 & 1.3 & 25.3 & 23.8 & 7.7 & 10.7 & 17.0 & 16.1 \\

\rowcolor{black!8}
Gaussianisation & 16.6 & 11.3 & 16.6 & 22.0 & 13.2 & 10.9 & 6.8 & 26.6 & 19.6 & 5.6 & 6.9 & 14.6 & 12.5 & 0.3 & 25.6 & 42.0 & 14.3 & 17.2 & 23.0 & 23.4 \\

\rowcolor{black!8}
Voxel NeRF & 18.9 & 13.5 & 19.1 & 24.2 & 15.9 & 14.8 & 6.4 & 35.1 & 29.7 & 2.9 & 7.0 & 16.5 & 16.1 & 0.0 & 29.5 & 44.7 & 17.1 & 18.5 & 23.6 & 21.7 \\

\rowcolor{blue!10}
SQOcc (Ours) 
& \textbf{19.9} & \textbf{14.2} & \textbf{20.1} & \textbf{25.4} & \textbf{17.6} & \textbf{13.4} & \textbf{8.3} & 38.2 & \textbf{31.1} & 5.0 & 8.6 & \textbf{17.8} & \textbf{15.0} & \textbf{0.3} & \textbf{31.4} & \textbf{45.0} & \textbf{17.5} & \textbf{18.8} & 24.7 & \textbf{24.0} \\

    \bottomrule
  \end{tabular}
  \end{adjustbox}
  \label{tab:rayiou_class_openocc}
\end{table*}

In Table \ref{tab:sota_occ3d}, we provide class-wise metrics for the state-of-the-art methods alongside our rendering methods (MLGA, Gaussianisation, Voxel NeRF). In terms of our rendering methods, all models, except MLGA, report relatively similar scores, with SQOcc-R providing the most consistent performance overall, falling only slightly behind the best-performing method for individual classes. For example, for the vulnerable bicycle class, SQOcc-R is only 2\% below the top result. In comparison to other state-of-the-art methods, namely GaussianFlowOcc \cite{boeder2025gaussianflowocc}, our method is outperformed for multiple classes, namely dynamic classes such as car and pedestrian, although we maintain competitive results. This can be attributed to overly dense predictions due to the greater number of primitives deployed, and mIoU does not penalise overly dense predictions \cite{tang2024sparseocc}. This is further evidenced when examining RayIoU per class, for which our method outperforms.

In Tables \ref{tab:rayiou_class_SOTA_occ3d} and  \ref{tab:rayiou_class_openocc}, SQOcc-R outperforms all our other render methods (MLGA, Gaussianisation, Voxel NeRF) in RayIoU across all distance thresholds. For class-wise metrics, it also frequently achieves substantially higher scores than competing methods. For instance, in the bicycle class, SQOcc-R outperforms the next best method by 25\% on both datasets. Overall, SQOcc-R demonstrates superior detection performance. In comparison to state-of-the-art methods, SQOcc demonstrates superior performance, better segmenting dynamic classes such as bicycle and pedestrian, demonstrating the superior capability of superquadrics as a scene representation choice.

\subsection{Downstream Tasks Timewise Metrics}
In \autoref{tab:occ3d_forecasting_iou} and \autoref{tab:occ3d_forecasting_rayiou}, we report the complete timewise occupancy forecasting results for IoU, mIoU, and RayIoU. \methodShort{} consistently outperforms GaussianFlowOcc across all forecast horizons and occupancy metrics. Notably, \methodShort{} maintains a clear advantage in RayIoU at the 3s horizon, demonstrating the effectiveness of sparse scene representations, such as superquadrics, for long-term occupancy forecasting.

In \autoref{tab:occ3d_forecasting_trajectory}, we report time-wise trajectory estimation results using the L2 and collision rate (CR) metrics. \methodShort{} again outperforms GaussianFlowOcc across all evaluated metrics and horizons. In particular, at the 3s horizon, \methodShort{} achieves a substantial reduction in collision rate (-34\%), indicating that compact superquadric representations improve long-term trajectory estimation and may provide practical benefits for real-world deployment.

\begin{table*}[htpb]
\caption{\textbf{Downstream tasks.} $^\dagger$ - copying the current occupancy as future estimations (no forecasting).
$^*$ - locally trained and evaluated.
Sup. denotes supervision type.
FPS is the time from camera input to all occupancy forecasts (1s, 2s, and 3s) and trajectory estimations if applicable.
Metrics are averages over forecasting for 1s, 2s, and 3s.
The best result overall is highlighted in \textbf{bold}.}
\centering

\begin{subtable}{\linewidth}
\centering
\begin{adjustbox}{width=0.9\linewidth}
\begin{tabular}{l|c|c|cccc|cccc}
\toprule
&
&
&
\multicolumn{4}{c|}{\textbf{IoU}}
&
\multicolumn{4}{c}{\textbf{mIoU}} \\
\textbf{Method}
& \textbf{Sup.}
& \textbf{FPS}
& \textbf{1s}
& \textbf{2s}
& \textbf{3s}
& \textbf{Avg.}
& \textbf{1s}
& \textbf{2s}
& \textbf{3s}
& \textbf{Avg.} \\
\midrule
    \rowcolor{black!8}
    OccWorld \cite{zheng2024occworld}
    & Strong & 2.8
    & 18.9 & 16.3 & 14.4 & 16.5
    & 11.6 & 8.1 & 6.2 & 8.6 \\

    \rowcolor{black!8}
    OccLLaMa \cite{wei2024occllama}
    & Strong & --
    & 25.8 & 23.2 & 20.0 & 23.0
    & 10.3 & 8.7 & 7.0 & 8.7 \\

    \rowcolor{black!8}
    Occ-LLM \cite{xu2025occ}
    & Strong & --
    & 27.1 & 24.1 & 20.2 & 23.8
    & 11.3 & 10.2 & 9.1 & 10.2 \\

    \rowcolor{black!8}
    DOME-F \cite{gu2024dome}
    & Strong & --
    & 35.2 & 27.9 & 23.4 & 28.8
    & 24.1 & 17.4 & 13.2 & 18.3 \\

    \rowcolor{black!8}
    COME \cite{shi2026come}
    & Strong & --
    & 48.1 & 43.8 & 40.3 & 44.1
    & 26.6 & 21.7 & 18.5 & 22.3 \\

    \rowcolor{black!8}
    SparseWorld-TC \cite{du2026sparseworld}
    & Strong & 2.1
    & 55.3 & 53.6 & 51.7 & 53.5
    & 32.8 & 29.6 & 27.3 & 29.9 \\

    \midrule
    \rowcolor{black!8}
    GaussianFlowOcc$^\dagger$ \cite{boeder2025gaussianflowocc}
    & Self & 9.6 & 35.9 & 33.3 & 31.7 & 33.6 & 10.6 & 8.7 & 7.8 & 9.3 \\
    \rowcolor{black!8}
    \methodShort{}$^\dagger$ (Ours)
    & Self & 21.5 & 39.6 & 36.9 & 35.2 & 37.2 & 11.6 & 9.7 & 8.7 & 10.0 \\

    \midrule

    OccWorld \cite{zheng2024occworld}
    & Self & 2.8
    & 5.1 & 5.0 & 5.0 & 5.0
    & 0.3 & 0.3 & 0.2 & 0.3 \\

    GaussianFlowOcc \cite{boeder2025gaussianflowocc}
    & Self & 7.9
    & 38.5 & 37.9 & 36.6 & 37.7
    & 14.8 & 13.8 & 12.7 & 13.8 \\
\rowcolor{blue!10}
    \methodShort{} (Ours)
    & Self & \textbf{17.9}
    & \textbf{41.6} & \textbf{39.4} & \textbf{36.9} & \textbf{39.3}
    & \textbf{15.4} & \textbf{14.0} & \textbf{12.8} & \textbf{14.1} \\
\bottomrule
\end{tabular}
\end{adjustbox}
\caption{\textbf{Occupancy forecasting.} Standard occupancy metrics.}
\label{tab:occ3d_forecasting_iou}
\end{subtable}

\vspace{0.5em}

\begin{subtable}{\linewidth}
\centering
\begin{adjustbox}{width=0.85\linewidth}
\begin{tabular}{l|c|c|cccc|cccc}
\toprule
&
&
&
\multicolumn{4}{c|}{\textbf{RayIoU Occ3D}}
&
\multicolumn{4}{c}{\textbf{RayIoU OpenOccv2}} \\
\textbf{Method}
& \textbf{Sup.}
& \textbf{FPS}
& \textbf{1s}
& \textbf{2s}
& \textbf{3s}
& \textbf{Avg.}
& \textbf{1s}
& \textbf{2s}
& \textbf{3s}
& \textbf{Avg.} \\
\midrule

  \rowcolor{black!8}
    GaussianFlowOcc$^\dagger$ \cite{boeder2025gaussianflowocc}
    & Self & 9.6 & 11.8 & 9.5 & 8.4 & 10.3 & 11.7 & 9.4 & 8.3 & 9.8\\
  \rowcolor{black!8}
    \methodShort{}$^\dagger$ (Ours)
    & Self & 21.5 & 13.1 & 10.6 & 9.4 & 11.0 & 13.1 & 10.6 & 9.4 & 11.0 \\

\midrule

    GaussianFlowOcc \cite{boeder2025gaussianflowocc}
    & Self & 7.9 & 15.5 & 13.6 & 12.1 & 13.7 & 15.6 & 13.8 & 12.4 & 13.9 \\
\rowcolor{blue!10}
    \methodShort{} (Ours)
    & Self & \textbf{17.9} & \textbf{16.5} & \textbf{14.2} & \textbf{12.5} & \textbf{14.4} & \textbf{16.7} & \textbf{14.5} & \textbf{12.7} & \textbf{14.6} \\
\bottomrule
\end{tabular}
\end{adjustbox}
\caption{\textbf{Occupancy forecasting.} RayIoU metrics.}
\label{tab:occ3d_forecasting_rayiou}
\end{subtable}

\vspace{0.5em}

\begin{subtable}{\linewidth}
\centering
\begin{adjustbox}{width=\linewidth}
\begin{tabular}{l|c|c|c|cccc|cccc}
\toprule
&
&
&
&
\multicolumn{4}{c|}{\textbf{L2 (m) $\downarrow$}}
&
\multicolumn{4}{c}{\textbf{CR(\%) $\downarrow$}} \\
\textbf{Method}
& \textbf{Sup.}
& \textbf{Ego Traj.}
& \textbf{FPS}
& \textbf{1s}
& \textbf{2s}
& \textbf{3s}
& \textbf{Avg.}
& \textbf{1s}
& \textbf{2s}
& \textbf{3s}
& \textbf{Avg.} \\
\midrule

\rowcolor{black!8}
ST-P3 \cite{hu2022st}
& Strong & \xmark & 
& 1.59 & 2.64 & 3.73 & 2.65
& 0.69 & 3.62 & 8.39 & 4.23 \\

\rowcolor{black!8}
OccWorld \cite{zheng2024occworld}
& Strong & \xmark & 2.8
& 0.52 & 1.27 & 2.41 & 1.40
& 0.12 & 0.40 & 2.08 & 0.87 \\

\rowcolor{black!8}
OccLLaMa \cite{wei2024occllama}
& Strong & \xmark & 
& 0.38 & 1.07 & 2.15 & 1.20
& 0.06 & 0.39 & 1.65 & 0.70 \\

\rowcolor{black!8}
GaussianAD \cite{zheng2024gaussianad}
& Strong & \xmark & 
& 0.40 & 0.64 & 0.88 & 0.64
& 0.09 & 0.38 & 0.81 & 0.42 \\

\rowcolor{black!8}
Uni-AD \cite{hu2023planning}
& Strong & \checkmark & 
& 0.20 & 0.42 & 0.75 & 0.46
& 0.02 & 0.25 & 0.84 & 0.37 \\

\rowcolor{black!8}
BEV-Planner \cite{li2024ego}
& Self & \checkmark & 
& 0.16 & 0.32 & 0.57 & 0.35
& 0.04 & 0.29 & 0.94 & 0.42 \\

\midrule

\rowcolor{black!8}
ShelfGaussian \cite{zhao2026shelfgaussian}
& Self$^*$ & \xmark & 
& 1.05 & 1.76 & 2.50 & 1.77
& 0.31 & 2.38 & 5.31 & 2.67 \\

\rowcolor{black!8}
ShelfGaussian \cite{zhao2026shelfgaussian}
& Self$^*$ & \checkmark & 
& 0.16 & 0.32 & 0.58 & 0.35
& 0.00 & 0.18 & 0.78 & \textbf{0.32} \\

\midrule

GaussianFlowOcc$^*$ \cite{boeder2025gaussianflowocc}
& Self & \xmark & 7.9
& 1.13 & 2.37 & 3.73 & 2.41
& 1.21 & 2.42 & 4.43 & 2.69 \\

\rowcolor{blue!10}
\methodShort{} (Ours)
& Self & \xmark & \textbf{17.9}
& \textbf{0.98} & \textbf{2.14} & \textbf{3.55} & \textbf{2.22}
& \textbf{1.09} & \textbf{1.75} & \textbf{2.89} & \textbf{1.91} \\

\bottomrule
\end{tabular}
\end{adjustbox}
\caption{\textbf{Trajectory estimation.} Self$^*$ denotes the use of LiDAR for supervision. Ego Traj. denotes models that use ego trajectory.}
\label{tab:occ3d_forecasting_trajectory}
\end{subtable}

\label{tab:occ3d_forecasting}
\end{table*}
\subsection{SuperQuadric Parameter Distribution}
\begin{figure*}
    \centering
    \includegraphics[width=\textwidth]{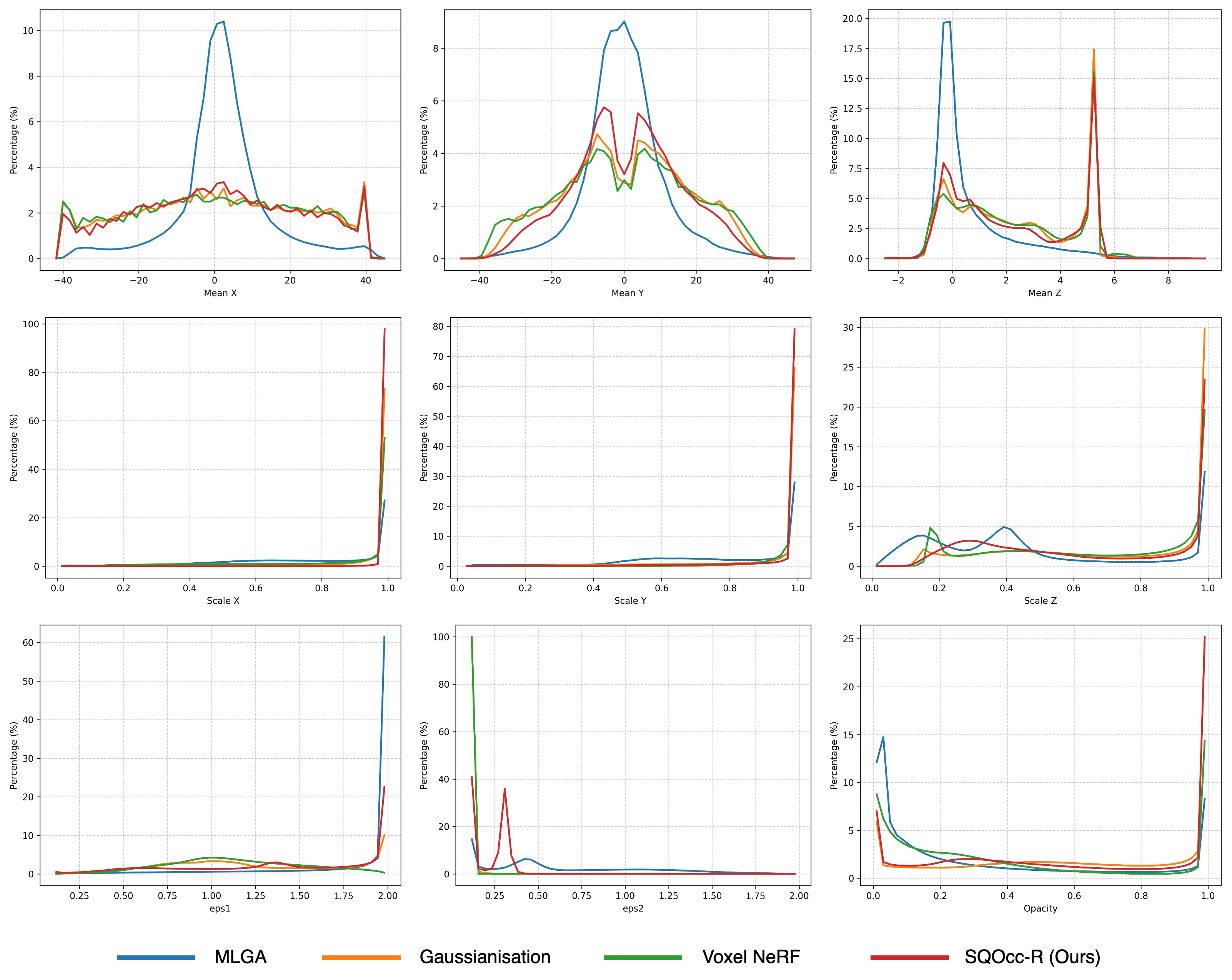}
     \caption{\textbf{Parameter comparison for all models.} Best viewed zoomed in.}
    \label{fig:param}
\end{figure*}

In Figure \ref{fig:param}, we compare the distribution of the superquadric parameters: mean $\mathbf{m}$, scale $\mathbf{s}$, epsilon $\boldsymbol{\varepsilon}$, and opacity $\sigma$ for SQOcc trained with different rendering supervisions: MLGA, Gaussianisation, Voxel NeRF, and SQOcc-R.

\textbf{Mean $\mathbf{m}$:} For the mean, models trained with the temporal flow module exhibit similar distributions, with superquadrics spread throughout the voxel grid. For MLGA, the $xy$ mean shows superquadrics concentrated closer to the origin, likely due to insufficient supervision in distant regions caused by the absence of temporal flow. For the $z$ mean, a spike around $5m$ appears for models using temporal flow, which may result from the larger region covered by supervision.

\textbf{Scale $\mathbf{s}$:} For scale, all models show a similar distribution for the $xy$ scale, generally approaching the maximum value of 1 to increase coverage. The $z$ scale shows a slightly broader spread, possibly reflecting the need to model flatter structures such as roads and sidewalks.

\textbf{Epsilon $\boldsymbol{\varepsilon}$:} For epsilon, the average values are approximately $(\varepsilon_1, \varepsilon_2) = (2.0, 0.2)$, corresponding to a diamond-like shape. MLGA and SQOcc-R display slightly more diverse representations, particularly for $\varepsilon_2$, suggesting that regression to voxel space may produce less varied superquadric shapes.

\textbf{Opacity $\sigma$:} For opacity, strong peaks appear at $\sigma=0$ and $\sigma=1$, indicating that the models tend to favour near-binary opacity decisions. SQOcc-R and Gaussianisation exhibit similar distributions, while MLGA and Voxel NeRF show slightly more variation at lower opacity values. 

\section{Supplementary Visualisations}
\label{sec:supp_vis}
In \autoref{subsec:vis_sqocc}, we provide additional semantic occupancy and depth estimation visualisations comparing GaussianFlowOcc and SuperQuadricOcc (SQOcc) (across our four rendering methods). In \autoref{subsec:sqocc-r} we present additional visualisations for \methodShort{} and depth visualisation results for \renderShort{}. Furthermore, we provide a visualisation video comparing \methodShort{} to the ground truth labels for various scenes attached.

\subsection{SuperQuadricOcc Comparisons}
\label{subsec:vis_sqocc}
Here, we provide additional qualitative visualisations of semantic occupancy and depth estimation, comparing GaussianFlowOcc with \method{} across all four rendering strategies.

\subsubsection{Semantic Occupancy}

\begin{figure*}[t]
    \centering
    \includegraphics[width=\textwidth]{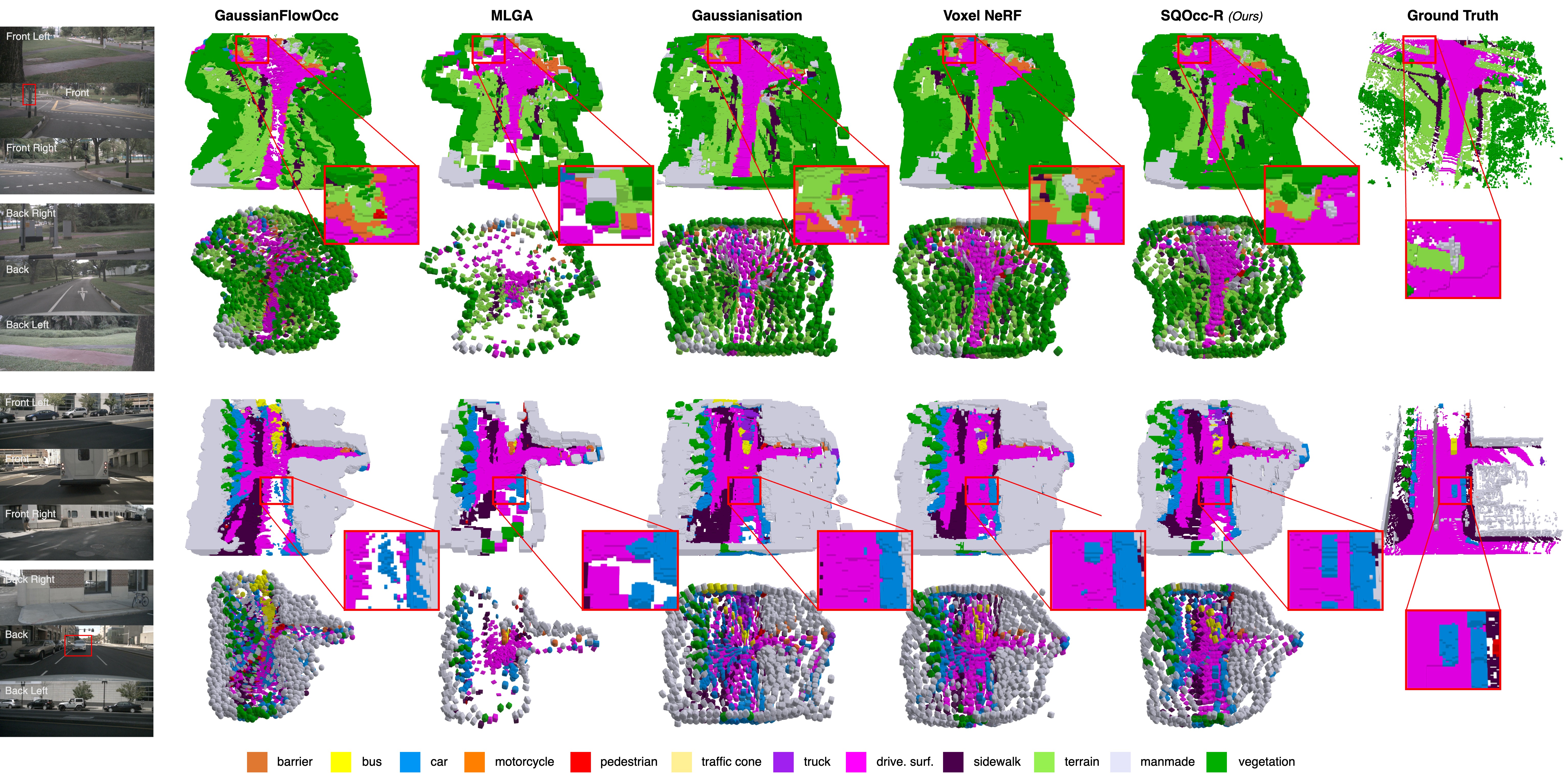}
     \caption{\textbf{Semantic occupancy visual analysis 1.} Best viewed zoomed in.}
     \vspace{-3mm}
    \label{fig:visual_voxel}
\end{figure*}

In \autoref{fig:visual_voxel}, we examine that in the top scene, several models struggle to reconstruct a pole visible in the front camera image. \renderShort{} predicts a single, coherent pole structure, whereas GaussianFlowOcc and Gaussianisation under-predict the object, and MLGA and Voxel NeRF over-predict it. This observation reflects the challenges self-supervised models face when reconstructing small objects under occlusion. For the bottom scene, \renderShort{} correctly reconstructs the full car and surrounding vehicles, with GaussianFlowOcc notably struggling in this area.

\begin{figure*}
    \centering
    \includegraphics[width=\textwidth]{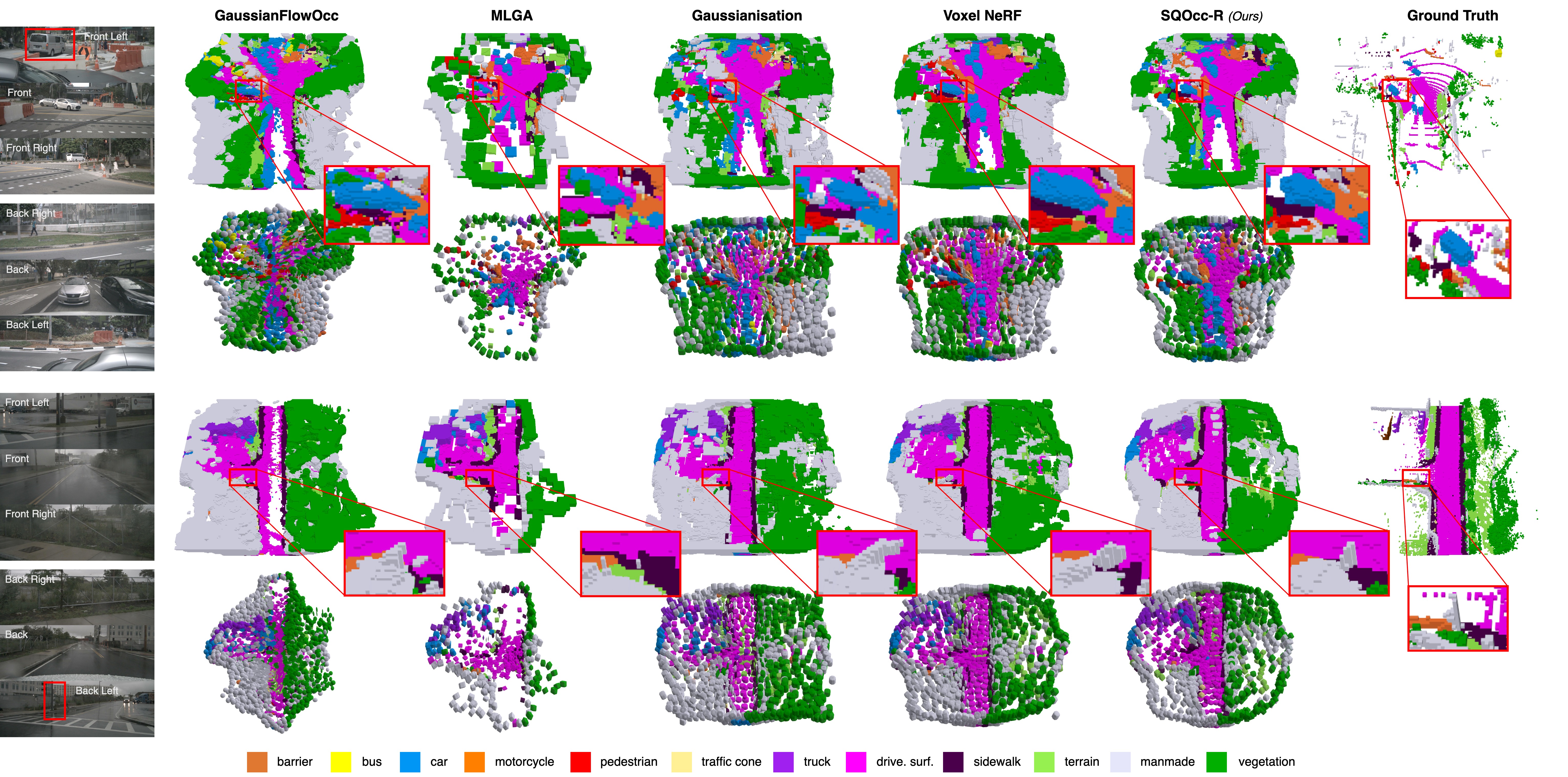}
     \caption{\textbf{Semantic occupancy visual analysis 2.} Best viewed zoomed in.}
    \label{fig:voxel_comp}
\end{figure*}

In \autoref{fig:voxel_comp}, we examine a region in the front-left image of the top scene, which contains several objects, including a vehicle and multiple construction elements. Due to this cluttered environment, many models struggle to reconstruct the vehicle correctly, often producing over-predictions (GaussianFlowOcc, Gaussianisation). In contrast, SQOcc-R reconstructs the vehicle more accurately. It is also important to note that the ground-truth labels for this scene contain inaccuracies, with significant misclassification of certain areas, particularly the road, indicating the need for improvement in labelling pipelines. In the bottom scene, the view is partially occluded by rain in front of the camera. Despite this, SQOcc-R reconstructs the pole in the back-left image more accurately than the other models. However, all models exhibit some degree of over-prediction, indicating that challenging conditions still lead to degraded performance.

\subsubsection{Depth Estimation}
\begin{figure*}
    \centering
    \includegraphics[width=\textwidth]{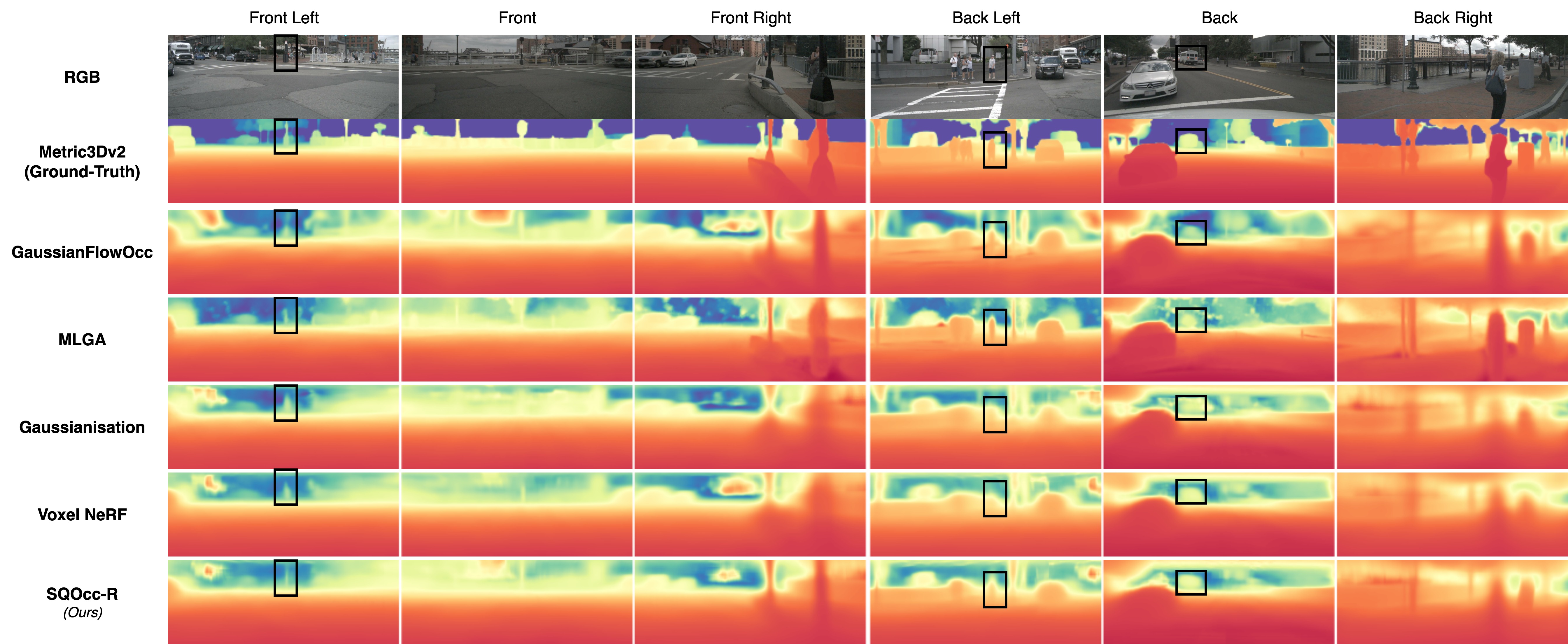}
     \caption{\textbf{Depth estimation comparison 1.} Boxes are overlaid for discussion. Best viewed zoomed in.}
    \label{fig:visual_depth}
\end{figure*}

In \autoref{fig:visual_depth}, we consider a pole in the front left camera view. \renderShort{} produces a clearer and more complete reconstruction, while other methods capture the base of the pole but struggle to reconstruct its thinner upper structure. The single pedestrian in the back-left image is most clearly captured by GaussianFlowOcc, MLGA, and \renderShort{}, all of which render the scene directly, unlike methods that regress to voxel space. The bus in the back camera view is more clearly defined by SQOcc-R and Voxel NeRF. These results highlight the effectiveness of our direct rendering strategy in \renderShort{}.

\begin{figure*}
    \centering
    \includegraphics[width=\textwidth]{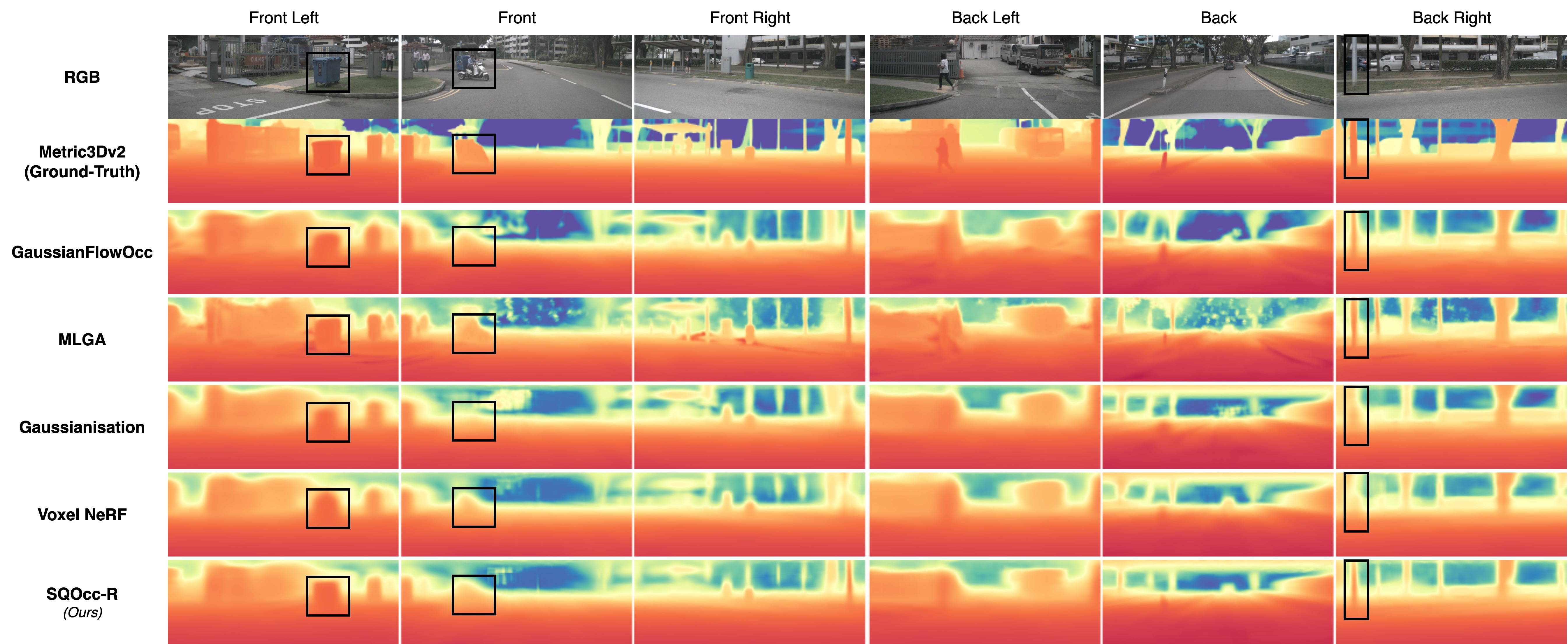}
     \caption{\textbf{Depth estimation comparison 2.} Boxes are overlaid for discussion. Best viewed zoomed in.}
    \label{fig:depth_supp}
\end{figure*}

In \autoref{fig:depth_supp}, we present a similar analysis for depth estimation. In the front-left image, MLGA and SQOcc-R correctly reconstruct the rigid structure of the trash can, while the other methods produce less precise results. Additionally, the motorcyclist in the front image is missing in Gaussianisation, whereas MLGA provides the most accurate reconstruction of the shape. In the back-right image, methods that rely on a voxel proxy (Gaussianisation and Voxel NeRF) struggle to reconstruct the thin pole, suggesting that such proxies can lead to fidelity loss, particularly for small or thin objects.

\subsection{Downstream Tasks}
\label{subsec:vis_downstream}

In this section, we present qualitative results for occupancy forecasting and trajectory estimation, comparing \method{} with GaussianFlowOcc.

\subsubsection{Occupancy Forecasting}
\begin{figure*}
    \centering
    \includegraphics[width=\textwidth]{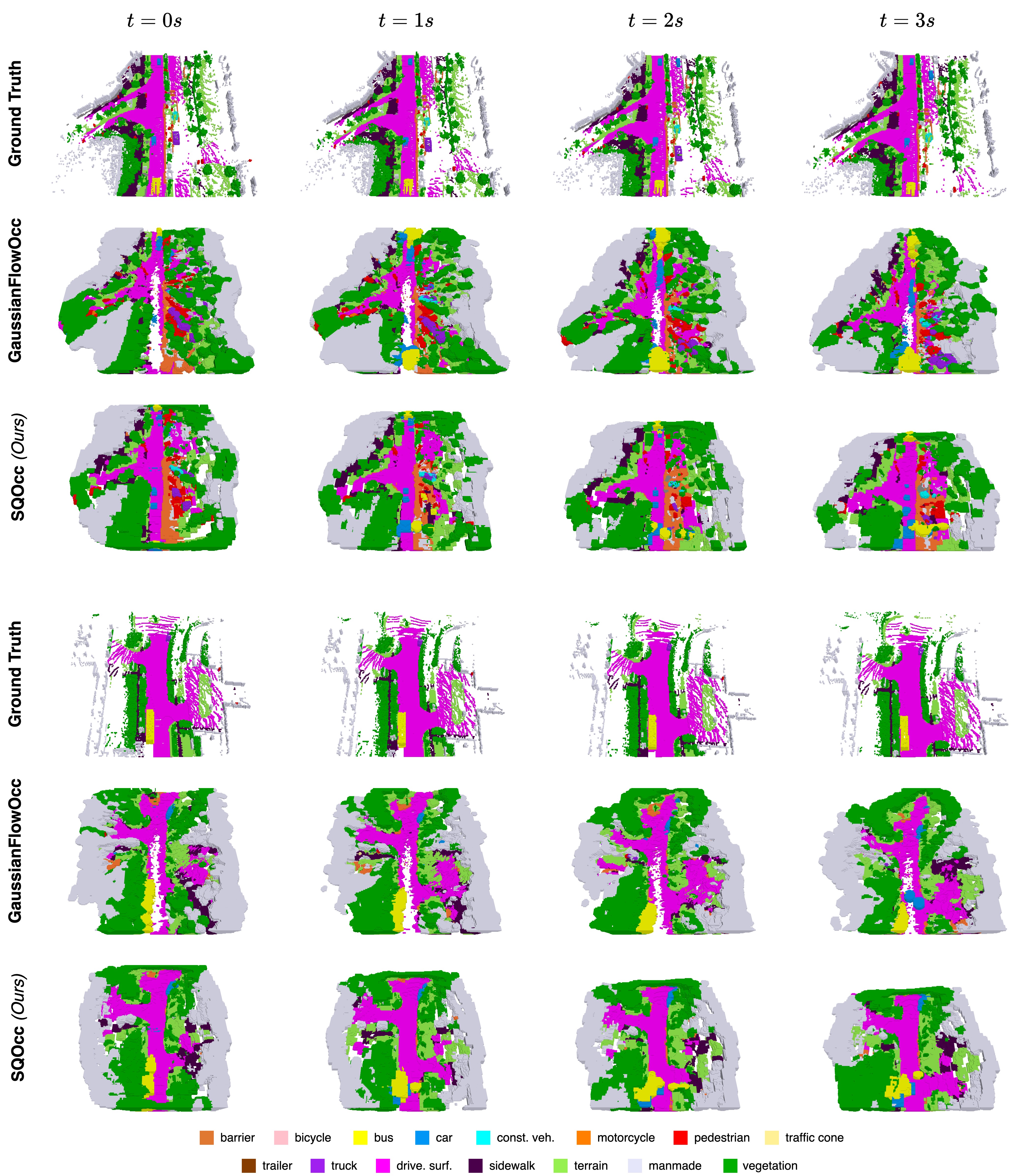}
     \caption{\textbf{Occupancy forecasting} for two separate scenes. Best viewed zoomed in.}
    \label{fig:forecasting_vis}
\end{figure*}

In \autoref{fig:forecasting_vis}, observe that \method{} produces better occupancy forecasts than GaussianFlowOcc. However, one limitation of \methodShort{} is far-region static completion. This is caused by the supervision used in \renderShort(), where the sampling range is set to $[t_{\text{near}}, t_{\text{far}}]=[0.1, 40.0]$, matching the perception range of the dataset. Thus, static regions not reconstructed at time $T$ cannot be introduced by temporal propagation alone. This is not a failure of the temporal flow module in our method. Increasing the sampling range would allow \methodShort{} to complete farther regions. A more robust solution would be to introduce a dedicated far-region completion module for occupancy forecasting, which is valid for both \methodShort{} and GaussianFlowOcc.

In contrast, GaussianFlowOcc is not constrained by this range in the same way, since Gaussian Splatting does not explicitly limit the perception range. Nevertheless, \methodShort{} achieves stronger forecasting metrics, suggesting that its compact superquadric representation better preserves temporally useful scene structure within the evaluated range.

\subsubsection{Trajectory Estimation}

\begin{figure*}[t]
    \centering
    \includegraphics[width=0.9\textwidth]{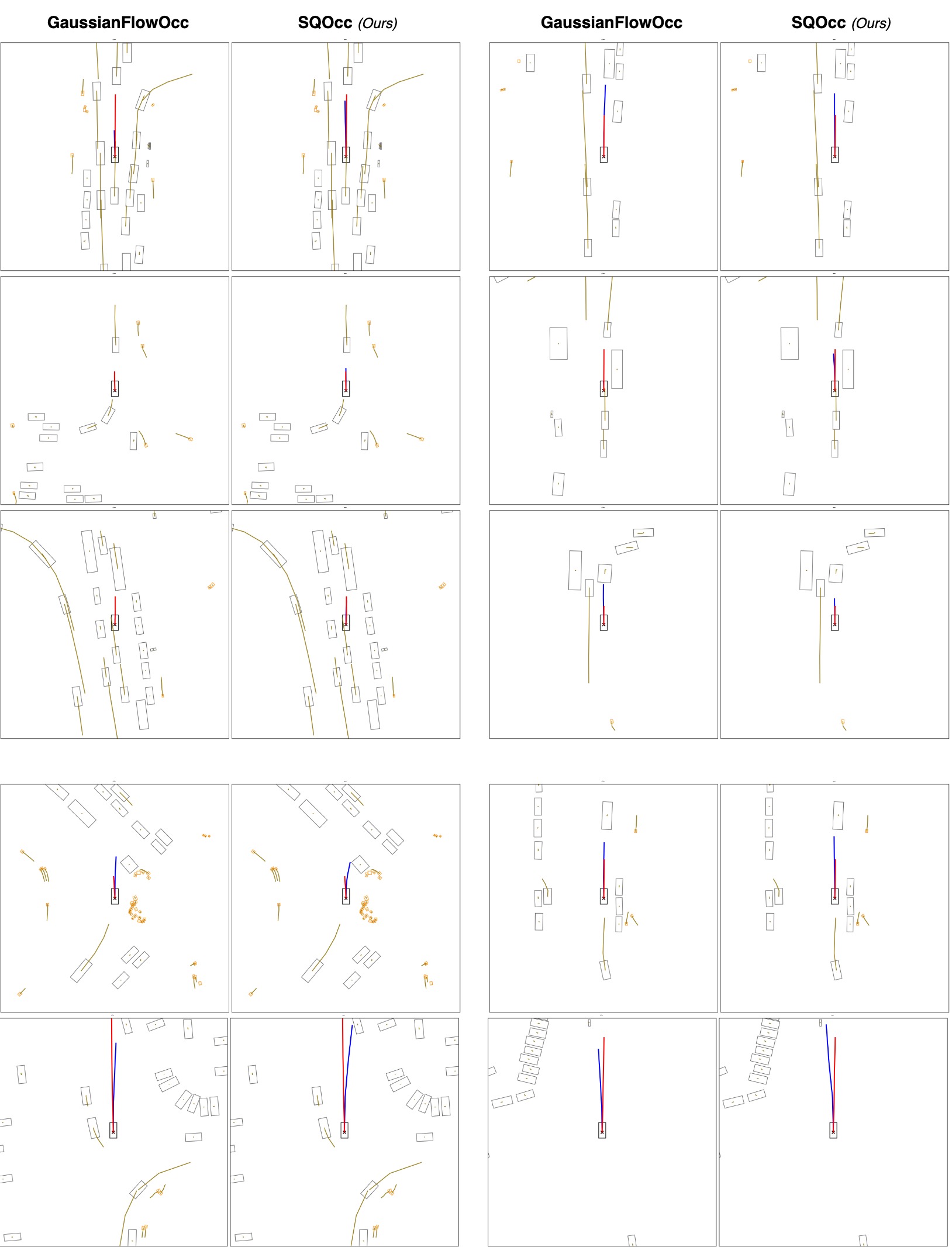}
     \caption{\textbf{Trajectory estimation} for ten scenes. Red line - ground truth trajectory. Blue line - estimation trajectory. Yellow line - ground truth agent trajectory. The top six scenes correspond to collisions in GaussianFlowOcc, with no collision in \methodShort{}. The bottom four scenes correspond to no collision in GaussianFlowOcc, with a collision in \methodShort{}. Best viewed zoomed in.}
    \label{fig:traj}
\end{figure*}

In \autoref{fig:traj}, we visualise ten scenes for trajectory estimation for GaussianFlowOcc and \methodShort{}. Evidently, GaussianFlowOcc typically struggles to move in some instances, with no forward movement resulting in it colliding with the vehicle behind it. Whereas \methodShort{} correctly estimates the movement. Other times, it overshoots and collides with the vehicle in front. In one instance, \methodShort{} does not collide, but it overshoots; however, it correctly moves left to avoid the agents to the right of the vehicle.

For failure instances in \methodShort{}, we mainly see collisions originating from oversteering in one direction. This could be remedied by incorporating further information, such as current ego velocity.

\subsection{NYUv2}
\begin{figure*}
    \centering
    \includegraphics[width=\textwidth]{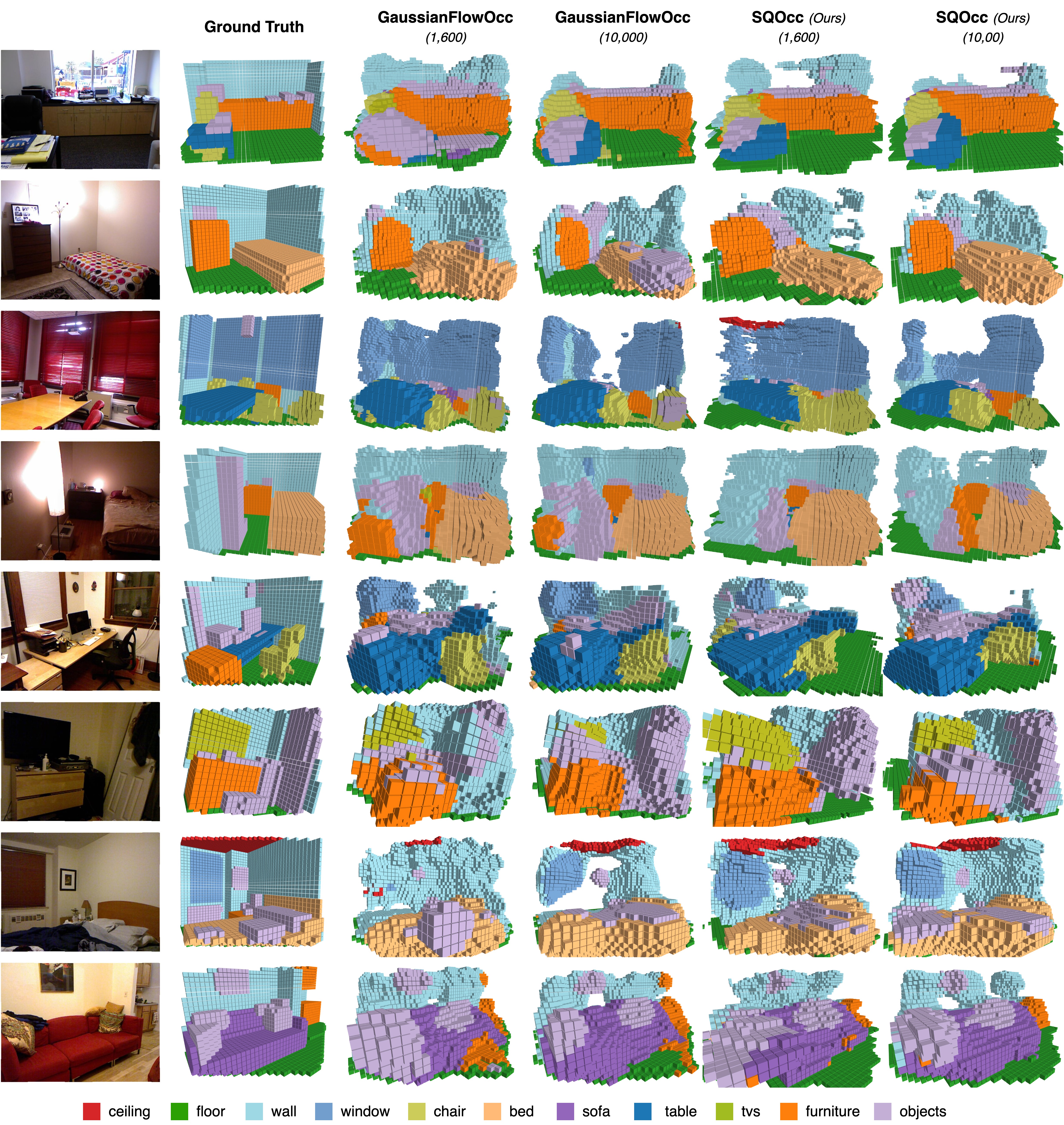}
     \caption{\textbf{Model comparison on the NYUv2 \cite{silberman2012indoor} dataset.} Best viewed zoomed in.}
    \label{fig:nyuv2_vis}
\end{figure*}

In \autoref{fig:nyuv2_vis}, we present visualisations for all models present in results \autoref{tab:nyuv2}. \methodShort{} displays superior performance as was evident in the main results table, providing better estimations for its 1,600 variant compared to the 10,000-primitive GaussianFlowOcc model. Although both models struggle in noticeable areas, namely fine detail objects such as \textit{'chair'}.

\subsection{Additional Visualisations}
\label{subsec:sqocc-r}
\begin{figure*}
    \centering
    \includegraphics[width=0.9\textwidth]{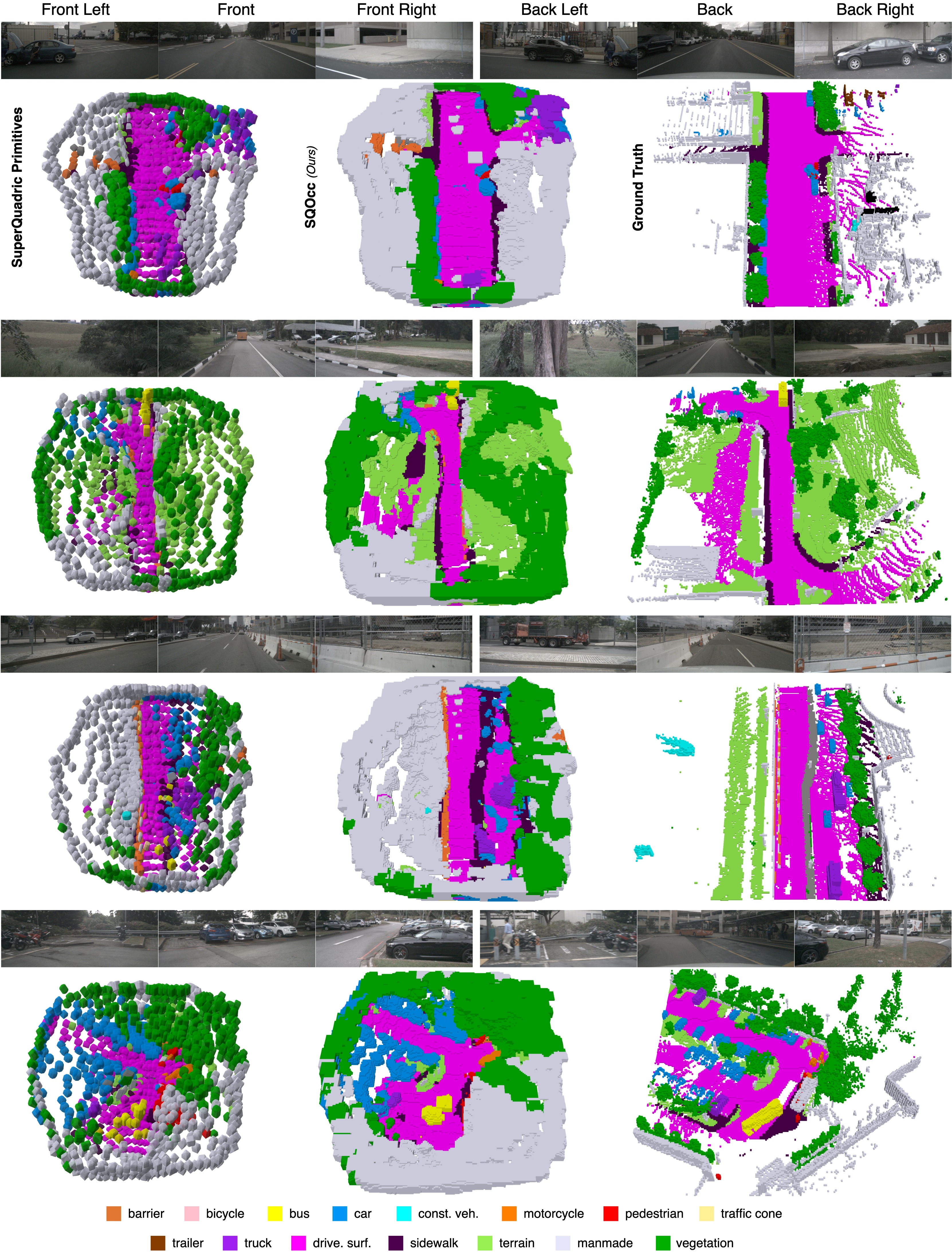}
     \caption{\textbf{Semantic occupancy estimation for SQOcc.} We visualise five samples. The left-most images show the superquadric primitives, the centre images show SQOcc predictions, and the right images show the ground-truth occupancy labels. Best viewed zoomed in.}
    \label{fig:vox_supp_sqnerf}
\end{figure*}

\begin{figure*}
    \centering
    \includegraphics[width=\textwidth]{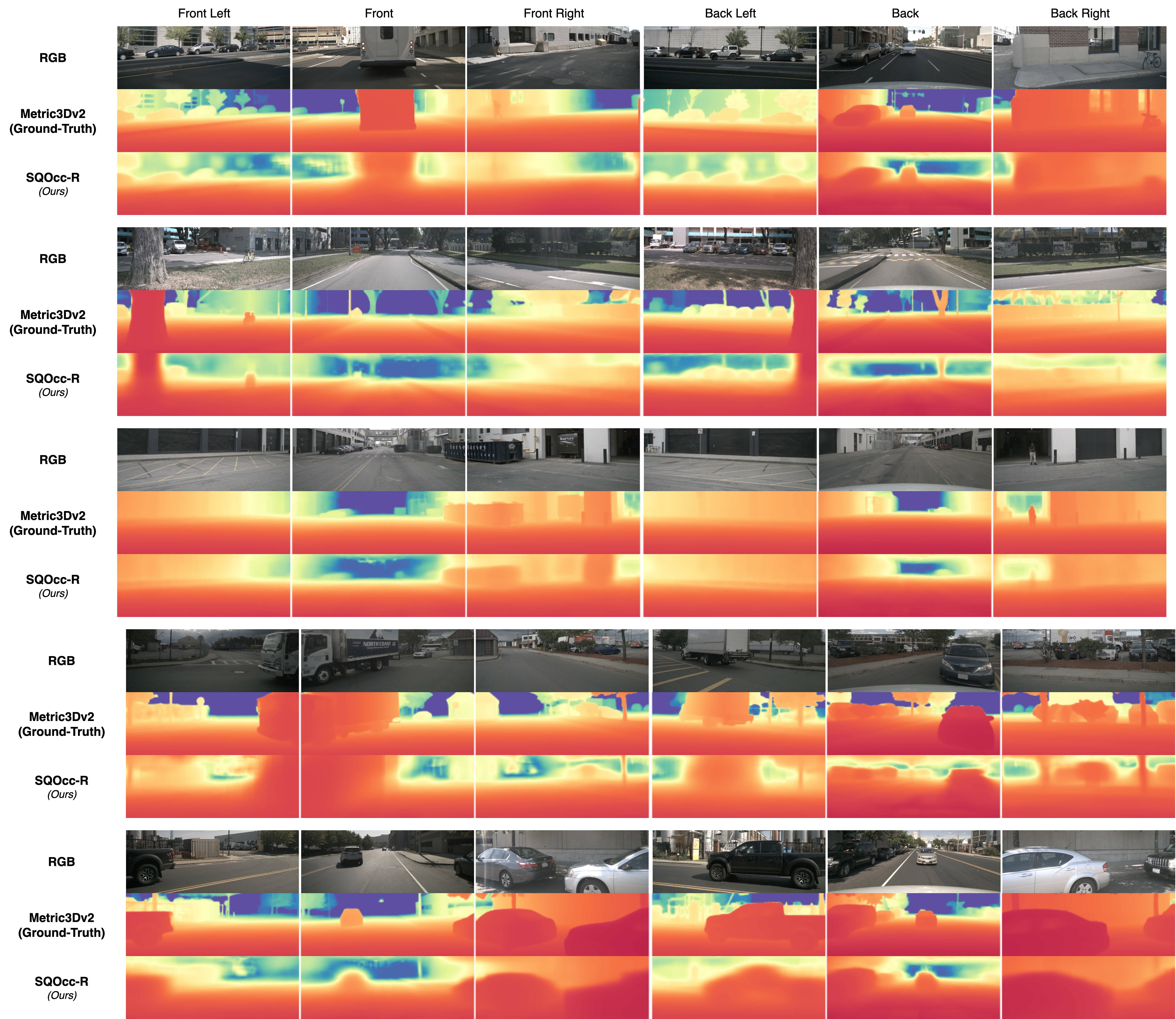}
     \caption{\textbf{Depth estimation for SQOcc-R.} Best viewed zoomed in.}
    \label{fig:depth_supp_sqoccr}
\end{figure*}

For the remaining figures, \autoref{fig:vox_supp_sqnerf} and \autoref{fig:depth_supp_sqoccr}, we present additional semantic occupancy visualisations for \methodShort{}, and depth estimation results for \renderShort{}. Our method displays accurate scene reconstruction alongside coherent depth detection.

\end{document}